\documentclass[Afour,sageh,times]{sagej}

\usepackage{moreverb,url}
\usepackage{multicol}
\usepackage{caption,tabularx,booktabs}
\usepackage[utf8]{inputenc} 
\usepackage[T1]{fontenc}    
\usepackage{hyperref}       
\usepackage{url}            
\usepackage{booktabs}       
\usepackage{amsfonts}       
\usepackage{nicefrac}       
\usepackage{microtype}      
\usepackage{xcolor}         
\usepackage{times}
\usepackage{multirow}
\usepackage{graphicx}
\usepackage{color}
\usepackage{array}
\usepackage{subfigure}
\usepackage{algorithm}
\usepackage{algpseudocode}
\usepackage{amssymb}

\usepackage{wrapfig}
\usepackage{bm}
\usepackage[justification=centering]{caption}
\usepackage[normalem]{ulem}
\newcolumntype{P}[1]{>{\centering\arraybackslash}p{#1}}
\usepackage{natbib}
\usepackage[font=small,skip=8pt]{caption}

\newcommand\BibTeX{{\rmfamily B\kern-.05em \textsc{i\kern-.025em b}\kern-.08em
T\kern-.1667em\lower.7ex\hbox{E}\kern-.125emX}}

\begin{document}
\runninghead{}

\title{Online Omnidirectional Jumping Trajectory Planning for Quadrupedal Robots on Uneven Terrains}

\author{Linzhu Yue\affilnum{1}, Zhitao Song\affilnum{1}, Jinhu Dong\affilnum{1}, Zhongyu Li\affilnum{2}, Hongbo Zhang\affilnum{1},\\ Lingwei Zhang\affilnum{1}, Xuanqi Zeng\affilnum{1}, Koushil Sreenath\affilnum{2}, Yun-hui Liu\affilnum{1}\vspace*{-1.1em}}

\affiliation{\affilnum{1}The Chinese University of Hong Kong, Hong Kong, China \\ 
\affilnum{2}University of California at Berkeley, CA, USA \\
}
\corrauth{Yun-hui Liu, Department of Mechanical and Automation Engineering, The Chinese University of Hong Kong, Shatin, Hong Kong, China}
\email{yhliu@mae.cuhk.edu.hk}

\keywords{Online Trajectory planning, Quadrupedal Jumps, Re-localization, Differential Evolution\vspace*{-0.5em}}

\begin{abstract}
Natural terrain complexity often necessitates agile movements like jumping in animals to improve traversal efficiency. To enable similar capabilities in quadruped robots, complex real-time jumping maneuvers are required. 
Current research does not adequately address the problem of online omnidirectional jumping and neglects the robot's kinodynamic constraints during trajectory generation. 
This paper proposes a general and complete cascade online optimization framework for omnidirectional jumping for quadruped robots. Our solution systematically encompasses jumping trajectory generation, a trajectory tracking controller, and a landing controller. 
It also incorporates environmental perception to navigate obstacles that standard locomotion cannot bypass, such as jumping from high platforms. 
We introduce a novel jumping plane to parameterize omnidirectional jumping motion and formulate a tightly coupled optimization problem accounting for the kinodynamic constraints, simultaneously optimizing CoM trajectory, Ground Reaction Forces (GRFs), and joint states. To meet the online requirements, we propose an accelerated evolutionary algorithm as the trajectory optimizer to address the complexity of kinodynamic constraints. 
To ensure stability and accuracy in environmental perception post-landing, we introduce a coarse-to-fine relocalization method that combines global Branch and Bound (BnB) search with Maximum a Posteriori (MAP) estimation for precise positioning during navigation and jumping. The proposed framework achieves jump trajectory generation in approximately 0.1 seconds with a warm start and has been successfully validated on two quadruped robots on uneven terrains. Additionally, we extend the framework's versatility to humanoid robots.
\end{abstract}

\maketitle

\section{Introduction} \label{sec:introduction}
Quadruped robots have demonstrated significant potential in navigating complex terrains and performing agile maneuvers, similar to natural quadrupeds. Achieving agile jumping capabilities—such as omnidirectional jumps and descending from high platforms—is crucial for expanding the operational scope of these robots in real-world environments. The primary challenge involves generating real-time jumping trajectories that adhere to the dynamic and kinematic constraints of underactuated systems while ensuring stable landings to minimize hardware stress. 

Previous studies have showcased quadruped robots executing backflips and forward jumps using offline trajectory optimization methods, such as~\cite{katz2018low,ding2020kinodynamic,nguyen2019optimized,gilroy2021autonomous}. However, the offline-generated trajectories are unsuitable for real-time applications and are limited to specific scenarios. Advances in locomotion gaits have also been achieved through model-based approaches, like \cite{di2018dynamic}, 
enabling robots to handle various obstacles. But, the obstacle-crossing ability of these pure locomotion methods is limited. Additionally, some works have focused on planning the Center of Mass (CoM) trajectory and employing kinodynamic tracking controllers for jumping tasks, like~\cite{chignoli2021rapid}.  Approaches that consider only CoM trajectory optimization often neglect comprehensive dynamic and kinematic constraints.
This requirement hinders direct implementation with tracking controllers or low-level controllers. Moreover, most existing studies construct the jumping problem depending on specific platforms affecting the algorithm's scalability and concentrate on the take-off phase without designing landing controllers capable of mitigating impact forces during descent, which is essential for reducing hardware damage. Many existing jumping frameworks may not fully integrate environmental perception or address relocalization challenges from map positioning failures caused by high-impact landings.

To address these issues, we introduce an omnidirectional jumping framework that enables real-time trajectory optimization while accounting for the robot's dynamic and kinematic constraints. An omnidirectional jumping plane is utilized to formulate the optimization problem for ground reaction forces (GRFs) and the CoM trajectory based on a reduced dynamics model. To accelerate optimization by meeting real-time requirements, the framework utilizes Latin Hypercube Sampling(LHS)\cite{ayyub1989structural}, enhancing the initial population quality and performing configuration space analysis intensively to reduce the searching space. Moreover, we propose a warm start strategy with the help of the Pre-motion library and
carefully design a prioritized fitness function to ensure the stable convergence of the algorithm.
\begin{table}[htbp]
\centering
\scriptsize
\caption{Supplementary video list in this paper}
\label{tab:video_list}
\begin{tabular}{ll}
\toprule
\textbf{Video} & \textbf{Links} \\
\midrule
1 & \href{https://youtu.be/fDye1gVxoMQ}{Summary Video} \\
2 & \href{https://youtu.be/1YGi2pMNIdI}{Omnidirectional Jumping} \\
3 & \href{https://youtu.be/PQldIYHTprs}{Agile Motions} \\
4 & \href{https://youtu.be/ny0Y5lCedNQ}{Jumping with Navigation} \\
\bottomrule
\end{tabular}
\end{table}
An evolutionary algorithm solves the complex constrained optimization problem, producing outputs that include GRFs, CoM trajectory, joint configurations, and timing for each jump stage. An impedance controller is implemented as an active compliance landing controller to handle large impact forces during landing. Additionally, a Whole Body Controller (WBC) is employed for trajectory tracking, and a proportional-derivative (PD) controller manages the flight phase, ensuring systematic control throughout the jumping process. Environmental perception is integrated to enable jumping on uneven terrains, combined with navigation capabilities. A coarse-to-fine relocalization method is proposed to address localization failures to enhance localization stability after high-impact landings.

\subsection{Contribution}
This work presents a novel and comprehensive framework for online jump trajectory optimization, with the following key contributions:
\begin{enumerate}
\item \textbf{Development of a novel and general framework for quadrupedal omnidirectional jumping}: 
We introduce a general and comprehensive omnidirectional jumping trajectory framework that includes online trajectory optimization, real-time trajectory tracking, and an active compliant landing controller. This framework enables effective jumping in multiple directions—including forward, backward, left, right, and diagonal movements like front-right. By integrating environmental perception, the framework allows the quadruped robot to overcome uneven terrain through jumping. Additionally, this method is successfully extended to support bipedal jumping.
\item \textbf{Novel online optimizable jumping problem definition and solution}: By constructing an omnidirectional jumping plane and formulating the optimization problem using a reduced-order model, we generate ground reaction forces (GRFs) and Center of Mass (CoM) trajectories along the jumping plane, with a feasible target point as input. Kinodynamic constraints are integrated into the optimization to ensure dynamic and kinematic feasibility. To achieve real-time optimization within approximately 0.1 seconds, we employ a differential evolution algorithm enhanced with strategies such as configuration space considerations, Latin Hypercube Sampling for efficient initial sampling, a pre-motion library as a warm start, and an innovatively designed fitness function (discussed in Sec. \ref{sec:jumping_to} and Sec. \ref{sec:application_task}).

\item \textbf{Advanced localization robustness with large impact upon landing}: To ensure the accuracy of environmental perception assisted jumping on uneven terrain. We introduce a robust localization recovery mechanism using a novel coarse-to-fine relocalization method within our jumping framework. In the coarse stage, a global Branch-and-Bound (BnB) search algorithm provides an accurate initial pose. The refinement stage involves a Maximum a Posteriori (MAP) estimation that fuses sensor data from Inertial Measurement Units (IMUs), motor encoders, and LiDAR to achieve precise localization (as detailed in Sec. \ref{sec1:reliable_localization}).

\item  \textbf{Extensive real-world experiments on two quadrupedal robots and simulations on a humanoid robot validate the effectiveness of our novel jumping framework for legged systems}: Through comprehensive simulations and real-world experiments—including multi-directional jumps and autonomous navigation (including indoor and outdoor) involving substantial landing impacts, such as jumping from elevated platforms—we have validated the robustness and practicality of our algorithm (see in Tab. \ref{tab:video_list}). Furthermore, the jumping framework has been extended to accommodate a full-sized humanoid robot in simulated environments, demonstrating its versatility and effectiveness (as detailed in Sec. \ref{sec:simulation_and_hardware}).
\end{enumerate}
\subsection{Related Works }
Model-free reinforcement learning (RL) has achieved significant advancements in quadruped locomotion, as highlighted in recent literature, like \cite{lee2020learning, li2023robust, choi2023learning}. However, the direct application of these techniques to jumping tasks faces challenges due to the limited jumping duration and high impact forces encountered during landing. This limitation has prompted researchers to explore the application of RL in enhancing jumping capabilities in robotics in~\cite{liu2023distance, bellegarda2024robust}. In particular, combining jumping with traditional locomotion has shown promise in enabling robots to navigate complex obstacles, such as desks and gaps in \cite{cheng2024extreme, zhuang2023robot}. Additionally, the integration of jumping functionalities has been explored in specific robotic tasks, including improving the agility of robotic goalkeepers in \cite{huang2023creating}. Notable achievements in this area include the development of robots capable of performing long-distance forward jumps, with some achieving distances greater than their length, and adopting cat-like landing strategies to ensure smooth touch-down and effective impact absorption like \cite{bellegarda2024robust, kurtz2022mini, rudin2021cat}. Despite these advancements, research on omnidirectional jumping and developing a unified control policy for executing various complex jumping maneuvers while ensuring post-landing hardware integrity remains limited.

Gradient-based trajectory optimization has emerged as an important technique in enhancing quadruped robot agility, enabling precise trajectory following through leveraging the robot's dynamic models for optimal joint trajectory computation, like \cite{bledt2018cheetah, mastalli2020motion, jenelten2020perceptive}. This approach has been similarly practical in augmenting jumping capabilities, as demonstrated by MIT researchers who utilized gradient-based algorithms to enable a Cheetah3 robot to ascend a 0.76m platform in~\cite{ nguyen2019optimized}. Similarly, this methodology facilitated the Mini Cheetah's ability to execute real-time, efficient, and dynamically stable 3D omnidirectional jumps predicated on predefined contact sequences and desired launch velocities in~\cite{chignoli2021rapid, chignoli2021humanoid}. Further advancements have been achieved through collocation-based optimization for jump execution within defined safety constraints in \cite{gilroy2021autonomous} and Mixed-Integer Convex Optimization for forward jumps that circumvent the need for initial guesses, targeting the robots with two legs in \cite{ding2020kinodynamic}. Despite advancements, existing approaches to quadruped robot jumping often mandate predefined contact sequences or velocities, do not consider kino-dynamics at the trajectory generation, and lack support for omnidirectional jumps. These methods, using online or offline optimization, have not achieved the simultaneous goals of speed, stability, and real-time execution. Furthermore, there is a pressing demand for an autonomous technique that enables omnidirectional jumping based directly on the target position, eliminating reliance on preset motion references.

Heuristic algorithms offer an efficient solution to optimization challenges with complex constraints, presenting novel avenues for jump motion generation. Differential Evolution (DE), introduced by~\cite{ahmad2022differential}, stands out as a heuristic algorithm that has seen applications across robotics, signal processing, and other domains for tackling intricate optimization tasks. Our prior research utilized DE to obtain offline jumping trajectories for quadruped robots and get impressive results in~\cite{song2022optimal}. Nonetheless, this offline methodology falls short for systems necessitating online re-planning. Additionally, \cite{yue2023evolutionary} using Latin hypercube sampling (LHS) for initializing DE populations has enhanced convergence rates in scenarios with low-dimensional spaces in~\cite{wang2022surrogate}. But the method introduced in \cite{yue2023evolutionary} can not achieve omnidirectional jumps.

In recent years, lidar odometry and mapping methods have been extensively utilized in autonomous driving and robotic navigation. Several notable frameworks have been proposed, including LOAM \citep{zhang2014loam}, LeGO-LOAM \citep{shan2018lego}, LIO-SAM \citep{liosam}, and FAST-LIO \citep{fastlio}. Among these, FAST-LIO has been widely adopted due to its superior accuracy and efficiency. However, these frameworks lack failure recovery strategies, making them unsuitable for direct application in quadrupedal robot navigation tasks involving jumping motions, where localization tracking failures frequently occur during landing. Therefore, a re-localization component is essential for quadrupedal robot navigation frameworks with jumping. The re-localization component in lidar SLAM essentially involves point cloud registration. To enhance the robustness of re-localization, we propose a global search method using the Branch-and-Bound (BnB) algorithm, which can obtain the global optimal solution and is well-suited for robust estimation problems. Unlike existing 6D point-to-point registration approaches employing BnB~\cite{yang2015go,campbell2016gogma,liu2018efficient}, we introduce a 3D point-to-plane registration algorithm tailored for our re-localization problem. This approach improves the algorithm's efficiency and accuracy while maintaining robustness.

\section{Overview of This Work}
The remaining content of this paper is structured as follows. Sec. \ref{sec:dyanmics_sec} briefly introduces the dynamics models of robots and defines omnidirectional jumping.
Sec. \ref{sec:jumping_to} details the construction of the jump trajectory optimization problem, including objectives, constraints, and acceleration methods. Sec. \ref{sec:wbc_tracking} covers using a whole-body controller for jump trajectory tracking. Sec. \ref{sec:impedance_landing} explains the simplified impedance control with active compliance to mitigate the impact of landing. Sec. \ref{sec1:reliable_localization} details our proposed reliable localization method to ensure accurate localization during jumps. Sec. \ref{sec:application_task} demonstrates the application of the method from Sec. \ref{sec:jumping_to} to both bipedal and quadrupedal jumps. Sec. \ref{sec:simulation_and_hardware} presents extensive jumping experiments that validate the stability and real-time performance of the proposed cascade jumping framework.
Finally, we summarize our paper and discuss future work in Sec. \ref{sec:conclution_results}.

We have significantly advanced our previous work at the conference, including~\cite{song2022optimal} and~\cite{yue2023evolutionary} by introducing several novel strategies. 1) we developed a novel jumping plane that unifies online and offline optimizations, enhancing the framework's versatility and efficiency. This advancement extends the framework to other legged robots, including successful implementations of bipedal robots. Utilizing the jumping plane, we formulated a new optimization problem enabling omnidirectional jumps on a two-dimensional plane, effectively reducing the action dimensionality for maneuvers from 12 to 6. Additionally, a pre-motion library facilitates real-time trajectory optimization within approximately 0.1s. 2) to minimize landing impact, we integrated an impedance controller, which significantly cushions the robot's descent, as evidenced by high-platform jumping tests. 3) for real-world applicability, we incorporated environment mapping for navigation and obstacle detection. We addressed jump-induced re-localization challenges by implementing a branch-and-bound search for coarse localization, followed by precise corrections using IMU data, joint angles, and LiDAR data. 4) we conducted additional experiments encompassing navigation-based jumping, obstacle recognition, and landing controller comparisons, further validating the framework's robustness.
\section{MODELS AND DYNAMICS}  \label{sec:dyanmics_sec}
\subsection{Ominidirectional Jumping Model}\label{sec:omini_jumping_mode}
\begin{figure}[tb]
\centering
\vspace{-0.2cm}
\includegraphics[width=\linewidth]{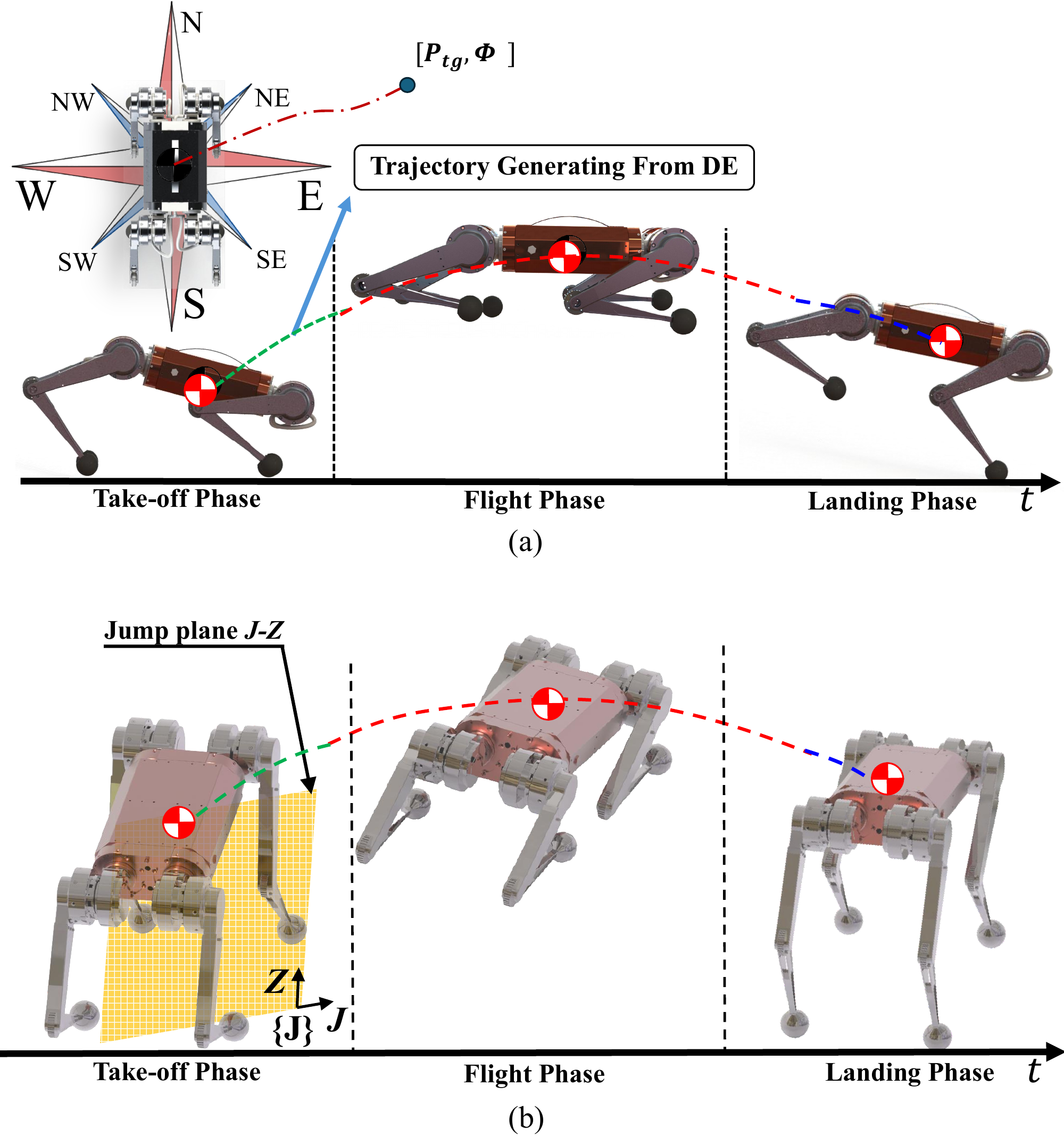}
\caption{The omnidirectional jumping task consists of Take-off, Flight, and Landing phases. The green, red, and blue dot lines show the jumping trajectory. The yellow plane shows the jumping plane. (a) Lateral view of the jumping task phases. (b) Top view of the jumping task phases.}\label{fig:omnidirectional_phase}
\vspace{-0.3cm}
\end{figure}

\begin{figure*}[h]
\centering
\includegraphics[width=\linewidth]{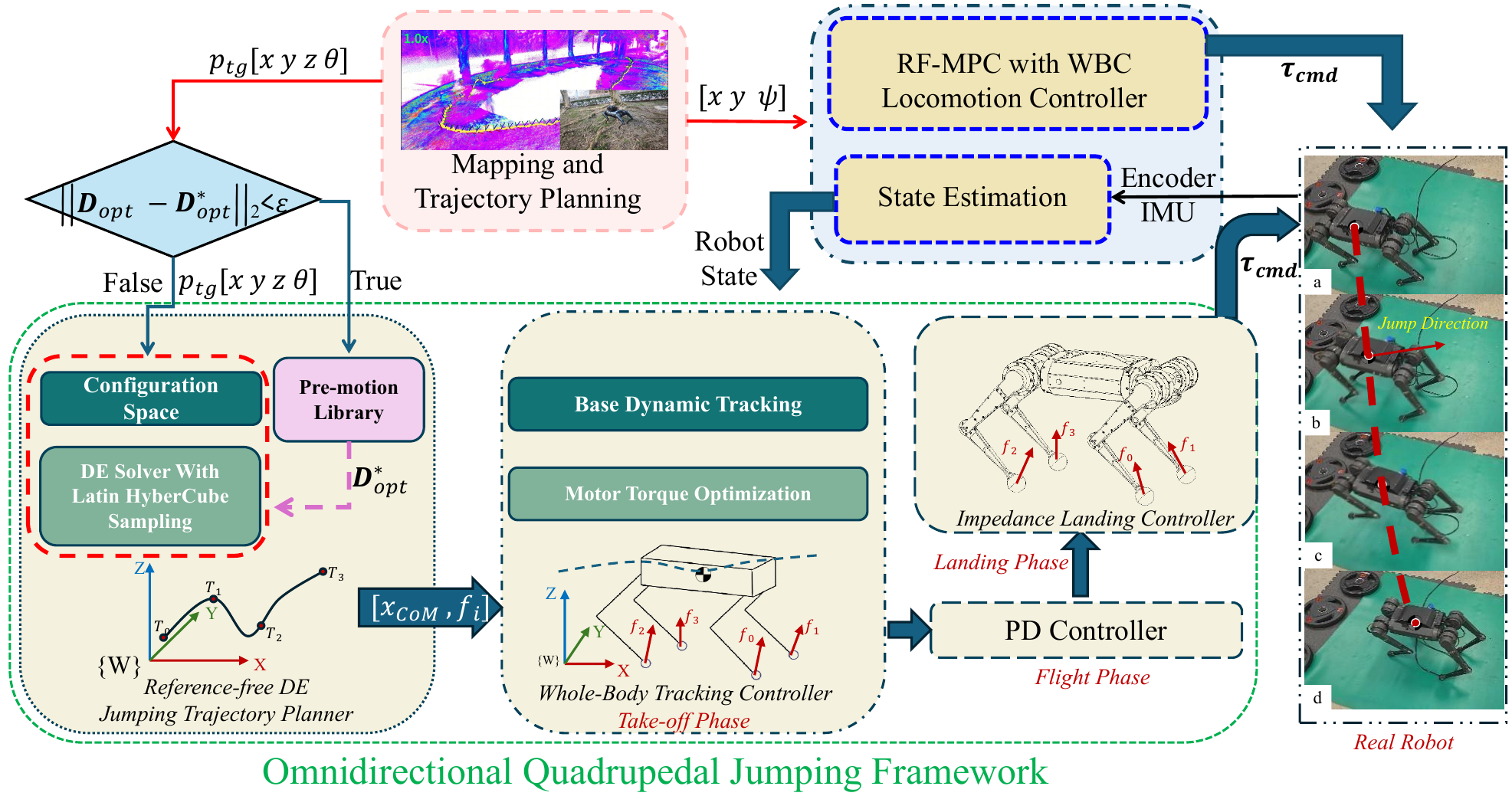}
\caption{An overview of the proposed online cascading omnidirectional jumping framework, incorporating Latin Hypercube Sampling 
 (LHS), Differential Evolution (DE), Offline initial guess library for Differential Evolution algorithm (Pre-Motion Library), configuration space (C-space), and navigation modules. The red dashed line represents the mapping and trajectory planning with navigation modules, while the blue blocks depict the navigation trajectory tracking controller. The green dashed line indicates the blocks corresponding to the jumping controller.}
\label{fig:jump_framework}
\end{figure*}

We segment omnidirectional jumping into three phases: take-off, flight, and landing (see Fig. \ref{fig:omnidirectional_phase}). Depending on the chosen jumping method, the take-off phase can involve either all four legs leaving the ground simultaneously or the front and rear feet lifting off sequentially (such as backflip). We define omnidirectional jumping as the ability to execute jumps at any specified angle around the plane composed of the x-direction and y-direction of the robot body following the desired CoM position and a desired yaw angular ($[ \bm P_{tg},\Phi]\in \mathbb{R}^{4}$). Then, the proposed framework effectively controls the robot to jump in desired directions, such as forward or forward-right jumps (details see Fig. \ref{fig:omini_jump_plane}). Each phase utilizes specific controllers that are switched via time schedules. During the take-off phase, an evolutionary algorithm–based trajectory planner generates the center of mass (CoM) trajectory, desired joint angles, and ground reaction forces (GRFs). These trajectories are then supplied to the Whole Body Control (WBC) controller for tracking using full-order dynamics. Given the legs' relatively small mass in the flight phase, a Proportional-Derivative (PD) controller maintains the preset leg joint angles as the body's inertia from leg movement alone is insufficient for significant displacement. Lastly, the landing phase employs an impedance controller integrating active compliance, as shown in Fig. \ref{fig:jump_framework}.

\subsection{Single Rigid Body (SRB) for Trajectory Optimization}\label{sec:srb_model}
This section outlines the robot system states and control variables, with a detailed description of the Single Rigid Body (SRB) model. This adapted model is crucially employed in the take-off phase of the jumping framework presented in this article. The reduced-order robot state $\boldsymbol{x}$ can be written as:
\begin{subequations}
\begin{eqnarray}
& \bm x:=\lbrack {\bm P_{CoM}^T} \quad {\bm \Theta^T}\quad {\bm V_{CoM}^T} \quad  {\bm \omega^T_B}\rbrack^T \in \mathbb{R}^{12}\label{q_joint_2}\\
&{\bm Q}:=[{\bm q_{i}}^T \quad {\dot{\bm q}_{i}}^T]^T \in \mathbb{R}^{24} \label{q_joint_1}
\end{eqnarray}
\end{subequations}
Where ${\bm P_{CoM}} \in \mathbb{R}^{3}$ is the position of the robot's body center of mass (CoM) w.r.t. inertial frame; ${\bm \Theta =[\psi \ \phi \ \theta]}$ represents the Euler angles of the robot's body; ${\bm V_{CoM} \in \mathbb{R}^{3}}$ is the velocity of the CoM. ${\bm \omega}_B \in \mathbb{R}^{3}$ is the angular velocity of CoM w.r.t. the robot frame $\{B\}$. $\bm q_i \in \mathbb{R}^3$ and $\bm \dot{\bm q}_i \in \mathbb{R}^3$ are the joint angles and velocities of each leg. $i=\{1,2,3,4\}$ is the number of feet. The ground reaction forces (GRF) $\bm f_{i} \in \mathbb{R}^3$ at each contact point consists of the dynamic system control input ${\bm u}:=[{\bm f_{i}}]$. ${\bm r_i}$ is the vector from CoM to the robot foot. Then, The linear acceleration of CoM and the angular acceleration of the base are shown:
\begin{subequations}
\begin{align}
m\ddot{\bm P}_{com}&={\sum_{i=0}^{n_c}\boldsymbol{f}_i}- m{\bm g} \label{acc}\\
\frac{\mathrm{d}}{\mathrm{d} t}({ \bm \omega_B})& = ^B\mathbf{I}^{-1}(\sum_0^{n_c}\bm{r}_i\times  \boldsymbol{f}_i -{\bm \omega_B}\times{^B\mathbf{I}}{\bm \omega}_B)
\label{rcc},
\end{align}
\label{centroidal_model}
\end{subequations}
\noindent Where ${\bm g}\in\mathbb{R}^3$ represents gravitational acceleration. $n_c$ represents the number of contacts. ${^B\mathbf{I} } \in \mathbb{R}^{3\times3}$ is the robot's rotation inertial tensor, which is assumed as a constant in this work (details in Table \ref{tab:robot_mass_param}).

\subsection{Floating Base Full-order Model for Tracking Controller} \label{floating_model}
This section introduces the full-order dynamic model based on the floating base theory for whole-body control (WBC) to track the trajectory in the take-off phase. The whole body dynamics of the floating base can be written in the following standard form.
\begin{equation}
\mathbf{A}(\boldsymbol{q}) \ddot{\boldsymbol{q}}+\bm{D}(\boldsymbol{q}, \dot{\boldsymbol{q}})=\mathbf{S}_f \boldsymbol{\tau}+\mathbf{J}_c^{T} \boldsymbol{f}_c
\label{full_order_dy}
\end{equation}
Where $\bm \ddot{q}= \left[\begin{array}{c}
\ddot{\mathbf{q}}_f \\
\ddot{\mathbf{q}}_j
\end{array}\right]$ is general joint acceleration. $\ddot{\mathbf{q}}_f \in \mathbb{R}^{6}$ and $\ddot{\mathbf{q}}_j \in \mathbb{R}^{12}$ are the acceleration of floating base and joint accelerations, respectively. $\mathbf{J}_c\in \mathbb{R}^{6 \times 12}$ is the contact Jacobian Matrix. $\mathbf{A}(\boldsymbol{q})$ is the joint inertial matrix, $\bm D(\boldsymbol{q}, \dot{\boldsymbol{q}})$ is the bias force matrix including Coriolis force, gravitation force. $\mathbf{S}_f=\left[\begin{array}{cc}
\mathbf{0}_{6 \times 6} & \mathbf{0} \\
\mathbf{0} & \mathbb{I}_{12 \times 12}
\end{array}\right]$ is the input mapping matrix.
\section{Trajectory Optimization for Jumping }
\label{sec:jumping_to}

This section details the general formulation of optimization equations (details in Sec. \ref{sec:optimization_formulation}), optimization variables (details in Sec. \ref{sec:opt_variables}), and cost function (details in Sec. \ref{sec:p_fit_func}) based on the dynamic model of a legged robot's jumping motion. It also presents a method for transforming optimization variables into physically meaningful robot parameters (details in Sec. \ref{sec:opt_variables}), enabling the use of configuration space and kino-dynamics to reduce the optimization space for real-time optimization (details in Sec. \ref{sec:kino_c_space}). Unlike other studies that employ gradient-based approaches, our proposed method constructs the optimization problem using an evolutionary-based solver to determine jumping motions. 

\subsection{Jumping Phase Trajectory Optimization Formulation}\label{sec:optimization_formulation}

Jumping generally consists of three phases: take-off, flight, and landing. By optimizing the take-off phase, we can obtain a jump trajectory that satisfies kino-dynamic constraints and has optimal energy consumption. To this end, we establish a priority hierarchical fitness function for this problem. 

\begin{eqnarray}
\begin{aligned}
& \bm S_{\textrm{opt}}=\underset{\bm S}{\text{argmin}}
& & \sum\limits_{n=1}^{Level}W_{n}(10^{n+3}+10^{n}\sigma_{n})+ \zeta \\
& \text{subject to}
& & \bm{x}_{k+1} = \bm{x}_k + g(\bm u_k,\bm x_k){\Delta t}  \\
&&&\boldsymbol{x}_{k} \in \mathbb{X}, k=1,2, \cdots, N \\
&&&\boldsymbol{u}_{k} \in \mathbb{U}, k=1,2, \cdots, N\\
&&&\boldsymbol{q}_{k} \in \mathbb{Q}, k=1,2, \cdots, N\\
&&&\bm{x}_{0} = \bm{x}_{\mathrm{init}},
\bm{x}_N =\bm{x}_{\textrm{target}}
\end{aligned}
\label{optimization_formula}
\end{eqnarray}
where $\bm S \in \mathbb{R}^{12}$ represents the trajectory optimization variable. $ \bm S_{\textrm{opt}}$ is the optimal results in this problem. ${\sigma}_n \in \mathbb{R}$ represents penalty function of kino-dynamic constraints. $\zeta=\int_{0}^T(|\bm{\tau}(t)\bm{\dot{q}}(t)|)dt$ indicates the mechanical energy consumption of the take-off phase. $\Delta t$ is the integration step time.
$Level \in \mathbb{R}$ is the maximum number of constraints. $W_n$ is the penalty coefficient. $\bm u_k$ is the robot control input, which is the ground reaction force in this paper. $N$ represents the total number of trajectory track points. $\mathbb{X}$, $\mathbb{U}$ and $\mathbb{Q}$ are the feasible sets that satisfies the constraints. $\bm{x}_0$ and $\bm{x}_N$ are robot's initial and the desired state; $\bm{x}_{k+1} = \bm{x}_k + g(\bm u_k,\bm x_k){\Delta t}$ represents the SRBD combination of  (\ref{acc}) and  (\ref{rcc}) which used for forward integration.

\subsection{Ominidirectional Jumping Plane}\label{sec:jumping_plane_sec}
\begin{figure}[htbp]
\centering
\includegraphics[width=\linewidth]{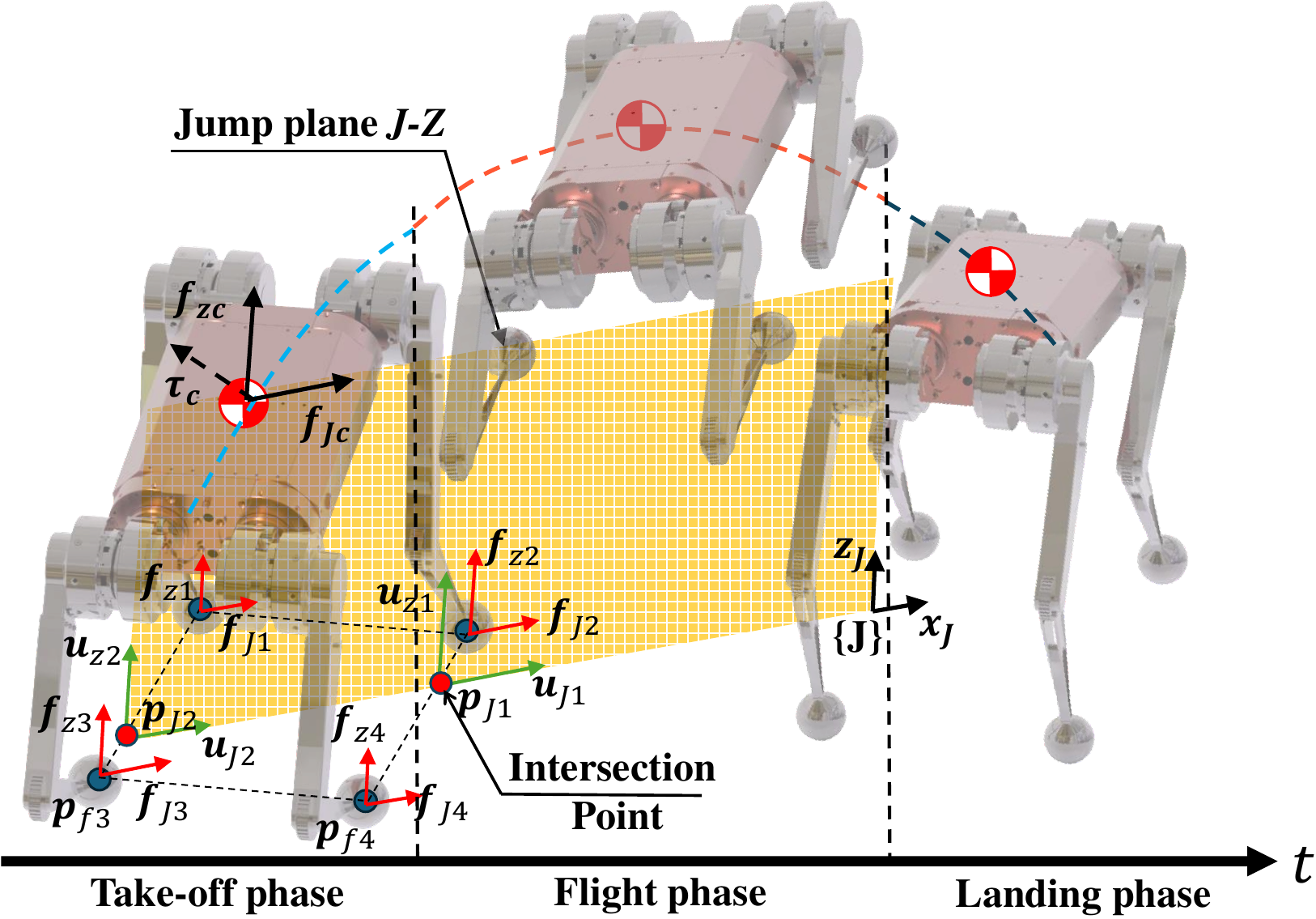}
\caption{Side view of the omnidirectional jump. The orange region denotes the jump plane $J$-$Z$. The green coordinate system represents the intersection between the jumping plane and the polygon formed by the robot's four feet projected onto a 2D plane. The coordinates of this intersection in the body frame are denoted as $\bm{p}_{j1}$ and $\bm{p}_{j2}$. The variables $\bm{u}_{j1}$ and $\bm{u}_{j2}$ represent the resultant forces along the jumping plane, while $\bm{u}_{z1}$ and $\bm{u}_{z2}$ denote the resultant forces along the z-axis in frame \{J\}. The positions of the feet in the body frame are indicated by $\bm{p}_{f1}, \bm{p}_{f2}, \bm{p}_{f3}$, and $\bm{p}_{f4}$. The components of the jumping-plane resultant forces acting on each foot are represented by $\bm{f}_{j1}, \bm{f}_{j2}, \bm{f}_{j3}$, and $\bm{f}_{j4}$. The forces $\bm{f}_{zc}$ and $\bm{f}_{jc}$ denote the resultant forces at the center of mass (CoM), with $\bm{\tau}_{c}$ indicating the resultant torque in the jumping direction. Detailed explanations and applications are provided in Sec. \ref{sec:opt_variables} and in Sec. \ref{sec:application_omini}.}
\label{fig:omini_jump_plane_detail}
\end{figure}
\begin{figure}[htbp]
\centering
\includegraphics[width=0.8\linewidth]{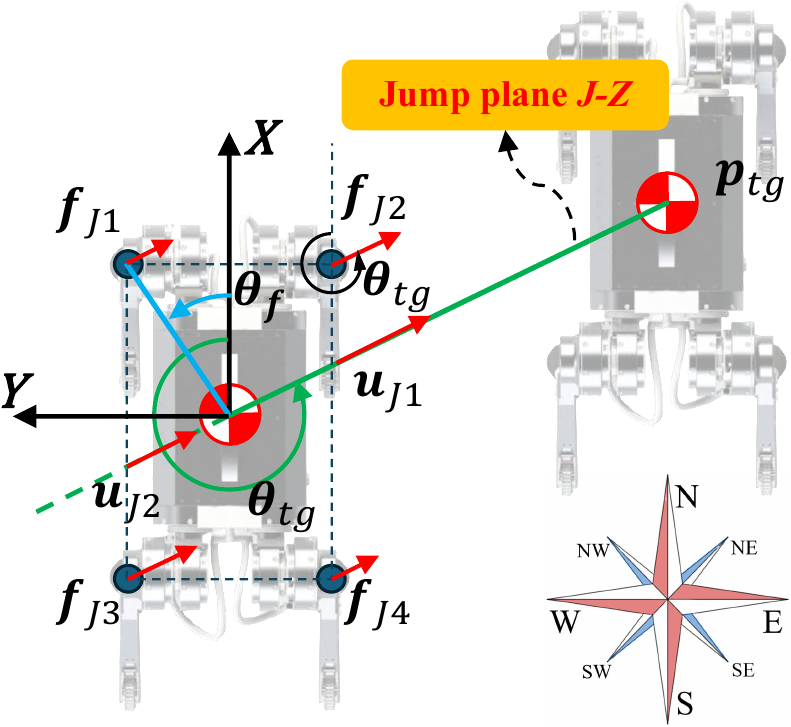}
\caption{
Top view of the omnidirectional jump. The green line denotes the jump plane $J$-$Z$, while \( \bm{p}_{tg} \) and \( \bm{\theta}_{tg} \) represent the target position within the body coordinate system and the angle of rotation of the jump plane relative to the X-axis, respectively. The angle \( \bm{\theta}_f \) defines the foot's orientation around the z-axis in the body coordinate system, distinguishing the direction of the jump plane. Red arrows illustrate the resultant forces exerted by the jump plane on the center of mass. \( \bm{u}_{J1} \) and \( \bm{u}_{J2} \) are the components of the resultant force decomposed for each foot. Detailed explanations are provided in Sec. \ref{sec:opt_variables}.}
\label{fig:omini_jump_plane}
\end{figure}

To achieve an omnidirectional jump, we define a jump plane $J$-$Z$ that is perpendicular to the $X$-$Y$ plane and forms an angle $\theta_{tg} \in$ [0,\:360]$^\circ$ with the $X$-$Z$ plane, as illustrated in Fig. \ref{fig:omini_jump_plane_detail} and Fig. \ref{fig:omini_jump_plane}. To streamline the complexity of omnidirectional jump planning, we assume the force perpendicular to the $J$-$Z$ plane to be zero. Consequently, we focus on planning only the forces $\bm u_{Jj},\bm u_{zj}$
within the $J$-$Z$ plane (Where $j = 1,2$ is the intersection of the quadrilateral formed by the jumping plane and the four-foot positions on the 2-dimensional projection surface). These forces are directly related to the forces at the foot's end as follows:
\begin{subequations}\label{fj}
\begin{align}
& \bm f_{Jl} = \bm u_{Jj}\frac{|\bm p_{Jj}-\bm p_{fm}|}{|\bm p_{fm}-\bm p_{fl}|} \label{fj:A}\\
& \bm f_{Jm} = \bm u_{Jj}-\bm f_{Jl} \label{fj:B}\\
& \bm f_{zl} = \bm u_{zj}\frac{|\bm p_{Jj}-\bm p_{fm}|}{|\bm p_{fm}-\bm p_{fl}|} \label{fj:C}\\
& \bm f_{zm} = \bm u_{zj}-\bm f_{zl} \label{fj:D}\\
 & \bm f_x = \bm f_Jcos(\theta_{tg}) \label{fj:E}\\
  & \bm f_y = \bm f_Jsin(\theta_{tg}) \label{fj:F}
\end{align}
\label{eqn:jumping_plane}
\end{subequations}

\noindent where $\bm p_{Jj}$ represents the position of $\bm u_{Jj}$ w.r.t body frame. $\bm p_{fm}$ and $\bm p_{fl}$ represent the foot position in the body frame (Where m,n  represent the indices of the foot coordinates, the value refers to  (\ref{eqn:jumping_case})). $\bm f_{Jl}$,$\bm f_{zl}$,$\bm f_{Jm}$,$\bm f_{zm}$represent the foot force. $\mathbf{f}_x$ and $\mathbf{f}_y$ represent the force components along the x-axis and y-axis directions in the jumping plane for each foot, as shown in Fig. \ref{fig:omini_jump_plane_detail}. We categorize the foot force plan into four distinct scenarios, differentiated by the jumping directions as follows: 
\begin{equation}
\begin{cases}
m=1,2,l = 3,4,& \theta_{f3}<\theta_{tg}\leq\theta_{f1} \\ 
m=3,1,l = 4,2,& \theta_{f3}<\theta_{tg}\leq\theta_{f4} \\ 
m=2,1,l = 4,3,& \theta_{f4}<\theta_{tg}\leq\theta_{f2} \\ 
m=1,3,l = 2,4,&{\text{otherwise.}} 
\end{cases}
\label{eqn:jumping_case}
\end{equation} 
where $\bm \theta_{tg}$ $\in $[0,\:360]$^\circ$ is the foot angle as shown in Fig. \ref{fig:omini_jump_plane}. In each scenario, the planned forces are strategically decomposed across the robot's feet and guarantee that the torques acting on the robot's CoM $\tau_c$ are exclusively perpendicular to the jump plane. 
\subsection{Optimization Variables}\label{sec:opt_variables}
In this section, we outline our approach for planning jumps within the $J$-$Z$ plane. Drawing upon insights from previous research in \cite{yue2023evolutionary}, we transform this challenge into an optimization problem. First, we plan the GRFs ($\bm u = [u_{J1},u_{J2},u_{z1},u_{z2}]$) in the take-off phase through a polynomial curve about time.
\begin{subequations}
\begin{align}
\bm {u}&=\left\{
\begin{array}{rcl}
\bm a_1 t+a_0 & & {{t} \in [0,\:{t}_1]}\\
\gamma(\bm b_2t^2+ \bm b_1t+\bm b_0) & & {{t} \in \left[{t}_1,\:{t}_2 \right]},\gamma=[0,1]\\
0 & & {{t} \in \left [{t}_2,\:{t}_3\right ]}
\end{array} \right. \label{coefficient_front}\\
\Lambda&=[\bm a_0^{{T}} ,\bm a_1^{{T}}, \bm b_0^{{T}}, \bm b_1^{{T}}, \bm b_2^{{T}}]^{{T}} \in \mathbb{R}^M
\label{coefficient_d},
\end{align}
\label{eqn:poly_coeff}
\end{subequations}

\noindent where $\Lambda$ represents the polynomial coefficients set of the polynomial curve. $t_1$ and $t_2$ represent the end time of take-off phases, and the $t_3$ represents the flight phase. When the take-off phase is segmented into two distinct steps, the parameter $\gamma$ is set to 1. During step 1 ($t< t_1$), two of the robot's legs lift from the ground. Subsequently, in the second step ($t< t_2$), the remaining two legs lift off. This parameter configuration facilitates maneuvers such as back-flip or jumping onto a platform. Refer to Fig. \ref{fig:flip_show} for a schematic illustration.
\begin{center}
\vspace{-0.3cm}
\begin{figure}[tb]
\centering
\includegraphics[width=\linewidth]{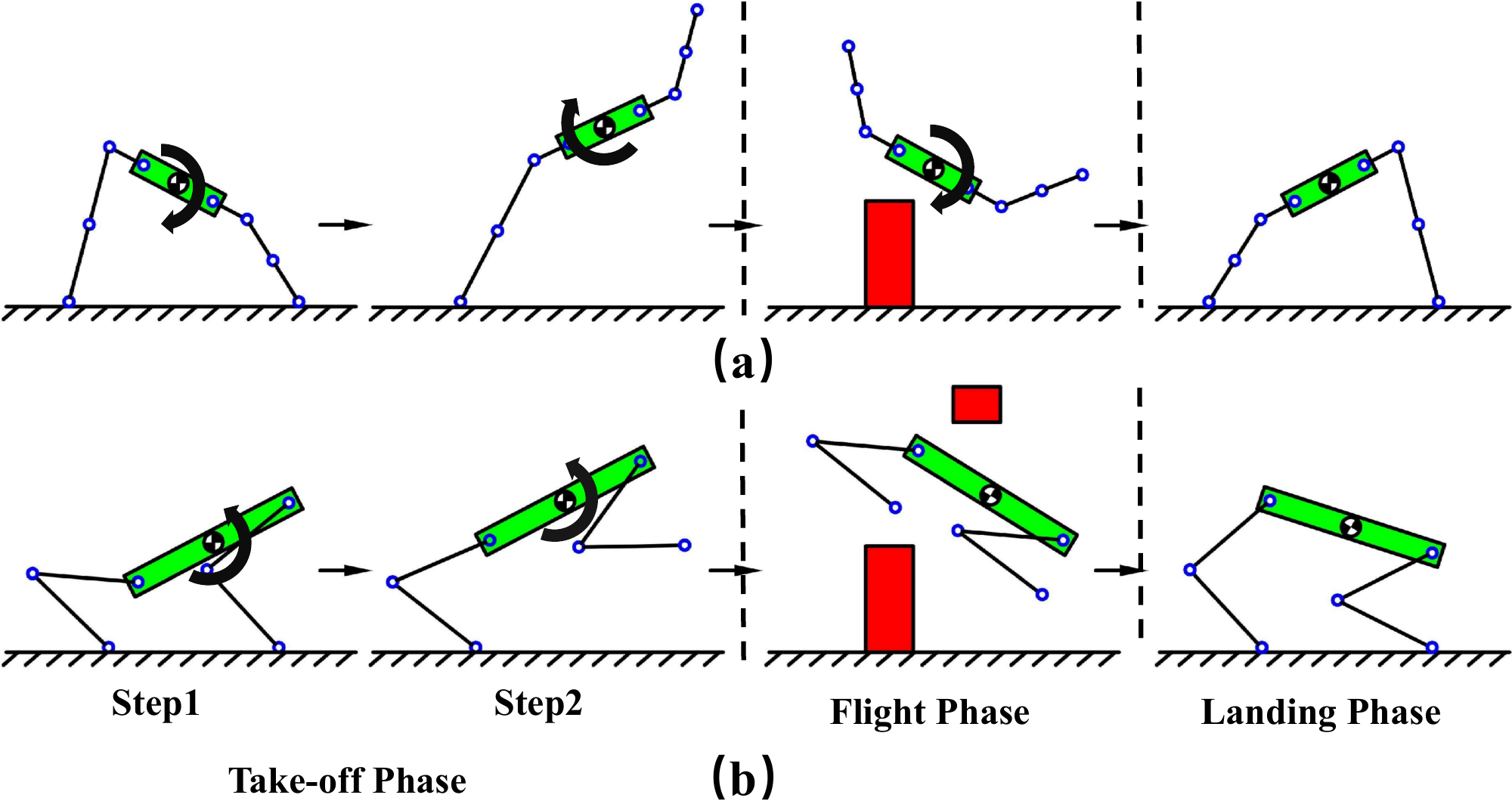}
\caption{Illustration of the two-step take-off phase during a jump. The black curved arrow indicates the direction of rotation. (a) The sequence of foot departures for a left flip. (b) The front foot lifts off first, followed by the back foot, enabling obstacle crossing.}
\label{fig:flip_show}
\end{figure}
\end{center}
Through the SRB dynamics model, the traditional method of directly optimizing polynomial coefficients is converted to optimizing CoM trajectory, different jumping phase times, and GRFs. Considering our omnidirectional jumping motion is constrained to the $J$-$Z$ plane, the reduced-order robot state can be represented as $\bm s_\Omega(t) = [x_{Jc},z_{Jc},\theta_{Jc}]$, where $[x_{Jc},z_{Jc}]$ are the CoM position in the ${J}$ coordinate system and $\theta_{Jc}$ is the rotation angle in $J$-$Z$ plane. By combining  (\ref{coefficient_front}) and SRB dynamics in Sec. \ref{sec:srb_model}, the analytical expression of ${\bm s_\Omega(t)}$ can be solved. 
Consequently, by specifying a series of  $\bm s_\Omega(t)$, the corresponding $\Lambda$ can be obtained. we choose robot states under three time points ($[\frac{t_1}{2},t_2,t_3]$). 
Then, the optimization variables can be given by: 
\begin{eqnarray}
\bm{S}_{\textrm{opt}}:=[\bm{s}_\Omega(\frac{t_1}{2}),\bm{s}_\Omega(t_2),\bm{s}_\Omega(t_3),t_{1},t_{2},t_3]^T \in \mathbb{R}^{12}, \label{d_opt}
\end{eqnarray}

\subsection{Kinodynamic 
Constraints and C-Space}\label{sec:kino_c_space}

\begin{figure*}[ht]
\centering
\includegraphics[width=\linewidth]{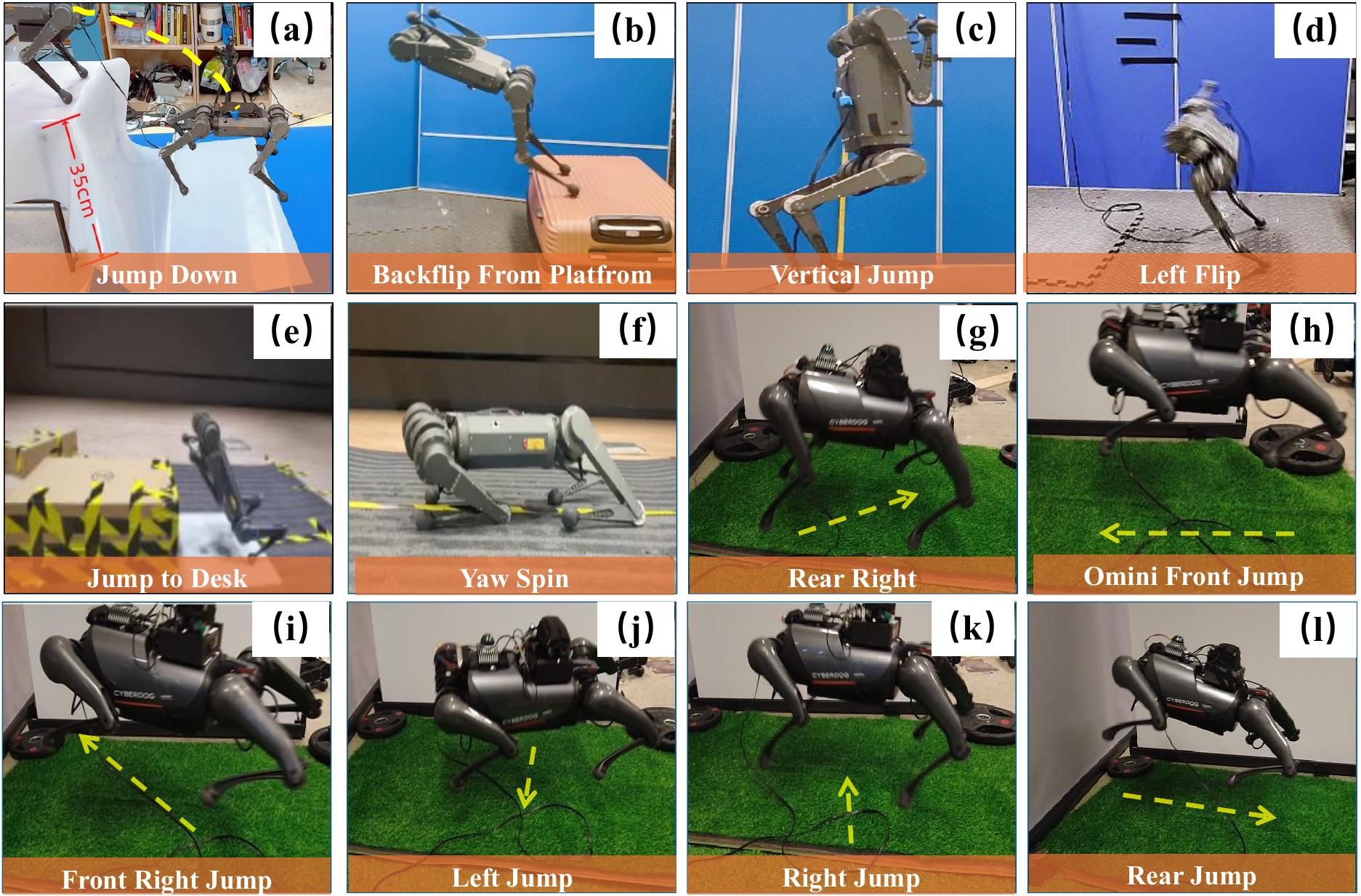}
\caption{
Jumping maneuvers executed using the proposed framework: (a) Jumping down from a 35cm high platform; (b) Backflip from a desk; (c) Vertical jump with both legs; (d) Left somersault; (e) Jumping onto a 30cm high platform; (f) Yaw spin; (g) Omnidirectional rear-right jump; (h) Forward jump; (i) Front-right jump; (j) Left jump; (k) Right jump; (l) Rear jump. ( Details in Video  \href{https://youtu.be/1YGi2pMNIdI}{Omni-Jumping} and \href{https://youtu.be/PQldIYHTprs}{Agile Motions} }
\label{fig:show_experim}
\end{figure*}
In addressing the jumping optimization problem within the constraints of the robot's hardware and environmental limitations, we integrate a series of constraints as follows, prioritized manually for effective resolution:
\begin{itemize}
\item Contact Force: \quad $\bm f_{iz} > f_{zmin}$ . 
\item Friction Cone: \quad $|\bm{f}_{J,i}/\bm{f}_{z,i}|<\mu$ .
\item Joint Angle: \quad $\bm q_{max}>\bm{q}_{ij}>\bm q_{min}$ . 
\item Joint Velocity: \quad $|\bm{\dot{q}}_{ij}|<\bm \dot{q}_{max}$.
\item Joint Torque: \quad $|\bm{\tau}_{ij}|<\bm \tau_{max}$.
\item Joint Position: \quad $\bm{z}_{ij}>\bm z_{min},j\neq2$.
\end{itemize}
Where $j = 0, 1, 2$ and $i = 1,2,3,4$ represent the joint and leg numbers, respectively. The variables $\bm{q}_{ij}$, $\bm{\dot{q}}_{ij}$, and $\bm{\tau}_{ij}$ denote joint angles, joint velocities, and joint torques, respectively. The ground reaction forces (GRFs) on each foot are represented by $\bm{f}_{J, i}$ and $\bm{f}_{z,i}$, as illustrated in Fig.~\ref{fig:omini_jump_plane}. Additionally, $\bm{z}_{ij}$ denotes the joint position in the $z$-direction.

The minimum GRF in the $z$-direction, denoted as $f_{z,\text{min}}$, is set to 1 Nm to reflect the principle that the ground can only exert a pushing force on the robot. The friction coefficient, represented by $\mu$ and set to 0.7, aims to prevent slippage by ensuring adequate grip. Mechanical constraints on joint movements are enforced by setting $q_{\text{min}}$ and $q_{\text{max}}$ for the knee joints to $10^\circ$ and $170^\circ$, respectively, while other joints are not subject to these specific limits. Actuator capabilities are also considered, with the maximum torque ($\tau_{\text{max}}$) and maximum joint velocity ($\dot{q}_{\text{max}}$) for the hip joints set to 24Nm and 300 RPM, respectively, whereas knee joints are constrained to 36Nm and 193 RPM, respectively. To safeguard against the risk of joint-ground collisions during jumps, the minimum permissible joint elevation, $z_{\text{min}}$, is specified as 0.05m.

To further improve the search efficiency of the optimization algorithm, we introduce the joint angle and position constraints to generate the C-space like \cite{yue2023evolutionary}, which makes this optimization problem lie in a much smaller searching region.
Inspired by Ding's work in \cite{ding2020kinodynamic}, we search for $\bm s_\Omega (\bm t_{\textrm{opt}})$ and $\bm t_{\textrm{opt}}$ of $\bm S_{\textrm{opt}}$ in two independent spaces, configuration space (C-space) $\bm {\Omega}_C \subset  \mathbb{R}^{3}$ and time-space (T-space) $\bm {\Omega}_T \subset \mathbb{R}^{3}$, respectively. The definition of $\bm {\Omega}_C$ and $\bm {\Omega}_T$ are as follows:
\begin{eqnarray}
\begin{aligned}
\bm {\Omega}_C:= \{\bm S_\Omega \in \mathbb{R}^{3} \ \arrowvert \ \bm q_{\textrm{min}}<\bm{q}(\bm {S_\Omega})<\bm q_{\textrm{max}}, \\
\bm{z}_{ij}>\bm z_{min},j\neq2 \},\\
\bm {\Omega}_T:= \{t_{1}, t_{2},t_{3}\in \mathbb{R} \ \arrowvert \ 0.1<t_{1}, t_{2}<0.5,t_{2}<0.3 \},
\end{aligned}
\end{eqnarray}
where $\bm {\Omega}_C$ is a set of robot's configurations $\bm s_\Omega$ in different jumping tasks, which satisfies joint angle and position constraints. $\bm {\Omega}_T$ is the time range of take-off and flight phases.
\subsection{Prioritization Fitness Function}\label{sec:p_fit_func}
In addressing the kino-dynamic constraints within the jump optimization problem, we carefully design the penalty function, as shown in
 ~(\ref{optimization_formula}). This approach involves converting the inequality constraints into penalty functions proportionate to their degree of kino-dynamic constraint violation, which are then integrated into the fitness function. The penalty coefficient, denoted as $W_n$, is assigned a value of 1 if a constraint is violated, which means ${\sigma}_n >0$. Conversely, if the constraint is met, 
  $W_n$ is set to 0. For example, if $ \bm f_{iz} < \bm f_{zmin}$ then the contact force constraint is violated and ${\sigma}_n = \bm f_{zmin} - \bm f_{iz}$. Such integration enables the algorithm to iteratively converge towards a trajectory that not only adheres to all imposed constraints but also minimizes energy consumption like \cite{back1997handbook}. To enhance the efficacy of our optimization process, we introduce a prioritization mechanism within our fitness function. This mechanism differentiates the impact of penalty values based on the significance of each constraint, adhering to a hierarchy of predetermined priorities. For instance, we assign a higher priority to the contact force constraint, reflected by setting the cost $W_{n}(10^{n+3}+10^{n}\sigma_{n})$ index $n=15$, whereas the joint torque constraint, deemed lower priority, is assigned $n=6$.
\subsection{Pre-motion Library}
This section aims to establish an offline library about a set of the optimal solutions under the constraints set ($\bm S_{\textrm{opt}}$) called the Pre-motion Library to accelerate online DE optimization. 

The central idea for building the Pre-motion Library is to generate a set of $\bm S_{\textrm{opt}}$ offline with $M$-dimensional optimal solutions ($\bm{S}_{\textrm{pre}}$). According to the fixed step size ($\sim0.05\ m$), $\bm{s}_{\Omega}$ is uniformly divided to obtain the target state of the robot. Then, the obtained target robot states are input into the proposed online omnidirectional jumping framework. The corresponding $\bm{S}_{\textrm{pre}}$ are saved and collected to form the Pre-motion Library. The library comprises 3867 $\bm{S}_{\textrm{pre}}$ representing various kinds of jumping motion (e.g., front/rear/side, backflip, side flip, two/four-leg jump). When re-planning is required, pre-calculated optimization variables can be shared with new evolution as a warm start.
Furthermore, in addition to the $S$, each record also includes the desired position and orientation of the robot CoM. There is also geometric information about the obstacles, such as the window-shaped obstacles, represented by the rectangles above and below.
\begin{figure}[!h]
  \centering
 \subfigure[]{
    \label{CaRe} 
    \includegraphics[width=0.9\linewidth]{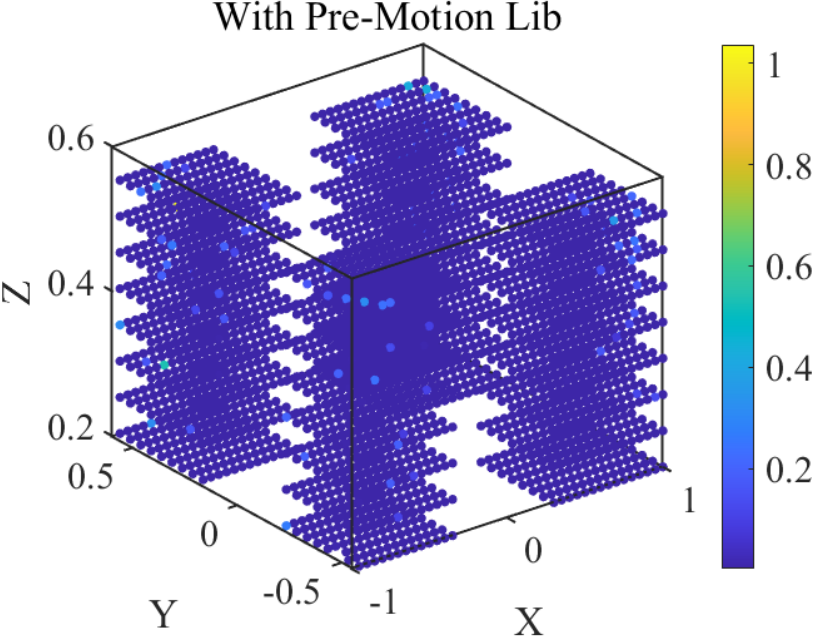}}
  \subfigure[]{
    \label{CaPe} 
    \includegraphics[width=0.9\linewidth]{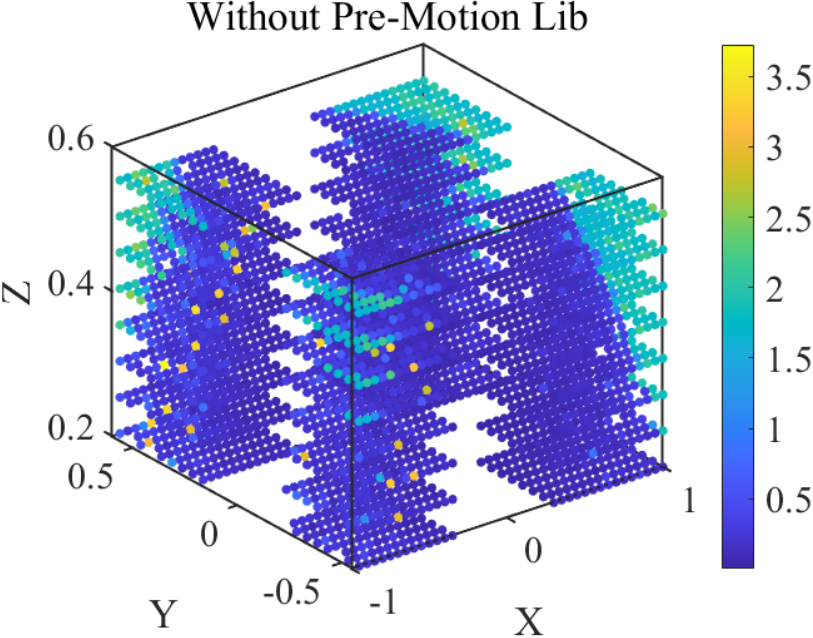}}
  \caption{
The figure compares optimization times when utilizing the pre-motion library as an initial guess versus not using it within the omnidirectional jumping framework proposed in this article. Based on a resolution of 0.05 m, 3,807 jumping target points were sampled. The X, Y, and Z axes represent the three-dimensional coordinates of the target positions for the jump optimizer, defined in the body coordinate system. The color bar indicates the time required (units: second) to optimize each target point. Subfigure (a) uses the pre-motion library as the initial guess, while (b) represents direct optimization with a random initial trajectory.}
  \label{fig:omini_time_compar} 
  \vspace{-0.3cm}
\end{figure}
An index file (Yaml-file) maintains all of the CoM's pre-motion trajectories. In addition, the Pre-motion file is $\sim0.34$ Megabytes in size (MB). It will be loaded into memory at the start of the controller's engine.
The desired trajectory from the library is based on the minimum Euclidean distance with a specified threshold (0.05 m); the input is the high-level information $\bm s_{\Omega}=\{\bm p_{tg},\bm O_k\}$. 
\begin{table}[t]
\center
\setlength{\tabcolsep}{1mm}{
\caption{Average solve times for forward/backward jump motion type}\label{tab:avg_time}
\begin{tabular}{c|c}
Jumping Framework & Forward/Backward  \\ \hline \hline 
Proposed With Lib. & 0.088 (s) \\
Proposed Without Lib.  &  0.42 (s)\\ \hline 
~\cite{song2022optimal} &65 (s)\\ 
~\cite{yue2023evolutionary} &2.19 (s)\\ \hline \hline 
\end{tabular}}

\vspace{-0.3cm}
\end{table}
\subsection{Trajectory Planning with DE}\label{sec:tpde}
  \begin{figure}[htbp]
\centering
\includegraphics[width=\linewidth]{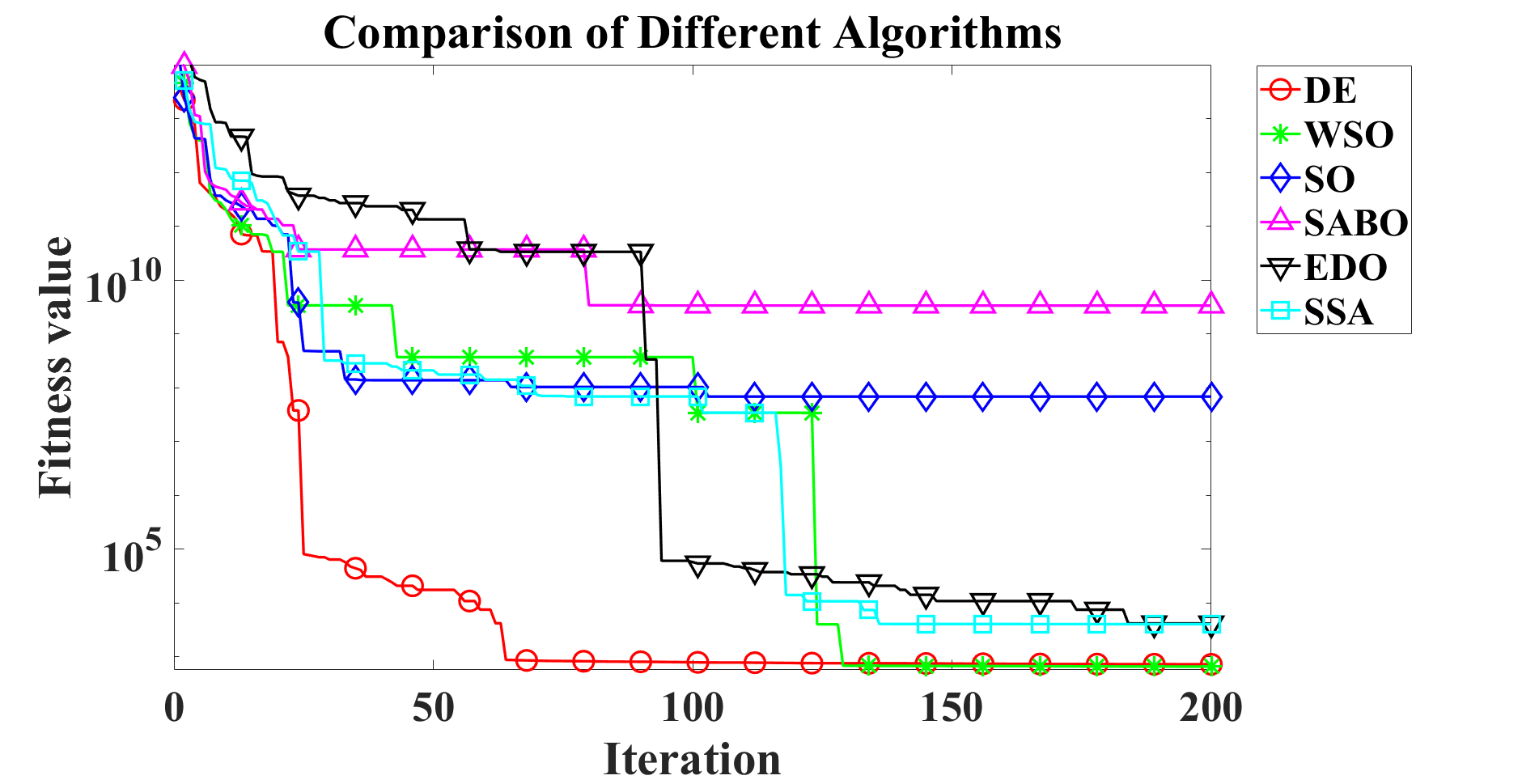}
\caption{
The figure illustrates the comparison between the DE algorithm and other evolutionary algorithms. The green line marked with asterisks represents the War Strategy Optimization (WSO) algorithm in~\cite{WSO_A}. The blue line marked with rhombus represents the Snake Optimizer (SO) in~\cite{SO_A}. The magenta line marked with scalene triangles represents the Subtraction-Average-Based Optimizer (SABO) in~\cite{SABO_A}. The black line marked with infra triangles represents the Exponential Distribution Optimizer (EDO) in~\cite{EDO_A}. The cyan line marked with squares represents the Salp Swarm Algorithm (SSA) in~\cite{SSA_A}.}
\label{fig:de_comparision}
\vspace{-0.5cm}
\end{figure}
After the C-Space, Pre-motion Library, and optimization variables are defined, Online Trajectory Planning with Differential Evolution (OTP-DE) can be introduced. 
Since the jump optimization problem presented in this paper does not require reference motion or the specification of jump times for each stage and complex kino-dynamics constraints, we use an evolutionary algorithm to solve it. Among evolutionary algorithms,  DE is widely favored for its stability and rapid solution speed. In this study, we use the same jump target position and fitness function to compare DE with five other evolutionary algorithms. As shown in Fig. \ref{fig:de_comparision}, DE demonstrates a clear advantage in the number of convergence iterations in our problem.

The OTP-DE utilizes the prioritization fitness function (see  (\ref{optimization_formula})) to search for solutions in C-space. 
The Pre-motion Library to make the online trajectory planning converge faster. Online DE optimization generates $\bm S_{\textrm{opt}}$ by combining high-level information about the environment and ${{\bm S}_{\textrm{pre}}}$ from Pre-motion Library and transmitting it to the robot dynamics model to generate trajectories. 
To enable crossover and mutation operations in the differential evolution (DE) method, the optimization parameter vector is $\bm S$ (see Algorithm. \ref{algo_de}).
We introduce ${\bm \Omega_{C}}$ and ${\bm \Omega_{T}}$ into the search space ${\bm \Omega_{s}}$ to limit the mutation region. Then, in contrast to the conventional DE algorithm's random population initiation, we use Latin Hypercube Sampling (LHS) in~\cite{ayyub1989structural} to produce a more uniform initial population distribution, which improves algorithm iteration convergence speed.  For the objective robot state $\bm{p}_{tg}$ whose Euclidean distance to $\bm{p}_\textit{pre}$ is less than the threshold $r$, it is obtained from the Pre-motion library. Otherwise, it continues to use the LHS for initialization.

The details of the OTP-DE algorithm are shown in Algorithm \ref{algo_de}, where ${\bm p_{tg}} \in {\mathbb{R}^3}$ is the desired position and Euler angular of the CoM, $\bm{p}_{pre}$ is the target point of a trajectory in the Pre-motion library and ${\bm O_k} \in {\mathbb{R}^{12}}$ is the location of obstacles coming from the high-level information in our framework. ${\textrm{Maxgen}}$ and ${NP}$ represent the maximum generation and population numbers, respectively. ${r}$ is the neighborhood radius of ${ {\bm S}_{\textrm{pre}}}$. ${\varepsilon}$ is the fitness value at which the algorithm stops and returns $\bm S_{\textrm{opt}}$. $g$ is the number of the DE generations, $M(\cdot)$, $C(\cdot)$, and $\textrm{LHS}(\cdot)$ present the mutation, crossover, and Latin hypercube sampling functions. ${\bm U_m(g)}$ is the unit vector w.r.t. optimization parameters.

\begin{algorithm}
\caption{OTP-DE Algorithm}\label{algo_de}
\begin{algorithmic}
\State \textbf{Input:} $ \bm p_{tg},\bm s_{\Omega},\bm O_k, {\bm S}_{\textrm{init}},{\textrm{Maxgen}},NP,r,\varepsilon$
\State \textbf{Output:} $\bm S_{\textrm{opt}}$

\State $g\leftarrow 1, \bm k \leftarrow s_{\Omega}=[{\bm p_{tg}},{\bm O_k}]$
\State $\bm {\Omega_s} \leftarrow \{\bm S \ \arrowvert \ \bm S_{1 \sim 9} \in {\bm \Omega_{C}}, \bm S_{10 \sim 12} \in {\bm \Omega_{T}}\}$
\If{$\Arrowvert \bm{p}_{tg} - {\bm p_{pre}} \Arrowvert_2 < r$}
    \State $\bm {S}(g)\leftarrow \mathrm{LHS}({\bm \Omega_{pre}}, NP)$
\Else
    \State $\bm {S}(g)\leftarrow \mathrm{LHS}({\bm \Omega}, NP)$
\EndIf
\While{Fitness($\bm{S}(g) , {\bm k})> {\varepsilon} \ \bm{or}\ g < \rm {Maxgen}$}
\For{$m\leftarrow 1$ to $NP$}
    \State Mutation and Crossover
    \For{$n\leftarrow 1$ to $12$}
    \State $v_{m,n}(g)\leftarrow M(\bm S_{m,n}(g))$;
    \State $u_{m,n}(g)\leftarrow C(\bm S_{m,n}(g),v_{m,n}(g))$
    \EndFor
    \State Selection
    \If{Fitness(${\bm U}_m(g),{\bm k}$)$<$Fitness(${\bm S}_m(g),{\bm k}$)}
    \State ${\bm S}_m(g)\leftarrow {\bm U_m(g)}$;
        \If{Fitness(${\bm S}_m(g),{\bm k}$)$<$Fitness(${\bm S}_{\textrm{opt}},{\bm k}$)}
        \State $\bm S_{\textrm{opt}}\leftarrow {\bm S}_m(g)$
        \EndIf
    \Else
        \State ${\bm S_m(g)}\leftarrow \bm S_m(g)$
    \EndIf
\EndFor
\State $g \leftarrow g+1$
\EndWhile
\end{algorithmic}
\end{algorithm}

\section{Whole-body tracking controller for Take-off Phase}\label{sec:wbc_tracking}
This section employs task-priority Whole-Body Control (WBC) for the take-off phase in the jumping task, as described in \cite{kim2019highly}, to perform full-order dynamics calculations and provide more accurate joint torques for the jumping take-off phase. Accurate torque and joint position calculations are crucial for using a hybrid controller for motor control methods if the actuator motor can not provide exact torque in a real robot. 

This work primarily uses task-priority WBC for joint position compensation and precise joint torques, utilizing Ground Reaction Forces (GRFs) and CoM states from the jumping trajectory generator. The following Quadratic Program (QP) optimization problem is constructed:
\begin{subequations}
\begin{align}
\min _{\boldsymbol{\delta}_{\boldsymbol{q}_f}, \boldsymbol{\delta}_f}& J\left(\boldsymbol{\delta}_{\boldsymbol{q}_f}, \boldsymbol{\delta}_f\right) = \boldsymbol{\delta}_{\ddot{q}_f}^T \mathbf{Q}_1 \boldsymbol{\delta}_{\ddot{q}_f}+\boldsymbol{\delta}_f^T \mathbf{Q}_2 \boldsymbol{\delta}_f \\
\text { s.t. } &\quad \mathbf{S}_f \mathbf{A} \ddot{\boldsymbol{q}_f}+\mathbf{S}_f \mathbf{D}=\mathbf{S}_f \mathbf{J}_c^{T} \boldsymbol{f}_c \label{floating_cons}\\
& \;\;\;\;\ddot{\boldsymbol{q}}_f =\ddot{\boldsymbol{q}}_f^{cmd}+\left[\begin{array}{c}
\boldsymbol{\delta}_{\ddot{q}_f} \\
\mathbf{0}
\end{array}\right] \label{q_1}\\
&\;\;\;\; \boldsymbol{f}_c  =\boldsymbol{f}_{d}+\boldsymbol{\delta}_f \label{q_2}\\
 &\;\;\;\; \boldsymbol{c}_l \leqslant \mathbf{G} \cdot{ }\boldsymbol{f}_c \leqslant \boldsymbol{c}_u \label{friction_cone_cons}
\end{align}
\end{subequations}

Where \(\ddot{\boldsymbol{q}}_f\) is the expected generalized acceleration, and \(\ddot{\boldsymbol{q}}_f^{cmd}\) is the generalized acceleration from the task priority. The slack variables \(\boldsymbol{\delta}_{\ddot{q}_f}\) and \(\boldsymbol{\delta}_f\) from WBC are expected to be minimized. The vector \(\boldsymbol{f}_d\) represents the GRFs from  (\ref{optimization_formula}), while \(\boldsymbol{f}_c\) is the coupled force from  (\ref{q_2}). \(\mathbf{Q}_1\) and \(\mathbf{Q}_2\) are positive definite weight matrices. Equation (\ref{friction_cone_cons}) defines the friction cone matrix. Equation (\ref{floating_cons}) describes the floating base dynamics, where \(\boldsymbol{S}_f\) is a selection matrix that considers only the first six rows. After obtaining the local optimal GRFs from the QP and joint acceleration from the task priority, the motor torque can be calculated as follows:
\begin{equation}
\left[\begin{array}{c}
\mathbf{0}_{6 \times 1} \\
\boldsymbol{\tau}
\end{array}\right]=\mathbf{A} (\boldsymbol{q}) \ddot{\boldsymbol{q}}+\mathbf{D}(\boldsymbol{q}, \dot{\boldsymbol{q}}) - \mathbf{J}_c^{T} \boldsymbol{f}_c \label{tau_cmd}
\end{equation}
The more accurate motor torque commands from the whole-body controller come from:
\begin{equation}
    \bm \tau^{J}_{cmd} = \bm{\tau_{ff}} + \bm k^J_p(\bm q^r - \bm q) + \bm k^J_d({\bm{\dot{q}}^r - \bm \dot{\bm q}})
\end{equation}
where $\bm{\tau_{ff}}$ is feed-forward torque from (\ref{tau_cmd}), $\bm k^J_p$ and $\bm k^J_d$ are the PD gains matrices. $\bm q^r$ and $\bm{\dot{q}}^r$ are the reference motor position and velocity vectors. $\bm q$ and $\bm{\dot{q}}$ are the actual motor position and velocity vectors.
\section{Impedance landing controller For Landing Large Impact}\label{sec:impedance_landing}
 \begin{figure}[htbp]
\centering
\includegraphics[width=0.6\linewidth]{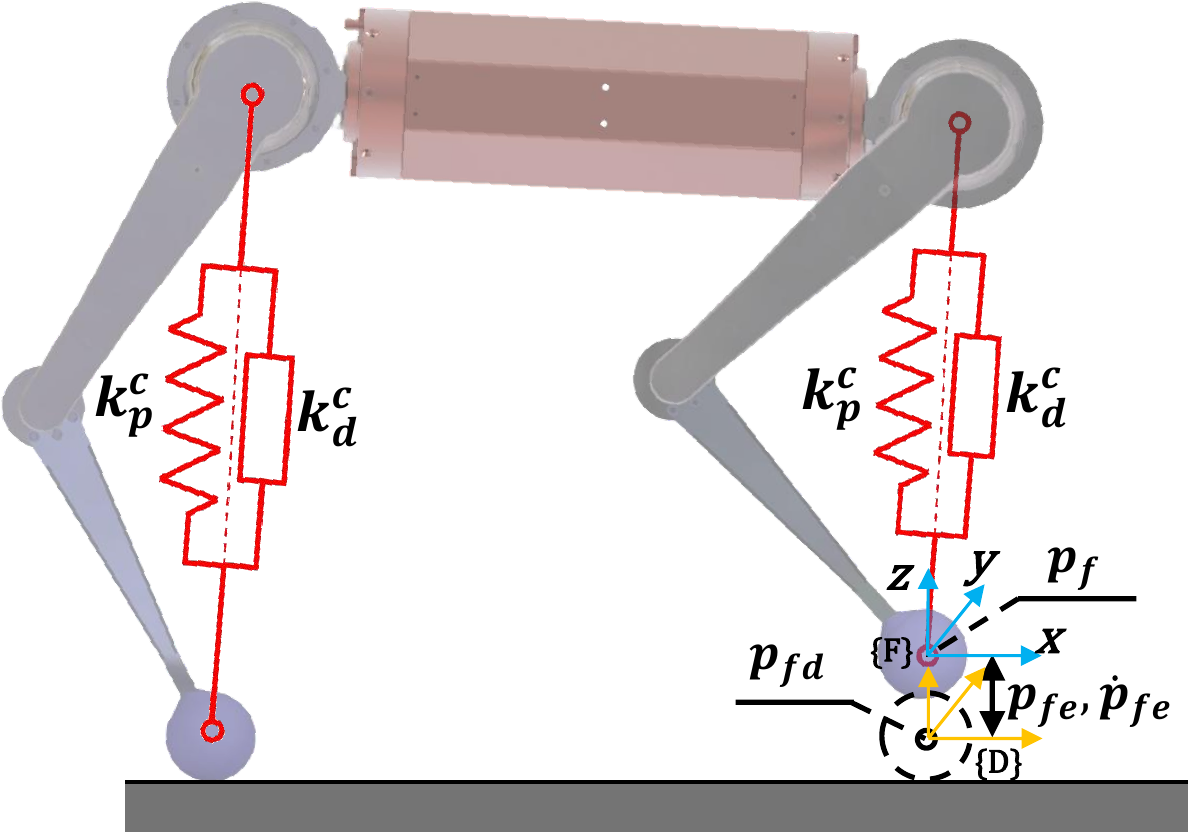}
\caption{
The figure illustrates the impedance control diagram in Cartesian space during the flight and landing phases. The red color represents the virtual spring damping. In this diagram, $\bm p_f$ denotes the current feet coordinate within the world system, $\bm p_{fd}$ is the target feet position in the world system, and $p_{fe}$ is the difference between the $\bm p_f$ and $\bm p_{fd}$.
}
\label{fig:im_Control}
\end{figure}

Impedance control, a control methodology integrating force and position considerations, has exhibited significant efficacy in robotic arms like \cite{caccavale1999six, lin2021unified}. Building upon biological insights, like \cite{hyun2014high} and \cite{lee2014dynamics} pioneered the application of proprioceptive impedance control exclusively within leg mechanisms, achieving remarkable milestones such as high-speed trot-running with the MIT Cheetah. Drawing inspiration from these advancements, we have incorporated impedance control into our omnidirectional jumping framework to manage impact responses during jumping and landing maneuvers effectively. Notably, impedance control can be categorized into joint space and Cartesian space representations, with this study opting for the latter to mitigate potential shortcomings associated with dynamic system control at singular positions. 
To characterize GRFs, foot position
of each foot, the dynamic equations and joint state quantities can be expressed by Lagrangian as shown in~\cite{lee2014dynamics}:
\begin{eqnarray}
\begin{aligned}
&  \mathbf{B} (\bm q) \ddot{ \bm q}+  \bm n( \bm q, \dot{ \bm q})= \mathbf{S}_f (\bm q)  \bm \tau+\sum_i  \mathbf{J}_i( \bm q)^T  \bm f_{i} \label{imd:A}\\
&  \bm n(\bm q, \dot{\bm q}) = \mathbf{C}(\bm q, \dot{\bm q})\dot{\bm q}+ \mathbf{G}(\bm q)
\end{aligned}
\end{eqnarray}
where $\mathbf{B}(\bm q)$, $\mathbf{C}(\bm q, \dot{\bm q})$, $\mathbf{G}(\bm q)$ and $\mathbf{S}(\bm q)$ are the inertial matrices, Coriolis and centrifugal terms, gravitational terms, and the select input matrix, respectively. And $\bm\tau$ is the actuator torques, $ \mathbf{J}_i(\bm q)=\frac{\partial \bm p_{fi}(\bm q)}{\partial \bm q}$ is contact Jacobian w.r.t inertial frame, $\bm p_{fi}$ is the contact foot position w.r.t inertial frame, $\bm f_i$ is the contact forces.
According to the \cite{Siciliano_}, we have the general mechanical impedance equation as follows:
\begin{equation}
\begin{aligned}
&\mathbf{M}_d \ddot{\widetilde{\boldsymbol{x}}}+\mathbf{K}_D \dot{\widetilde{\boldsymbol{x}}}+\mathbf{K}_P \widetilde{\boldsymbol{x}}=\mathbf{M}_d \mathbf{B}^{-1}(\boldsymbol{q}) \boldsymbol{f}\\
&\mathbf{B}(\boldsymbol{q})=\mathbf{J}^{-T}(\boldsymbol{q}) \mathbf{B}(\boldsymbol{q}) \mathbf{J}^{-1}(\boldsymbol{q})
\end{aligned}
\end{equation} where $\mathbf M(d)$ is a positive definite diagonal matrix, $\widetilde{\boldsymbol{x}}=\bm p_{fd_i}-\bm p_{fi}$. Let frame $\{F\}$ and frame $\{D\}$ denote the contact foot frame and desired frame (see Fig. \ref{fig:im_Control}), respectively. Then, we can get the homogeneous transformation matrices from frame $\{F\}$ to frame $\{D\}$. 
\begin{equation}
\begin{aligned}
&\mathbf{T}_f^d=\left[\begin{array}{cc}
\mathbf{R}_f^d & \boldsymbol{t}_{d, f}^d \\
\mathbf{0}^T & 1
\end{array}\right]\\
&\mathbf{R}_f^d=\mathbf{R}_d^T \mathbf{R}_f,
&\boldsymbol{t}_{d, f}^d=\mathbf{R}_d^T\left(\boldsymbol{t}_f-\boldsymbol{t}_d\right). 
\label{jocaon_q1}
\end{aligned}
\end{equation}
where $\mathbf{R}_f^d$, $\mathbf{R}_d$, $\mathbf{R}_f$, $\boldsymbol{t}_{d, f}^d$, $\boldsymbol{t}_f$ and $\boldsymbol{t}_d$ are rotation matrix from feet to desired feet position, feet desired rotation matrix, feet rotation matrix, translation from frame $\{F\}$ to frame $\{D\}$, current feet position, and feet desired translation vector.

We assume the coordinate $\bm p_{fd}$ is time-varying, and then the relationship between foot velocity in Cartesian space and joint velocity is shown as follows in ~\cite{Siciliano_}:

\begin{equation}
\begin{aligned}
&\dot{\widetilde{\boldsymbol{x}}}=-\mathbf{J}(\boldsymbol{q}, \widetilde{\boldsymbol{x}}) \dot{\boldsymbol{q}}+\boldsymbol{b}\left(\widetilde{\boldsymbol{x}}, \mathbf{R}_d, \dot{\boldsymbol{t}}_d, \boldsymbol{\omega}_d\right)\\
& \boldsymbol{b}\left(\widetilde{\boldsymbol{x}}, \mathbf{R}_d, \dot{\boldsymbol{t}}_d, \boldsymbol{\omega}_d\right)=\left[\begin{array}{c}
\mathbf{R}_d^T \dot{\boldsymbol{t}}_d+\mathbf{S}\left(\omega_d^d\right) \boldsymbol{t}_{d, e}^d \\
\mathbf{T}^{-1}\left(\phi_{d, e}\right) \boldsymbol{\omega}_d^d
\end{array}\right]
\label{jocaon_q}
\end{aligned}
\end{equation}
where if we consider that the sole of the quadrupedal robot is in point contact, there is no orientation. The legs cannot stick or slip when in contact with the ground.
Then, computing the derivate of time about  \ref{jocaon_q}, we have 
\begin{equation}
\ddot{\widetilde{\boldsymbol{x}}}=-\mathbf{J} \ddot{\boldsymbol{q}}-\dot{\mathbf{J}}\dot{\boldsymbol{q}}=0
\end{equation}
then, for simplicity, omit the dependence of the functions on their arguments and consider the case where a quadruped robot has no force sensor at its feet and only uses proprioceptive sensors. We obtain a linear impedance equation that is independent of configuration. 
\begin{equation}
\mathbf{K}^C_D \dot{\tilde{\boldsymbol{x}}}+\mathbf{K}^C_P \widetilde{\boldsymbol{x}}\approx \boldsymbol{h}_{ext}^d
\end{equation}
where the $\mathbf{K}^C_D$, and $\mathbf{K}^C_P$ is environment-related active impedance matrices (see Tab. \ref{hyper-parameters}), $\boldsymbol{h}_{ext}^d$ is the linear impedance force vector. To compensate for the absence of a contact force sensor, the feedforward technology is used to calculate the expected GRFs, and the formula is changed to the following (see Fig. \ref{fig:im_Control}):
\begin{equation}
\begin{aligned}
\bm \tau_{ff} &=-\mathbf{S}_f\mathbf{J}^T\bm f_{ff} \\
\bm \tau_{cmd} &= \bm{\tau_{ff}} + \mathbf{K}^C_{P}(\bm p_{fd} - \bm p_f) + \mathbf{K}^C_{D}({\bm{\dot{p}}_{fd} - \bm \dot{\bm p}_f}) 
\end{aligned}
\end{equation}
Here, $\bm f_{ff}$ is the reference force given half of the gravity.
\begin{table}[t]
\scriptsize
\vspace{0.3cm}
\caption{HYPERPARAMETERS\label{hyper-parameters}}
\centering
\renewcommand\arraystretch{1.3}
\scalebox{0.9}{
\begin{tabular}{l|l|l}
\hline
Parameters & Symbol &Values \\
\hline
\multicolumn{3}{c}{\textbf{Impedance Landing}}\\
\hline
Cartesian Proportional Gains & $\mathbf{K}^C_P$ & $diag(500,500,350)$\\
Cartesian Derivative Gains & $\mathbf{K}^C_D$  & $diag(14,14,14)$\\
\hline
\multicolumn{3}{c}{\textbf{DE Optimization Paramters}}\\
\hline

DE-scaling Factor (Without Initial Guess) & $ F$ & 0.85 \\
Crossover Probability (Without Initial Guess)& $CR$ & 0.75 \\
Mutation Probability (Without Initial Guess) & $Pmu$& 0.05\\
DE-scaling Factor (With Initial Guess) & $F$ & 0.9 \\
Crossover Probability (With Initial Guess)& $CR$ & 0.95 \\
Mutation Probability (With Initial Guess)& $ Pmu$& 0.5\\
Number of parameters & $Npar$& 6\\
Population size & $Npop$ &20\\
Number of iterations & $MaxIter$& 200 \\
Friction Coefficient & $\mu$ & 0.7 \\
\hline
\end{tabular}
}
\end{table}

\begin{table}[h]
\scriptsize
\center
\caption{Legged Robot Parameter Table}
\begin{tabular}{c|c|c|c}
\hline 
Parameter  & Symbol & Value & Units\\ \hline \hline 
\multicolumn{3}{c}{\textbf{Duplicated Open Source Mini Cheetah}}\\
\hline
Mass & m & 11.4&Kg\\
Inertia &  [$ I_{xx}$,\quad$ I_{yy}$,\quad$ I_{zz}$]& [0.07,\quad0.3,\quad0.34]&$kg.m^2$\\
Leg Length &[$ L_0$,\quad$ L_1$,\quad$ L_2$] & [0.072,\quad0.211,\quad0.2]&m\\ \hline \hline 
\multicolumn{3}{c}{\textbf{Cyberdog 1}}\\
\hline
Mass & m & 14.0&Kg\\
Inertia &  [$ I_{xx}$,\quad$ I_{yy}$,\quad$ I_{zz}$]& [0.08,\quad0.4,\quad0.45]&$kg.m^2$\\
Leg Length &[$ L_0$,\quad$ L_1$,\quad$ L_2$] & [0.107,\quad0.2,\quad0.217]&m\\ \hline \hline 
\multicolumn{3}{c}{\textbf{Humanoid Robot}}\\
\hline
Mass & m & 47.0&Kg\\
Inertia &  [$ I_{xx}$,\quad$ I_{yy}$,\quad$ I_{zz}$]& [11.6,\quad9.9,\quad2.0]&$kg.m^2$\\
Leg Length &[$L_{upper}$,\quad$ L_{lower}$] & [0.366,\quad0.340]&m\\ 

Foot Length &[$L_{toe}$,\quad$ L_{heel}$] & [0.180,\quad0.120]&m\\ \hline \hline 
\end{tabular}
\label{tab:robot_mass_param}
\vspace{-0.5cm}
\end{table}
\section{Reliable Localization On Jumping Task}\label{sec1:reliable_localization}
When a robot performs a jumping task, the intense landing impulses can significantly increase the risk of localization failure, which poses substantial challenges to reliable navigation. To address this issue, we introduce a robust localization recovery mechanism within our navigation framework by proposing a novel coarse-to-fine re-localization method. In the coarse stage, a global Branch and Bound (BnB) search algorithm is employed to ensure robust re-localization and to provide an accurate initial pose for subsequent refinement. In the refinement stage, we formulate a Maximum a Posteriori (MAP) estimation problem that fuses periodic observations from multiple sensors, including Inertial Measurement Units (IMU), motor encoders, and LiDAR, to achieve more precise localization results. This coarse-to-fine approach enhances the resilience of the navigation system against localization failures induced by high-impact landings, thereby improving the overall reliability of legged robot navigation during dynamic maneuvers.
\subsection{Coarse Re-localization Using Global BnB Search}
To ensure robustness, we employ a Branch and Bound (BnB) algorithm for global search. However, the computational complexity of BnB grows exponentially with the dimension of the solution space. To enhance the algorithm's efficiency, we first reduce the dimensionality of the solution space.

In the general point cloud registration problem, the solution space is six-dimensional, encompassing 3D rotation and translation. In our re-localization problem, we reduce this to three dimensions by leveraging observations from auxiliary sensors. Specifically, accelerometer data provides an accurate gravity direction, constraining the rotation to a single degree of freedom and allowing us to perform a one-dimensional rotation search. For the translation vector \(\bm{t} = [x, y, z]^{\mathrm{T}}\), the \(z\)-component ($z^*$) is determined by calculating the foot end position using encoder measurements, thereby limiting the translation search to two dimensions.

Consequently, the solution space for our re-localization problem is defined as $\mathcal{B} = \{\bm{b} | \bm{b} \in \mathbb{R}^3, \bm{b} = [\theta, x, y]^{\mathrm{T}}, |\theta| < u^b_{\theta}, |x| < u^b_{x}, |y|< u^b_{y}\}$, where $u^b_{\theta}$, $u^b_{x}$ and $u^b_{y}$ are the upper of $\theta, x, y$ and we choose $u^b_{x} = \pi, u^b_{x} = u^b_{y} = 2 \;(m)$ in this paper according to the 2-times largest jumping region of our robots. 
Within this reduced solution space, the re-localization problem is formulated as a maximum consistency problem, enabling efficient and robust pose estimation despite the high-dimensional challenges typically associated with BnB algorithms.
\begin{equation}
    \begin{split}
    \mathrm{E}(\bm{b}) = \sum \limits_{i=0}^{|\mathrm{P}|} \max \limits_{j \in \lbrack0
    , |\mathrm{N}| \rbrack} \lfloor | &\bm{n}_j^{\mathrm{T}}(\mathbf{R}(\theta) \cdot \bm{p}_i+\bm{t}) + d_j| \leq \varepsilon_r \rfloor\\
    \bm{t} = \begin{bmatrix}
        x \\ y \\ z^*
    \end{bmatrix}\;\;\;\;\;
    \mathbf{R}(\theta) &= \begin{bmatrix}
        \cos \theta & -\sin\theta & 0\\
        \sin \theta & \cos \theta & 0\\
        0 & 0 & 1
    \end{bmatrix}  \\
    \bm{b}^* &= \arg\max \limits_{\mathbf{b} \in \mathcal{B}} \mathrm{E}(\bm{b}) 
    \end{split}
    \label{objectiveFunctionForrotationSearch} 
\end{equation}
where $\mathrm{N} = \{\bm{n}_j, d_j\}_{j = 0}^{|\mathrm{N}|}$ is the patch set in the pre-built map and $\mathrm{P} = \{\bm{p}\}_{i = 0}^{|\mathrm{P}|}$ is the point set downsampling from current lidar observation. According to the lidar observation noise, we choose $\varepsilon_r = 0.1$ in this paper. Then, we define the upper and lower bound of the above objective function $\mathrm{E}(\bm{b})$. 

\begin{equation}
    \widetilde{\mathrm{E}}(\mathcal{B}_k) = \sum \limits_{i=0}^{|\mathrm{P}|} \max \limits_{j \in \lbrack0
    , |\mathrm{N}| \rbrack} \lfloor | \bm{n}_j^{\mathrm{T}}(\mathbf{R}(\theta^l_{c}) \cdot \bm{p}_i+\bm{t}^l_{c}) + d_j| \leq \varepsilon_r \rfloor
\end{equation}
where $\mathcal{B}_k$ is a subspace of $\mathcal{B}$, $\bm{b}^l_{c} = [
    \theta^l_{c},x^l_{c},y^l_{c}
]^{\mathrm{T}}$ is the center of $\mathcal{B}_k$ and $\bm{t}^l_{c} = [
x^l_{c},y^l_{c},z^*
]^{\mathrm{T}}$. Since 
\begin{eqnarray}
    \widetilde{\mathrm{E}}(\mathcal{B}_k) \leq \arg\max \limits_{\mathbf{b} \in \mathcal{B}_k} \mathrm{E}(\bm{b})
\end{eqnarray}
we can get $\widetilde{\mathrm{E}}(\mathcal{B}_k)$ is the lower bound of objective function in $\mathcal{B}_k$. Next, we will give the objective function's upper bound.
\begin{equation}
    \widehat{\mathrm{E}}(\mathcal{B}_k) = \sum \limits_{i=0}^{|\mathrm{P}|} \max \limits_{j \in \lbrack0
    , |\mathrm{N}| \rbrack} \lfloor | \bm{n}_j^{\mathrm{T}}(\mathbf{R}(\theta^l_c) \cdot \bm{p}_i+\bm{t}^l_c) + d_j| \leq \varepsilon_r + \delta_j \rfloor 
\end{equation}
where 
\begin{equation}
    \delta_j = 2\sin(\frac{u^b_{\theta}}{2}) \Vert \bm{p}_j \Vert+ \sqrt{(u^b_x)^2 + (u^b_y)^2} 
\end{equation}

\begin{proof}
    Since $|\theta| < u^b_{\theta}$, we have $\angle(\mathbf{R}(\theta), \mathbf{R}(\theta^l_c)) < u^b_{\theta}$. Thus, 
    \begin{equation}
        \Vert\mathbf{R}(\theta) \cdot \bm{p}_i - \mathbf{R}(\theta^l_c) \cdot \bm{p}_i \Vert < 2\sin(\frac{u^b_{\theta}}{2}) \Vert \bm{p}_j \Vert
    \end{equation}
    given $|\bm{n}_j^{\mathrm{T}}(\mathbf{R}(\theta) \cdot \bm{p}_i+\bm{t}) + d_j| \leq \varepsilon_r$, we have
    \begin{equation}
    \begin{split}
        |\bm{n}_j^{\mathrm{T}}((\mathbf{R}(\theta) + \mathbf{R}(\theta^l_c) - \mathbf{R}(\theta^l_c)) \cdot \bm{p}_i+\bm{t} + \bm{t}^l_c - \bm{t}^l_c) + d_j| \leq \varepsilon_r
    \end{split}
    \end{equation}
    Since
    \begin{equation}
        \begin{split}
            &|\bm{n}_j^{\mathrm{T}}((\mathbf{R}(\theta) + \mathbf{R}(\theta^l_c) - \mathbf{R}(\theta^l_c)) \cdot \bm{p}_i+\bm{t} + \bm{t}^l_c - \bm{t}^l_c) + d_j| \\
            =&|\underbrace{\bm{n}_j^{\mathrm{T}}(\mathbf{R}(\theta^l_c) \cdot \bm{p}_i+\bm{t}^l_c) + d_j}_{e^l_c} + (\mathbf{R}(\theta) - \mathbf{R}(\theta^l_c))\cdot \bm{p}_i + \bm{t} - \bm{t}^l_c|
            \\
            \geq&
            \; |e^l_c| - \Vert\underbrace{(\mathbf{R}(\theta) - \mathbf{R}(\theta^l_c))\cdot \bm{p}_i \Vert}_{2\sin(\frac{u^b_{\theta}}{2}) \Vert \bm{p}_j \Vert} - \underbrace{\Vert\bm{t} - \bm{t}^l_c\Vert}_{\sqrt{(u^b_x)^2 + (u^b_y)^2} }
        \end{split}
    \end{equation}
    Thus $|\theta| < u^b_{\theta}$, we have $\angle(\mathbf{R}(\theta), \mathbf{R}(\theta^l_c)) < u^b_{\theta}$ is the sufficient condition for 
    \begin{equation}
        |\bm{n}_j^{\mathrm{T}}(\mathbf{R}(\theta^l_c) \cdot \bm{p}_i+\bm{t}^l_c) + d_j| < \epsilon_r + \delta_j
    \end{equation}
    So we can get $\widehat{\mathrm{E}}(\mathcal{B}_k)$ is the upper bound of the objective function in $\mathcal{B}_k$.
\end{proof}
Once the upper and lower bounds of the objective function are determined, a global Branch-and-Bound (BnB) search can be performed to obtain the global optimum of the proposed optimization problem, thereby achieving coarse re-localization.
\subsection{Localization Refinement}
In this section, we will introduce our localization refinement strategy. We choose the time when finishing coarse re-localization as zero point, and the refined pose at this time is $\mathbf{T} \in \mathrm{SE}(3)$. After that, we can get a period of observations of multi-sensors. Given IMU observations $\mathcal{U} = \{\bm{u}^i_k\}_{k = 0}^{T}$, encoder observations $\mathcal{Q} = \{\bm{q}_k\}_{k = 0}^{T}$ and lidar observations $\mathcal{L} = \{\bm{l}_k\}_{k = 0}^{T}$. We first employ an invariant filter from~\cite{hartley2020contact} performing proprioceptive state estimation and obtaining relative transformation $\Delta \mathbf{\widehat{T}}_k$ and its uncertainty.
\begin{eqnarray}
   \Delta \mathbf{T}_k = \Delta \widehat{\mathbf{T}}_k \mathrm{Exp}(\delta \bm{\xi}_k) \;\;\;\;\; \delta \bm{\xi}_k \sim \mathcal{N}(\bm{0}, \; \mathbf{P}_k) 
\end{eqnarray}
Then we form the following MAP problem to refine the coarse pose $\mathbf{T}$
\begin{eqnarray}
    \mathbf{T}^* = \arg \min_{\mathbf{T} \in \mathrm{SE}(3)} \sum_{k = 0}^{\mathrm{T}}\Vert f(\mathbf{T},\; \Delta \widehat{\mathbf{T}}_k, \; \bm{l}_k)\Vert_{\mathbf{\Sigma}_k}
\end{eqnarray}
where $f(\cdot)$ is the lidar observation model using point-to-plane constraint. We can solve the problem of the above least squares using the Lewenberg-Marquardt method to get a more accurate position estimate.
\section{APPLICATIONS}\label{sec:application_task}
Building upon the dynamics model in Section \ref{sec:dyanmics_sec} and the jump optimization framework in Section \ref{sec:jumping_to}, this section further details the specific solution processes for omnidirectional and agile jumping in quadruped robots, as well as $X$-$Z$ plane jumping in humanoid robots. These examples demonstrate the generalizability of the method proposed in this paper.
\subsection{Ominidirectional Jumping for Quadrupedal Robots}\label{sec:application_omini}
Following the discussion in Sec. \ref{sec:jumping_plane_sec}, we introduce a jumping plate $J$-$Z$  to achieve omnidirectional jumps (see Fig. 
\ref{fig:omini_jump_plane}). To clarify this method, we take the target point $p_{tg} = $ (0.5, -0.5, 0.5) m as an example. Under this condition, $\theta_{tg} =$ 225° which corresponds to case $\theta_{f4}<\theta_{tg}\leq\theta_{f2}$ in  (\ref{eqn:jumping_case}) and thus  (\ref{eqn:jumping_plane}) becomes as follows:
\begin{subequations}\label{fj_EXP}
\begin{align}
&\bm f_{J1} = \bm u_{J2}\frac{|\bm p_{J2}-\bm p_{f3}|}{|\bm p_{f1}-\bm p_{f3}|},\bm f_{J2} = \bm u_{J1}\frac{|\bm p_{J1}-\bm p_{f4}|}{|\bm p_{f2}-\bm p_{f4}|} \label{fj_EXP:A}\\
& \bm f_{J3} = \bm u_{J2}-\bm f_{J1} \quad \quad \  ,\bm f_{J4} = \bm u_{J1}-\bm f_{J2} \label{fj_EXP:B}\\
&\bm f_{z1} = \bm u_{z2}\frac{|\bm p_{J2}-\bm p_{f3}|}{|\bm p_{f1}-\bm p_{f3}|} \ , \bm f_{z2} = \bm u_{z1}\frac{|\bm p_{J1}-\bm p_{f4}|}{|\bm p_{f2}-\bm p_{f4}|} \label{fj_EXP:C}\\
& \bm f_{z3} = \bm u_{z2}-\bm f_{z1} \quad \quad \quad ,\bm f_{z4} = \bm u_{z1}-\bm f_{z2} \label{fj_EXP:D}
\end{align}
\end{subequations}

By doing so, the ground reaction forces (GRFs) $\bm u = [\bm f_{i}] \in \mathbb{R}^{12}$ planning is reduced to planning $\bm u = [u_{J1}, u_{J2}, u_{z1}, u_{z2}] \in \mathbb{R}^4$. We use linear equations in  (\ref{eqn:poly_coeff}) to represent $\bm u$ and omit the second step in the take-off phase by setting $\gamma = 0$. This results in eight polynomial coefficients $\Lambda = [\bm a_0, \bm a_1] \in \mathbb{R}^8$ and two time variables [$t_1$, $t_3$] for optimization, totaling 10 dimensions. To improve the algorithm's efficiency, we have reduced the design variables using two methods. First, we artificially specify $u_{J1} = u_{J2}$, introducing two equality constraints on the polynomial coefficients and reducing the design variables by two dimensions. Second, we adopt the approach from Sec. \ref{sec:opt_variables}, mapping polynomial coefficients with unspecified values to the robot state $\bm{s}_\Omega(t) = [x_{Jc}, z_{Jc}, \theta_{Jc}]$, while introducing two equality constraints $x_{Jc}(0) = 0$ and $\theta_{Jc}(0) = 0$. 
Since omnidirectional jumping only has one step at the take-off phase, it is a subset of the take-off phase, completed in two steps. Then, the optimization variables at this stage are as follows:
\begin{eqnarray}
\bm{S}_{\textrm{opt}}:=[\bm{s}_\Omega(\frac{t_1}{2}),\bm{0},\bm{0},t_{1},t_{1},t_3]^T \in \mathbb{R}^{12}, \label{d_opt}
\end{eqnarray}
After removing zero and redundant components, the expression can be simplified into the following compact form:
\begin{eqnarray}
\bm{S}_{\textrm{opt}}:=[\bm{s}_\Omega(\frac{t_1}{2}),t_{1},t_{3}]^T \in \mathbb{R}^{5}, \label{omini_d_opt}
\end{eqnarray}

Then, by using the prioritization fitness function and DE algorithm in Sec. \ref{sec:jumping_to}, optimal jumping motion trajectories can be obtained at different $\bm{p}_{tg}$, verified in Sec. \ref{sec:real_exp_vali} and proves the effectiveness of the method. 

\subsection{Agile Jumping for Quadrupedal Robots}

While our method is primarily applicable to four-legged jumps, it can also be extended to more agile two-legged jumps such as backflips, side flips, jumping over obstacles, and leaping onto high platforms, as demonstrated in our previously published articles like \cite{song2022optimal,yue2023evolutionary}. In this section, we provide a clearer explanation of how the methods in Section \ref{sec:dyanmics_sec} can be applied to these jumps (see Fig. \ref{fig:flip_show}). Unlike the previous omnidirectional jumping, agile jumping motions include Step 2: two-legged jumping during the take-off phase. Therefore, we use two polynomials (setting $\gamma = 1$) to plan the take-off GRF in  (\ref{eqn:poly_coeff}). Using similar concepts to those in omnidirectional jumping, we simplify the agile jumping problem within the jumping plane $J$-$Z$, which coincides with the $X$-$Z$ and $Y$-$Z$ planes for front and side jumps. This simplification allows  (\ref{eqn:jumping_plane}) to be further reduced. Taking the front jump as an example, we have $f_{x1} = f_{x2} = u_{x1}/2$, $f_{x3} = f_{x4} = u_{x2}/2$, $f_{z1} = f_{z2} = u_{z1}/2$, $f_{z3} = f_{z4} = u_{z2}/2$, and during the two-legged jumping phase, $u_{x1} = u_{z1} = 0$. In this way, we can use $\bm u = [u_{x1}, u_{x2}, u_{z1}, u_{z2}]$ instead of $\bm f_{i}$ to generate the jumping motion, reducing the original forty-two polynomial coefficients to fourteen ($\Lambda = [\bm a_0, \bm a_1, \bm b_0, \bm b_1, \bm b_2] \in \mathbb{R}^{14}$).

Further, we use a similar idea to omnidirectional jumping for downscaling and transforming the design variables. First, we also let $u_{x1}$=$u_{x2}$ to reduce $\Lambda$ to 12 dimensions. Then, the robot dynamics model in Sec. \ref{sec:jumping_to} allows these variables to be transformed into 12 robot state variables as shown in \cite{yue2023evolutionary}: $[\bm{s}_\Omega(\frac{t_1}{2}),\bm{s}_\Omega(t_1),\bm{s}_\Omega(t_2),\bm{s}_\Omega(t_3)]$. As $\bm{s}_\Omega(t_3)$ = $\bm p_{tg}$, the target position and angle are given as a high-level command. So, the design variables used in optimization are as follows:
\begin{eqnarray}
\bm{S}_{\textrm{opt}}:=[\bm{s}_\Omega(\frac{t_1}{2}),\bm{s}_\Omega(t_1),\bm{s}_\Omega(t_2),t_{1},t_{2},t_{3}]^T \in \mathbb{R}^{12}, \label{agile_d_opt}
\end{eqnarray}
Then, the optimization problem can be solved following the methods proposed in Sec. \ref{sec:tpde}, and these agile jumping motions are verified in Fig. \ref{fig:offline_flip}.

\subsection{Forward and Backward Jumping for Humanoid Robots}\label{sec:humanoid_extension}

\begin{figure}[htbp]
\centering
\includegraphics[width=3 in]{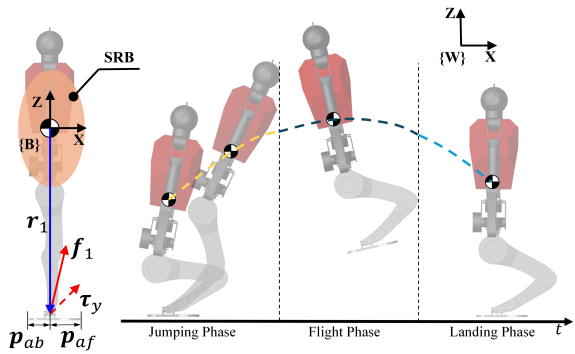}
\caption{Humanoid jumping diagram in the sagittal plane. The humanoid robot's body is simplified to an SRB model, as shown by the yellow ellipsoid and the dashed line representing the COM trajectory. We generate the jumping motion by planning the force $\boldsymbol{f}_1$ and torque $\boldsymbol{\tau}_{y}$ at the ankle joint.}
\label{Humanoid}
\end{figure}

Our method is not limited to quadrupedal robots; it also applies to the jumping of humanoid robots. Taking the forward and backward jumps as an example, unlike quadruped robots with point feet, humanoid robots require accounting for foot torque in force planning, as shown in Fig. \ref{Humanoid}. In this way, the dynamics model in ~(\ref{rcc}) becomes as follows:

\begin{equation}
\frac{\mathrm{d}}{\mathrm{d} t}({^B\mathbf{I} \bm \omega_B}) = \bm{r}_1\times  \boldsymbol{f}_1 + {\tau}_{fy} +{\bm \omega_B}\times^B\mathbf{I} {\bm \omega}_B
\label{humanoid_model}
\end{equation}
where ${\tau}_{fy}$ represents the component of the foot torque perpendicular to the sagittal plane. Then, the $\bm u$ in  (\ref{coefficient_front}) is convert into $[f_{1x},f_{1z},\tau_{fy}]$. By using the linear curve to plan foot force $\boldsymbol{f}_1$ and torque$\tau_{fy}$  and cancel the step 2 in take-off phase ($\gamma$ = 0), there are six polynomial parameters ($\Lambda =[\bm a_0,\bm a_1] \in \mathbb{R}^6$). Together with the times $t_1$ and $t_3$ of the jump and flight phases in  (\ref{coefficient_front}), this optimization problem has eight design variables. Using the same variable transformation step in Sec. \ref{sec:opt_variables}, these physically meaningless polynomial parameters can be transformed into physically meaningful trajectory position and velocity parameters so that the specific expressions for the design variables in  (\ref{new_d_opt}) are:
\begin{eqnarray}
\bm{S}_{\textrm{opt}}:=[\bm{s}_\Omega(0),\bm{s}_\Omega(\frac{t_1}{2}),t_{1},t_{3}]^T \in \mathbb{R}^{8}, \label{new_d_opt}
\end{eqnarray}
where $\bm{s}_\Omega(t) = [x_{Jc},z_{Jc},\theta_{Jc}]$ represents the CoM position and rotation angle of the SRB model in the $\bm{W}$ coordinate system of the sagittal plane as shown in Fig. \ref{Humanoid}.  

Meanwhile, to prevent the robot's feet from flipping over, Zero Moment Point (ZMP) constraints in~\cite{kajita2007zmp} are added to the kino-dynamic constraints in Sec. \ref{sec:kino_c_space}, which is given by:
\begin{eqnarray}
p_{ab} < p_{zmp} = -\frac{\tau_{fy}}{u_z} <p_{af}
\end{eqnarray}
where $p_{zmp}$ is the position of ZMP, $p_{ab}$ and $p_{af}$ are the toe and heel positions, respectively. Then, the jumping trajectory of the robot jumping back and forth can be obtained using the differential evolution algorithm and the Prioritization fitness function in the previous Sec. \ref{sec:jumping_to}, and we have carried out the simulation verification in the later Sec. \ref{sec:human_exp} to prove the feasibility of the method.
\begin{figure*}[htbp]
  \centering
 \subfigure[]{
    \label{four_jumping_outdoor} 
    \includegraphics[width=0.9\linewidth]{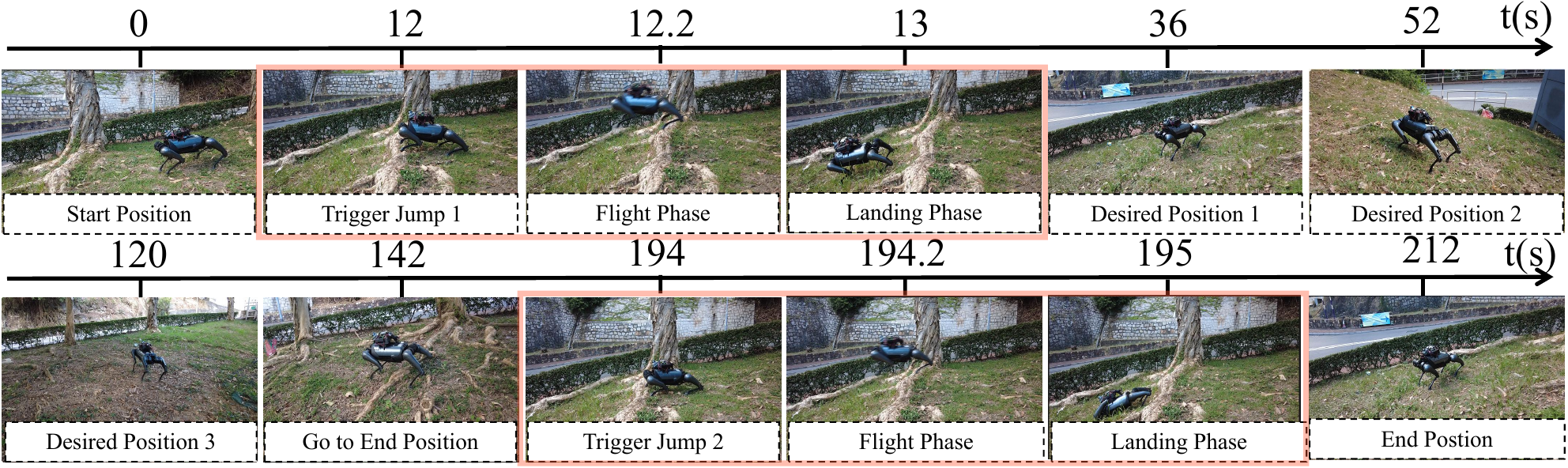}}\\
  \subfigure[]{
    \label{With_grid_map_outdoor} 
    \includegraphics[width=0.9\linewidth]{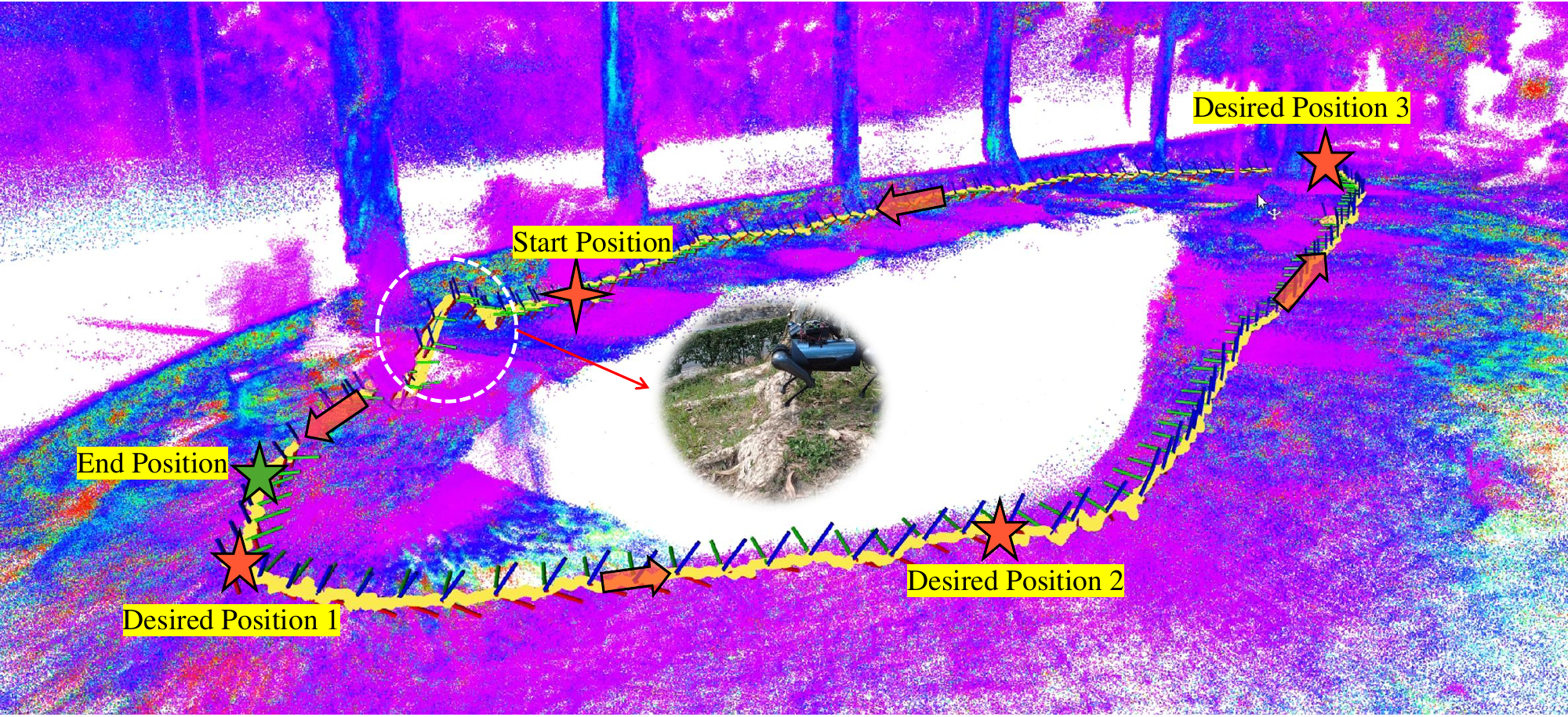}}
  \caption{The figure illustrates initiating a jump during large-scale outdoor navigation. In (a), an outdoor scene is depicted, with the pink box highlighting two distinct jump events. In (b), the constructed map and the corresponding robot trajectory are presented, where the yellow circle marks the robot's jumping trajectory.}
\end{figure*}
\begin{figure*}[htbp]
  \centering
 \subfigure[]{
    \label{four_jumping_two} 
    \includegraphics[width=0.43\linewidth]{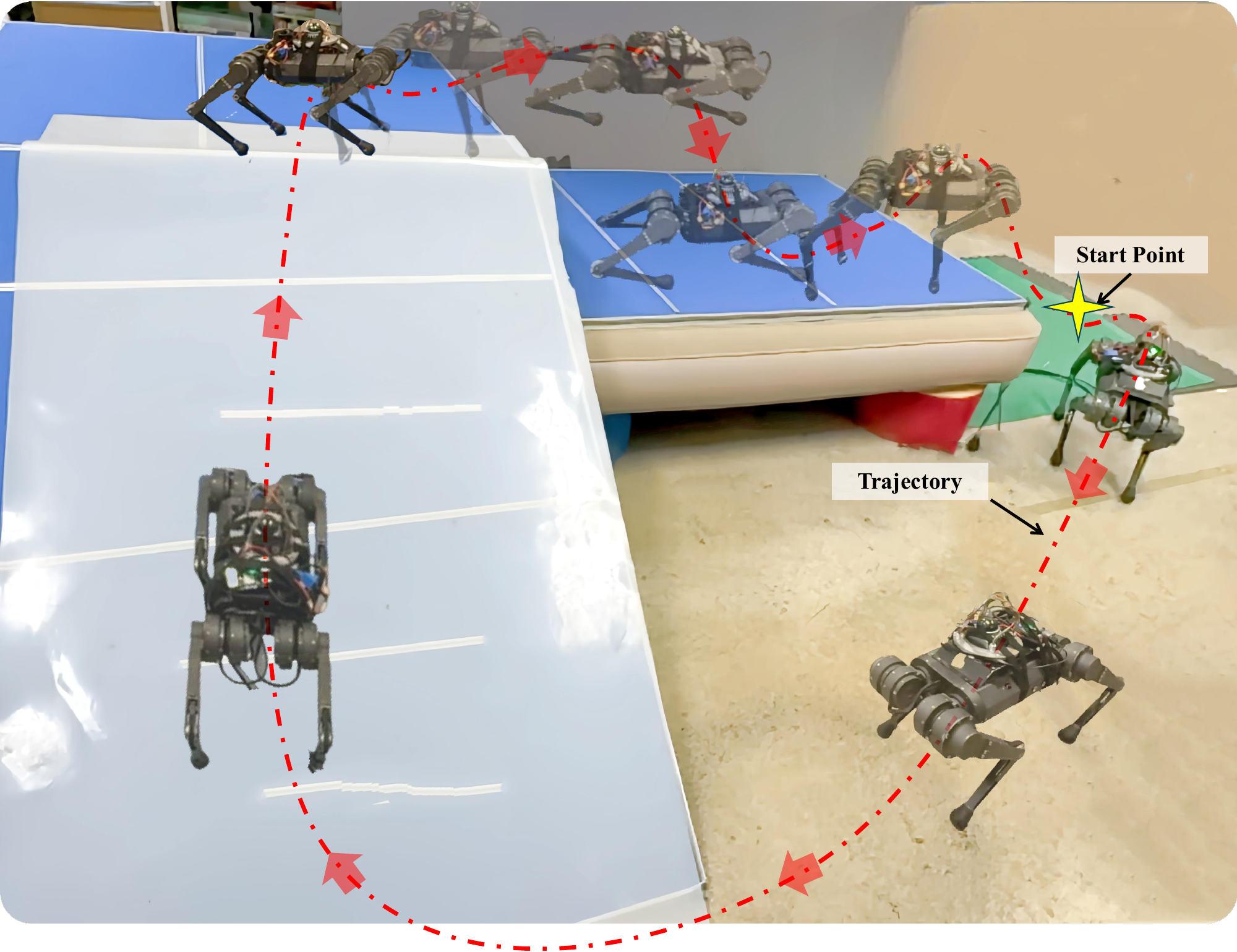}}
  \subfigure[]{
    \label{With_grid_map1} 
    \includegraphics[width=0.47\linewidth]{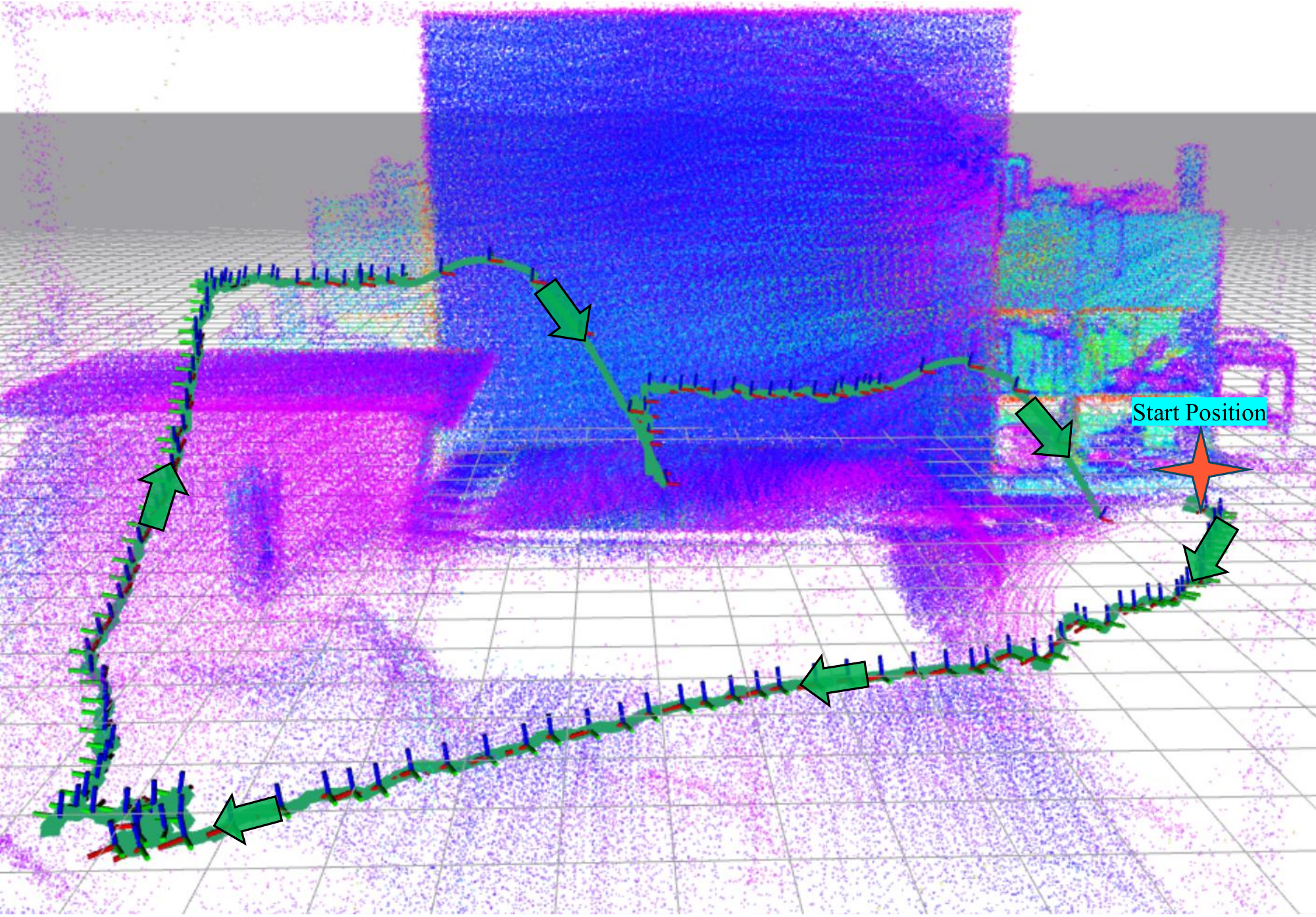}}
  \caption{The image depicts the robot climbing an indoor slope of approximately 35 degrees and performing two consecutive jumps based on varying landing positions on the map, with a maximum drop of 35 cm. In (a), the snapshot illustrates the navigation process and the two jumps. The red dashed arrow line represents the robot's walking direction and approximate running trajectory, while the yellow star marks the starting point. In (b), the point cloud map and the robot's center of mass (CoM) trajectory, generated using the improved mapping and reliable localization method discussed in this article, are shown.}
\end{figure*}

 \begin{figure}[htbp]
\centering
\includegraphics[width=\linewidth]{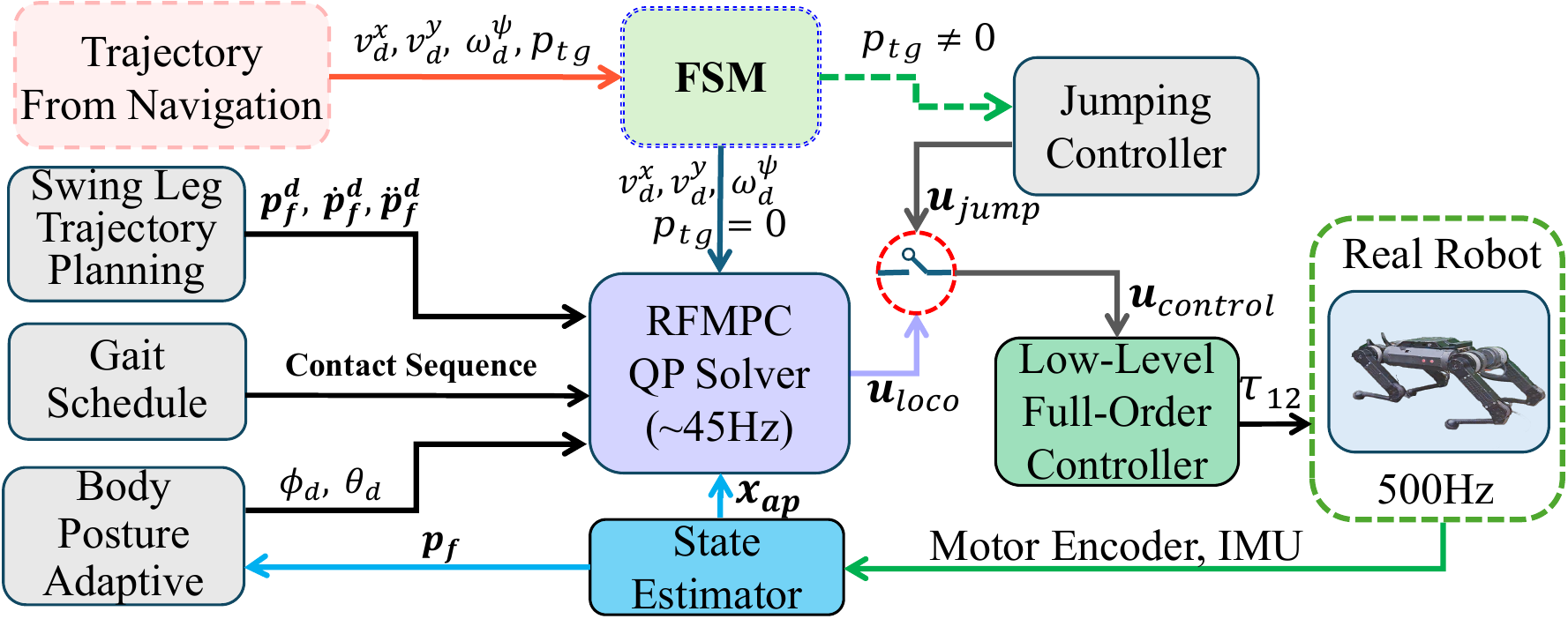}
\caption{The figure provides a detailed flowchart illustrating the use of the MPC+WBC combination as a tracking controller for the navigation trajectory. The navigation trajectory is first sent to the FSM to determine whether a jump is required. If $\bm{p}_{tg} = \bm{0}$, the controller tracks the navigation trajectory, which is passed to the RF-MPC as velocity commands. Otherwise, the jump controller is executed. $\bm{u}_{loco}$ represents the GRFs output by the locomotion controller, $\bm{u}_{jump}$ represents the GRFs output by the jumping controller, and $\bm{u}_{control}$ represents the forces directly sent to the low-level full-order controller by FSM modules. Refer to Fig. \ref{fig:jump_framework} for the jump controller flowchart.}
\label{fig:contro_framework}
\end{figure}
\begin{figure}[t]
  \centering
 \subfigure[]{
    \label{outdoor_suc} 
    \includegraphics[width=0.9\linewidth]{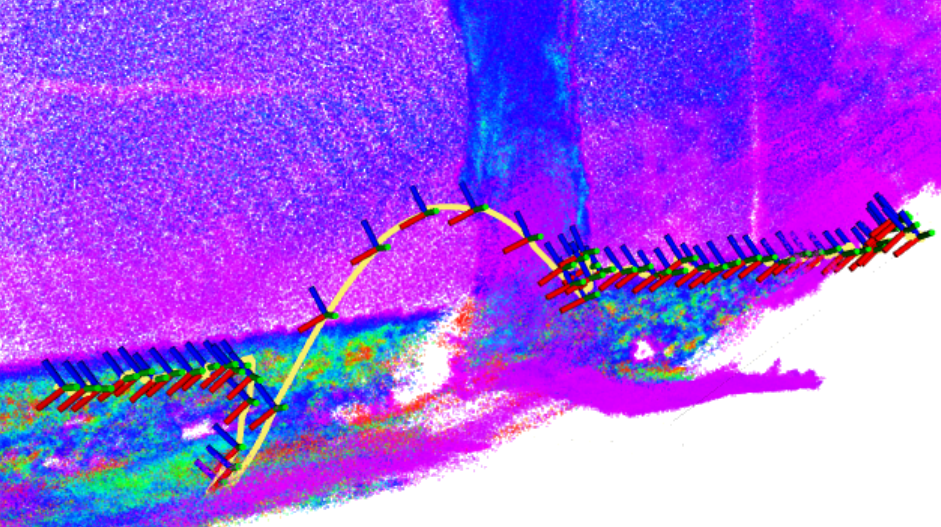}}
  \subfigure[]{
    \label{outdoor_fail} 
    \includegraphics[width=0.9\linewidth]{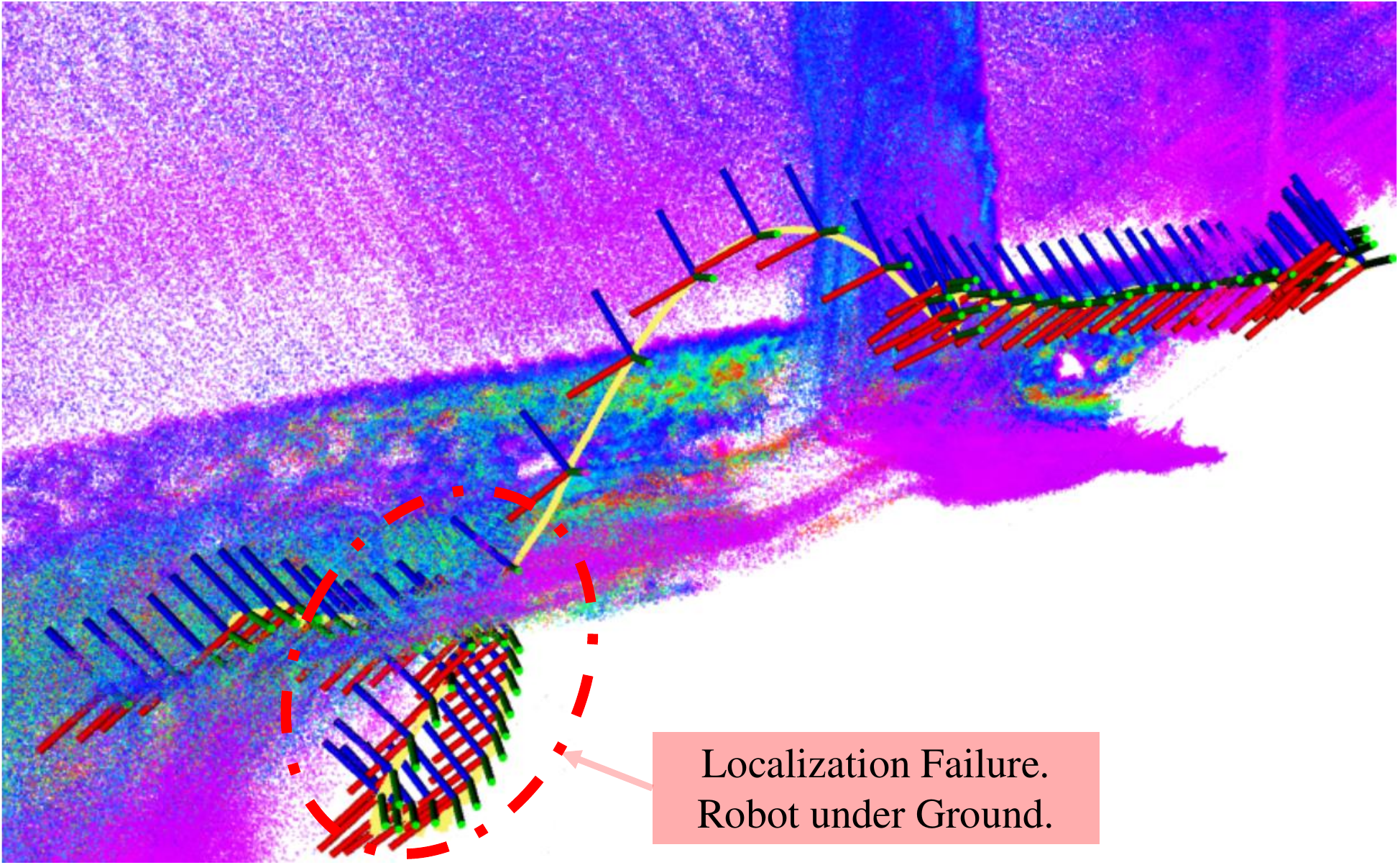}}
  \caption{
The figure illustrates the limitations of baseline\cite{fastlio} localization during two consecutive forward jumps, where landing impacts cause map deformation and localization failure in outdoor environments. In (a), the enhanced relocalization algorithm described in the article preserves map integrity and trajectory accuracy under the same conditions. In contrast, (b) shows that using the baseline mapping method results in ghosting after a single jump, leading to localization failure.}
  \label{fig:good_bad_localization}
\end{figure}
\begin{figure}[t]
  \centering
 \subfigure[]{
    \label{good} 
    \includegraphics[width=0.9\linewidth]{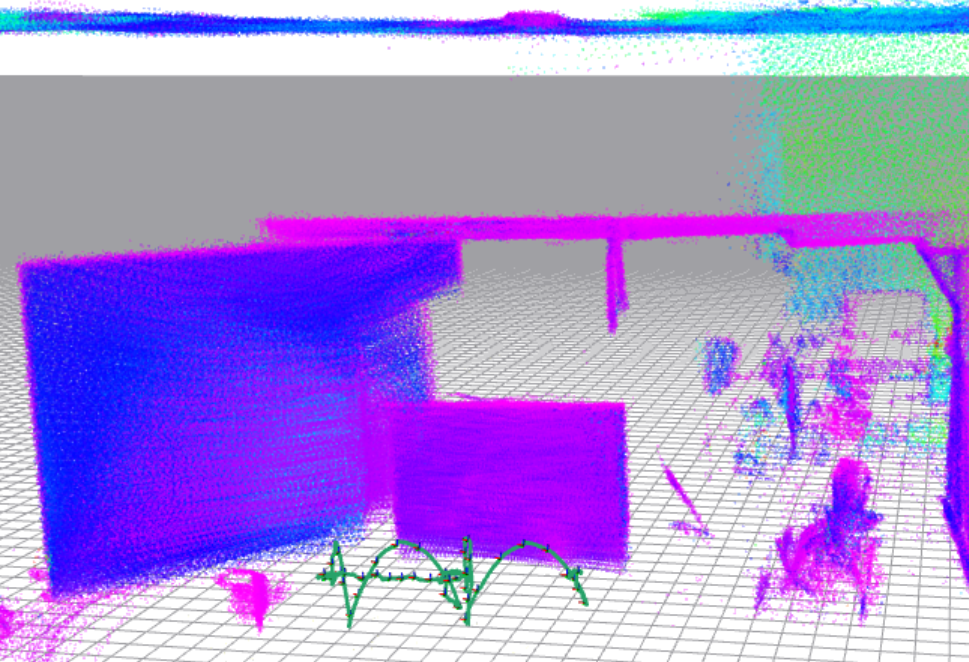}}
  \subfigure[]{
    \label{bad} 
    \includegraphics[width=0.9\linewidth]{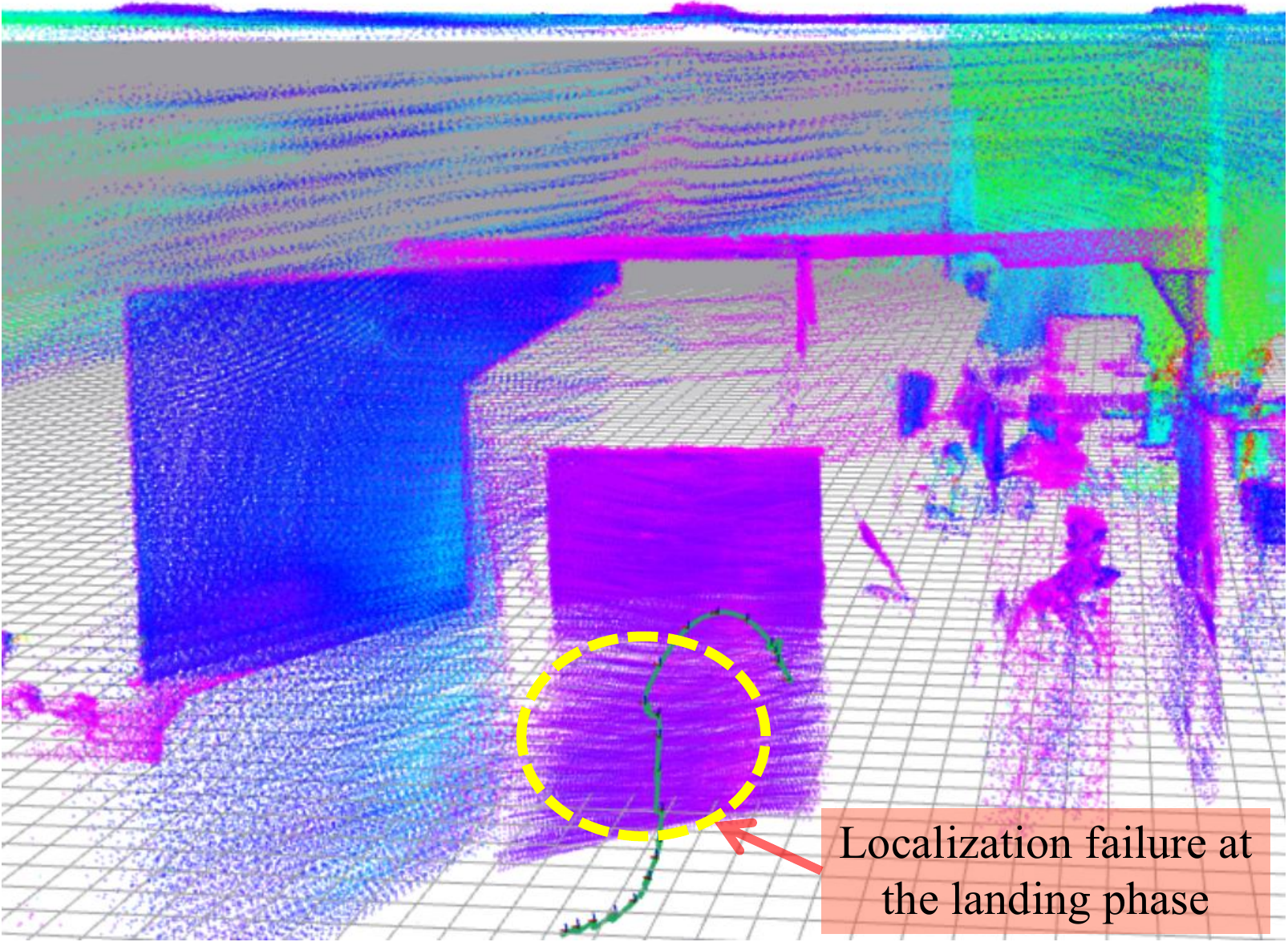}}
  \caption{
The figure shows localization failure in indoor environments. In (a), the enhanced relocalization algorithm maintains map integrity and trajectory accuracy. In contrast, (b) demonstrates localization failure (The robot is positioned below ground level.) using the baseline mapping method after a single jump.}
  \label{fig:good_bad_localization2}
\end{figure}

\begin{figure*}[t]
  \centering
 \subfigure[]{
    \label{four_jumping_two3} 
    \includegraphics[width=0.45\linewidth,height=1.65in]{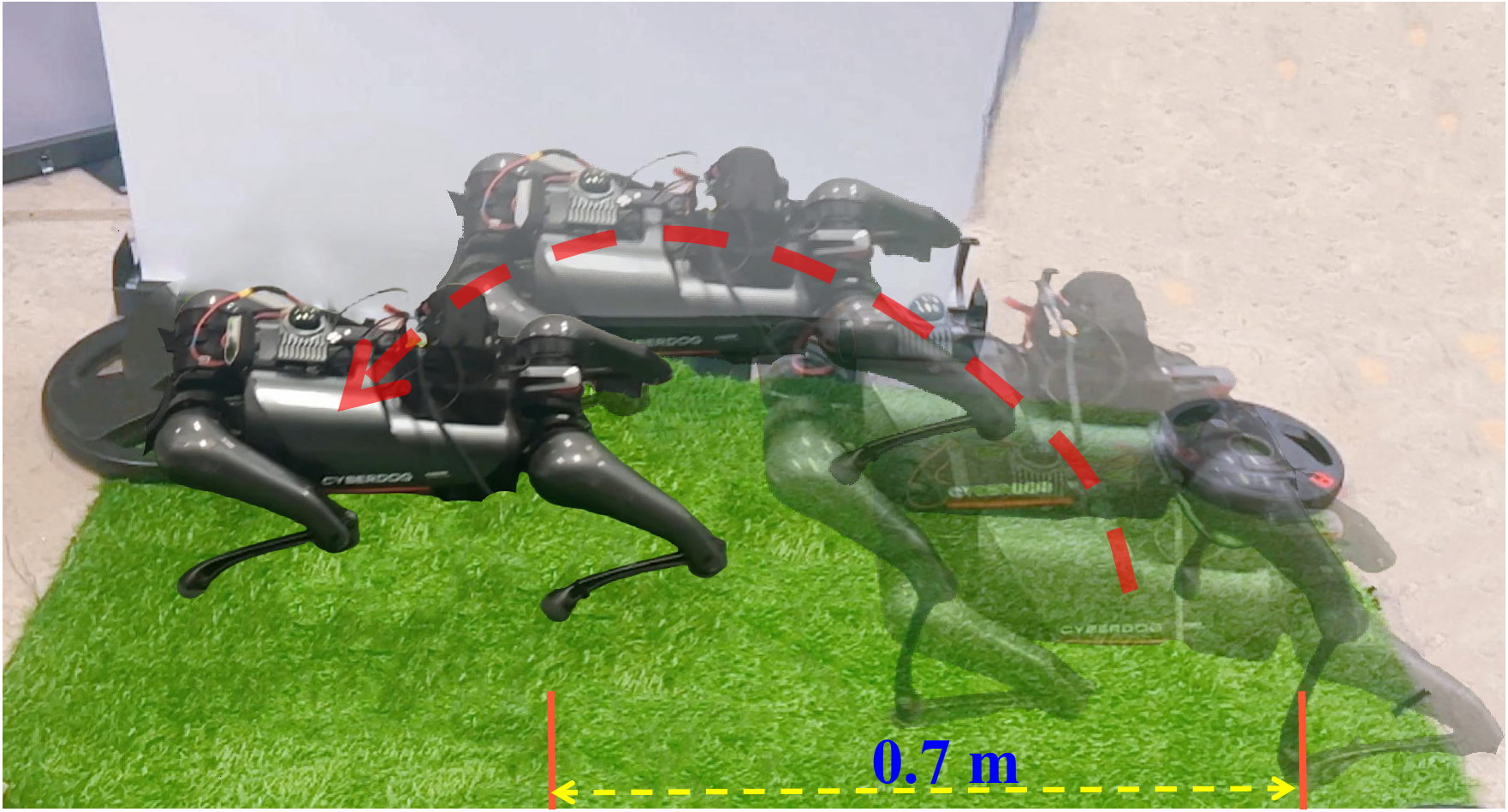}}
  \subfigure[]{
    \includegraphics[width=0.45\linewidth,height=1.65in]{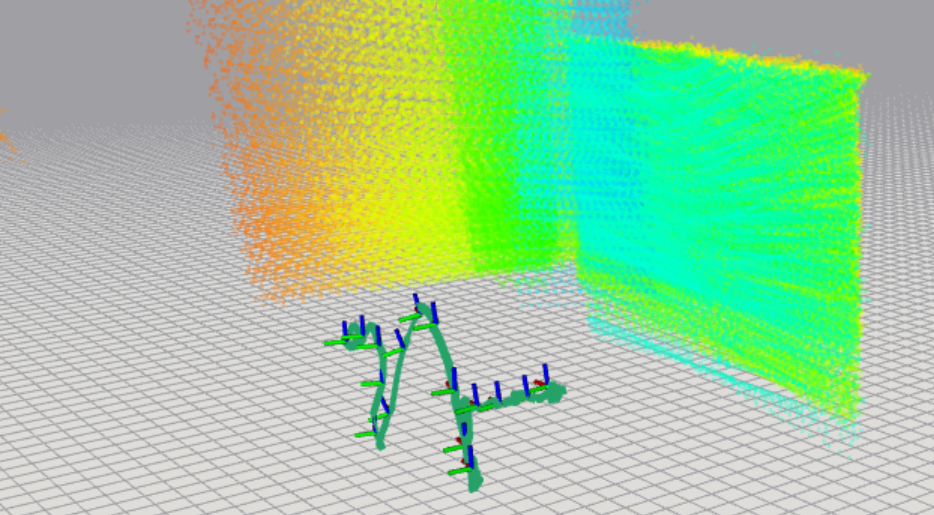}}
  \caption{Trajectory analysis of a quadruped robot executing a forward right jump to the right using the proposed reliable relocation method. (a) illustrates the motion-captured trajectory with a red dotted line. (b) displays the trajectory reconstructed by the localization module.}
  \label{With_grid_map_long} 
\end{figure*}
\begin{figure*}[htbp]
  \centering
 \subfigure[]{
    \label{four_jumping_two2} 
    \includegraphics[width=0.45\linewidth,height=1.6in]{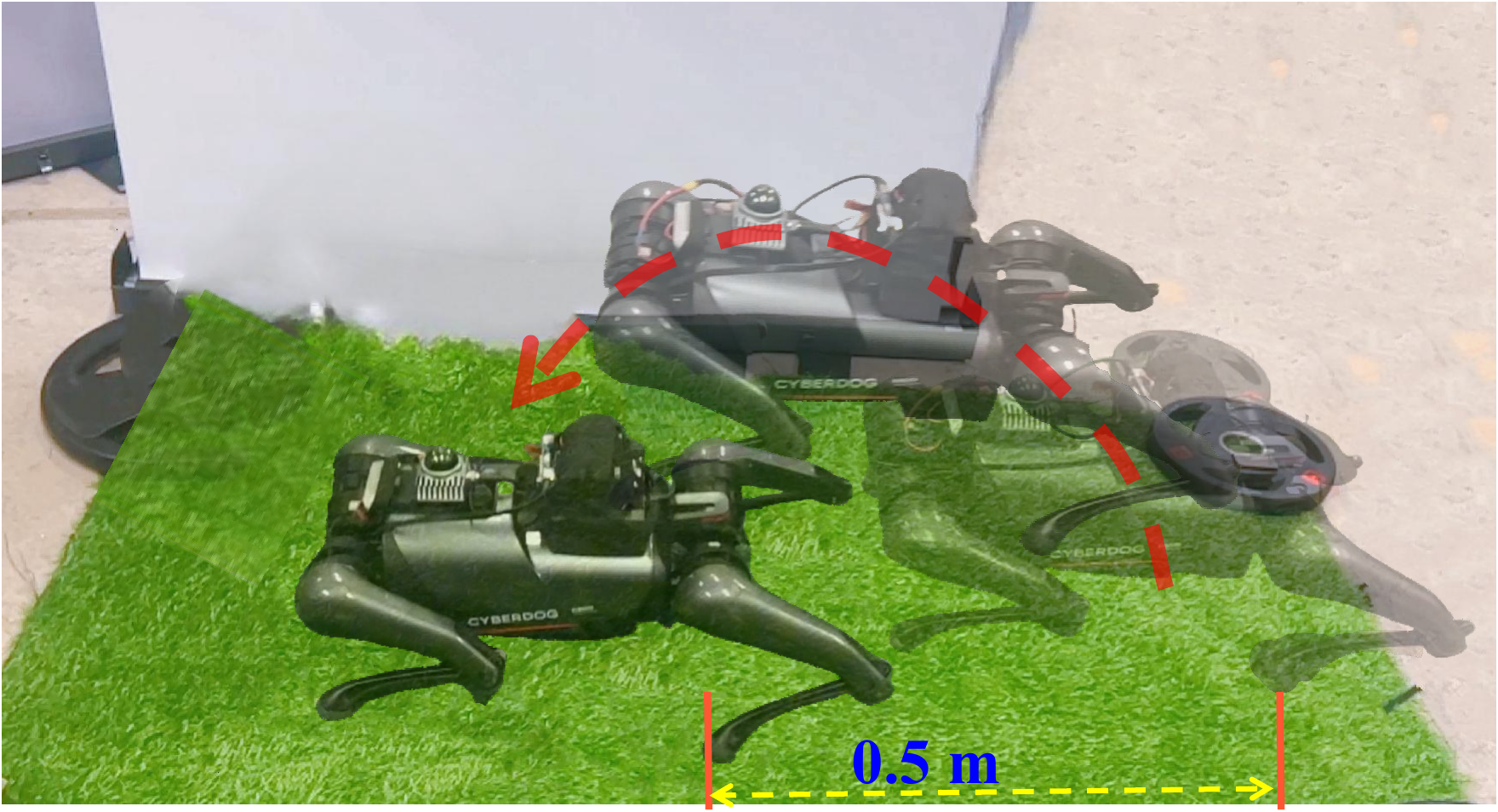}}
  \subfigure[]{
    \label{With_grid_map} 
    \includegraphics[width=0.44\linewidth,height=1.6in]{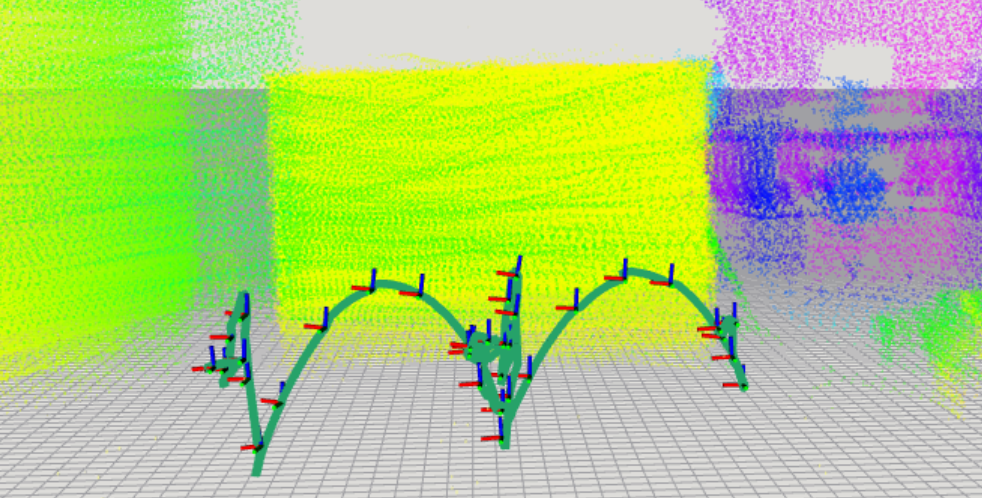}}
  \caption{Analysis of a quadruped robot's forward-right jump: motion snapshots and trajectory. (a) displays the motion-captured trajectory of the forward jump. (b) illustrates the trajectory reconstructed by the relocalization module based on the localization module.}
\end{figure*}
\begin{figure*}[htbp]
  \centering
 \subfigure[]{
    \label{front_right_jump} 
    \includegraphics[width=0.45\linewidth,height=1.6in]{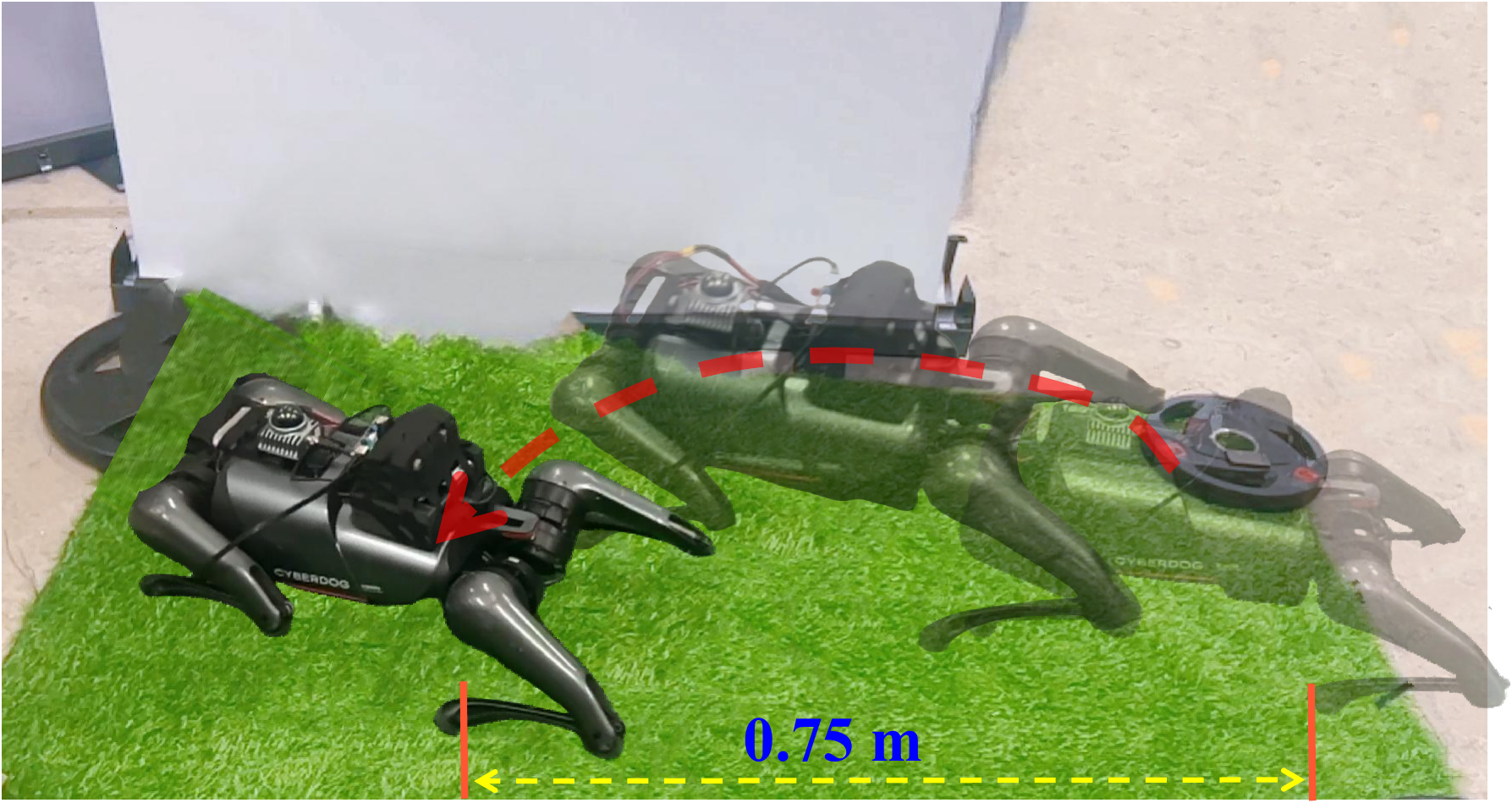}}
  \subfigure[]{
    \label{front_left_lidar} 
    \includegraphics[width=0.45\linewidth,height=1.6in]{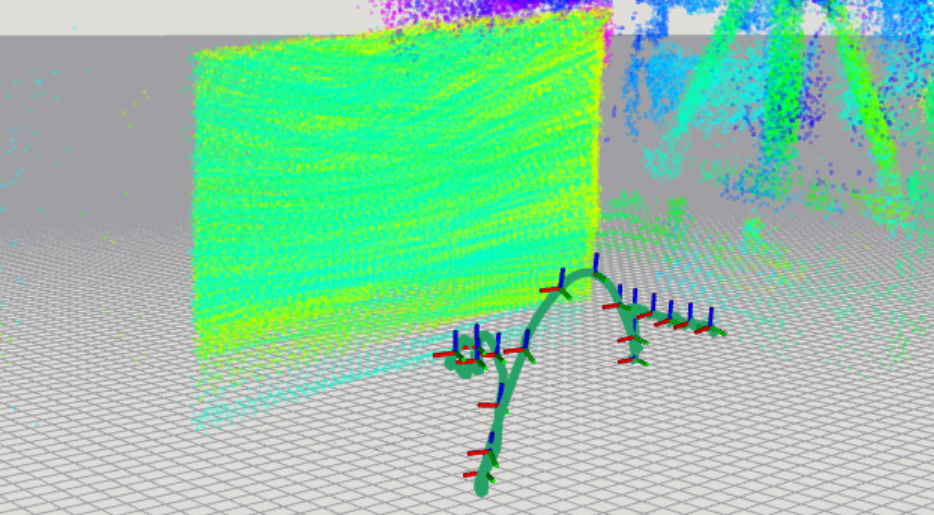}}
  \caption{The figure presents the forward-left jump. (a) shows the video snapshots of the jump, while (b) displays the actual trajectory reconstructed from the localization module.}
\end{figure*}
\section{SIMULATION AND HARDWARE EXPERIMENTS}\label{sec:simulation_and_hardware}

This section focuses on the hardware implementation specifics of the online omnidirectional jumping framework and experiments conducted on a real quadrupedal robot. 
We first outline the submodules of the cascade optimization and locomotion control framework, including the MPC controller, state and slope normal vector estimators, map construction, trajectory generation, and gait scheduler (discussed in Sec. \ref{sec:control_arch}).
In Sec. \ref{sec:conclution_results}, we conducted extensive simulation experiments and performed a statistical analysis of jump optimization time and success rates. We then introduce the robot parameters, Pre-motion computing resources, and the extension to humanoids (Sec. \ref{sec:exp_hard_fea}). In Sec. \ref{sec:real_exp_vali}, we present real-world experiments comparing multiple jump methods, landing controllers, and joint performance under different conditions. Finally, we validate the jump controller through Webots (refer to~\cite{Webots}) simulations, showcasing its versatility (Sec. \ref{sec:humanoid_extension}).

\subsection{Controller Architecture}\label{sec:control_arch}

In addition to developing the jump controller, this paper integrates obstacle terrain jumping with navigation capabilities. Merely generating jump trajectories and tracking Whole-Body Control (WBC) trajectories are insufficient for achieving precise navigation and Center of Mass (CoM) trajectory tracking on uneven terrains. Consequently, several additional sub-modules are necessary to realize comprehensive control. A simplified flowchart of the complete control framework is presented in Fig. \ref{fig:contro_framework}. To enable dynamic walking, this paper employs linear Model Predictive Control (MPC) to predict ground reaction forces (GRFs) during locomotion. 
Furthermore, we adopt the Representation-Free Model Predictive Control (RF-MPC) in~\cite{ding_2020} as the GRF planner for locomotion. Unlike traditional linear MPC approaches that utilize Euler angles for state representation, RF-MPC leverages rotation matrices. Compared to the linear MPC approach presented in ~\cite{kim2019highly}, the RF-MPC formulation incorporates the yaw angle and allows for more precise tracking of desired roll and pitch angles. This enhanced tracking capability significantly improves the navigation process, particularly in maintaining accurate pitch direction alignment when traversing slopes. Effective locomotion also requires integration with a state estimator. Given that the experimental scenarios include slope climbing, a least squares method estimates the slope's normal vector, facilitating necessary posture adjustments. Additionally, since navigation and mapping necessitate offline map construction and trajectory planning between start and target points, sub-modules for mapping and trajectory planning are incorporated into the framework. Each sub-module is briefly described below, with references provided for readers interested in a more in-depth exploration of these components like~\cite{bledt2018cheetah,kim2019highly,ding_2020}.
\subsubsection{RF-MPC for Locomotion}
To enhance slope adaptability in this study, RF-MPC in~\cite{ding_2020} is selected as the trajectory generator for the navigation planner. The primary differences between RF-MPC and Convex MPC in~\cite{kim2019highly} lie in two key aspects. First, RF-MPC employs a rotation matrix instead of Euler angles. This enables the system state equations to account for roll and pitch tracking, which is often neglected. Second, a variation-based linearization in~\cite{wu2015variation} and vectorized linearization method reduce the dimensionality of RF-MPC from 18 to 12 states, achieving a solution frequency of 250 Hz using qpSWFIT in~\cite{pandala2019qpswift}.
\subsubsection{State Estimator and Body Posture Slope Adjust }
The state estimator presented in this paper uses data from the IMU and motor encoders as input and outputs the robot's CoM state, which is crucial for other submodules. The estimator assumes that the robot's foot does not slip during the stance phase. Based on this assumption, a Kalman Filter is designed, leveraging the forward kinematics of the foot and its derivatives. Specifically, the foot's height ($p_{fz}$) is assumed to be zero when in the stance phase. As a result, the observation equation for the Kalman Filter is derived, and the state equation is established from the foot's velocity. The recursive Kalman Filter formula estimates the robot's state in~\cite{bledt2018cheetah} using these equations. Importantly, since the recursive equation includes data from the swing leg and relies on the assumption that the foot does not slip, assigning a high covariance to the swing leg is often necessary to improve state estimation accuracy.

Moreover, when the robot moves across uneven terrains such as a slope, initializing the foot’s early contact point to zero can result in instability of the body. To address this and improve terrain adaptability, particularly on sloped surfaces, this paper employs the slope estimator method introduced in \cite{bledt2018cheetah}. The approach centers on estimating the normal vector of the slope by applying a least squares technique to the position data from the contact leg’s sole. After normalizing the vector, it is used as the third column of the target posture rotation matrix. The desired Roll and Pitch angles, which are crucial for our jumping with navigation task on a slope, are then derived using the arc tangent and cosine functions, respectively. Additionally, by substituting the estimated normal vector into the slope equation, the foot's vertical position ($p_z$) concerning the terrain is calculated. This method ensures improved stability and adaptability to challenging surfaces.
\subsubsection{Gait Schedule}
This paper utilizes a time-driven gait scheduler for legged robots designed to manage the duration of the flight and stance phases. It applies a periodic time distribution to differentiate between various gaits across different legs. As described in \cite{kim2019highly}, the scheduler breaks down the timing of different gaits into duty cycles and phase offsets. The duty cycle refers to the proportion of time the leg is in the support phase relative to the entire gait cycle, which includes both the flight and support phases. The offset defines the start time of the support phase for each leg as a fraction of the total gait cycle. In this study, we focus exclusively on the trotting gait for our navigation and jumping tasks, which is noted for its stability across various terrains.
\begin{figure}[htbp]
\centering
\includegraphics[width=3.0in]{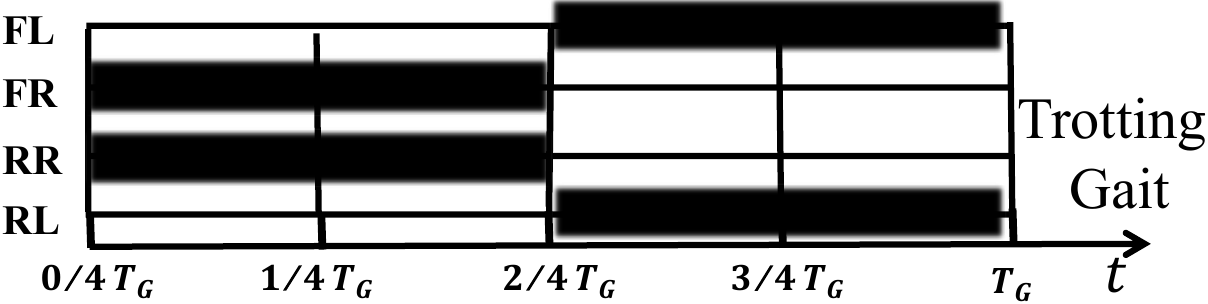}
\caption{The figure illustrates the time-based distribution of the trot gait employed by the quadrupedal robot while tracking the navigation trajectory in this study.}
\label{fig:trot_gait}
\end{figure}

\subsubsection{Mapping and Trajectory Planning for Navigation}
In our navigation system, we utilize FAST-LIO in~ \cite{fastlio} as the localization backbone, combined with our re-localization component. Following the approach in \cite{plan}, we use LiDAR observations to construct an occupancy grid map, obtaining geometrical information about the surroundings. Given a target point, we employ $\mathrm{RRT}^*$ in \cite{karaman2011incremental,lavalle1998rapidly} to generate a collision-free path for the PD tracking controller. We detect ground height in real time; when significant height differences are identified, we trigger the jump controller. This involves calculating the target position for the jump and optimizing the jump trajectory in real-time to successfully cross the current obstacle.

\subsection{Experiments Hardware and Features}\label{sec:exp_hard_fea}
For the real-world experiments, we utilized a quadruped robot based on the open-source MIT Cheetah hardware platform, with detailed robot parameters presented in Table \ref{tab:robot_mass_param}. To simulate and validate the jumping trajectories, a modified version of the MIT Cheetah simulator in~\cite{MIT_GitHub} was employed. Additionally, to demonstrate the versatility of the proposed algorithm, we conducted jumping experiments using Xiaomi’s Cyberdog-1 in~\cite{Midog_GitHub}, whose parameters are also listed in Table \ref{tab:robot_mass_param}. The jumping algorithm was further extended to a bipedal humanoid robot designed and developed by the authors, with its jumping trajectory simulated and verified in Webots; the bipedal robot’s parameters are similarly provided in Table \ref{tab:robot_mass_param}.

To achieve navigation-based jumping, the quadruped robot must accurately track pre-planned trajectories. For this purpose, we implemented Representation-Free Model Predictive Control (RF-MPC in~\cite{ding_2020}) as the locomotion GRFs planner. Unlike traditional linear MPC approaches that utilize Euler angles for state representation, RF-MPC leverages rotation matrices. This approach not only incorporates the yaw angle but also enables more precise tracking of the desired roll and pitch angles, as compared to the linear MPC method presented in \cite{kim2019highly}. The trotting gait, governed by time-based planning as illustrated in Fig. \ref{fig:trot_gait}, was employed for trajectory tracking. Given that the experimental setup includes a slope of approximately 30 degrees, we applied the method described in \cite{bledt2018cheetah} to estimate the slope’s normal vector using foot coordinates, thereby obtaining the desired roll and pitch angles for posture adjustment.

The upper-level navigation planner communicates desired speed commands and jumps target points to the quadruped body controller and Finite-state machine(FSM) via the LCM UDP message queue. Upon detecting a non-empty jump target point, the Finite-state machine seamlessly transitions to the jump controller, enabling coordinated navigation and obstacle traversal.

Performance analyses, including the success rate of jumping trajectory solutions for various target points, time statistics, and pre-motion library computations, were conducted on a laptop equipped with an Intel Core i7-10875H CPU @ 2.30 GHz. Real-time jump trajectory optimization and locomotion were executed on an Intel NUC (i3-8145U @ 2.1 GHz) mounted on the quadruped robot. The state estimation and Whole-Body Control (WBC) modules operated at 500 Hz, while the RF-MPC ran at 45.45 Hz. Furthermore, an additional Intel NUC10 (i7-10710U @ 4.7 GHz) was utilized for environmental mapping and navigation trajectory generation, employing DJI’s MID360 \citep{mid360_lidar} and Intel’s RealSense D430 sensors.

\subsection{Results of Jumping Framework Analysis}\label{sec:jumping_analysis}
To comprehensively verify the real-time performance, stability, versatility, and practical effectiveness of the cascading jumping framework proposed in this paper, several analyses were conducted. For timeliness, the optimization time and success rate for different take-off phases were statistically analyzed based on the x, y, and z coordinates of the jumping target point, with a resolution of 0.05 m. To further demonstrate the time improvement achieved by using the pre-motion library as the initial guess for optimization, the same jumping trajectory generation method was employed, with time statistics collected for cases using the pre-motion library as the initial guess.

Fig. \ref{fig:omini_time_compar} presents the time statistics for omnidirectional jumps with and without the pre-motion library. The figure indicates that, even without the pre-motion library, the average optimization time of the proposed jumping algorithm is mostly within 1 second, as shown by the dark blue region in Fig. \ref{fig:omini_time_compar}. When the pre-motion library is used, nearly all points are optimized within 0.2 seconds, with only a few around 0.4 seconds, demonstrating the efficiency of the optimization process.
The range of $p_{tg}$ for generating and solving the pre-motion library's jumping trajectory, along with the corresponding time consumption statistics, is presented as follows:
\begin{equation}
\begin{aligned}
\bm{p}_{tg}=&\left\{\bm{p}_{tg} \mid x_{tg} \in[-1.0 \, \mathrm{m}, 1.0 \, \mathrm{m}], \right. \\
&y_{tg} \in[-0.55 \, \mathrm{m}, 0.55 \, \mathrm{m}], \\
&z_{tg} \in[0.2 \, \mathrm{m}, 0.6 \, \mathrm{m}] \left.\right\}
\end{aligned}
\end{equation}
To assess the optimization success rate, we conducted a statistical analysis of the success rates for omnidirectional jumping trajectories generated without using the pre-motion library. As shown in Table \ref{tab:stat_compare}, the success rate of trajectory optimization generation exceeds 90\% in all directions and, in some cases, even reaches 100\%. The lowest success rate occurs in the rear-left jump, where a longer target distance results in a slight decrease in the optimization success rate. Table~\ref{tab:stat_range_data} displays the target ranges for the various jumping directions.

To validate the generalizability of the proposed algorithm, we extended the jumping algorithm to a full-size humanoid robot, as detailed in Sec. \ref{sec:humanoid_extension} and Sec. \ref{sec:human_exp}. These experiments highlight the algorithm’s versatility.

Finally, to verify the practicality of the algorithm, extensive simulations, and real-world experiments were conducted, including integration with navigation systems. The subsequent sections provide a detailed description of the experimental results related to the jumping framework.
\begin{figure*}[htbp]
\centering
\includegraphics[width=\linewidth]{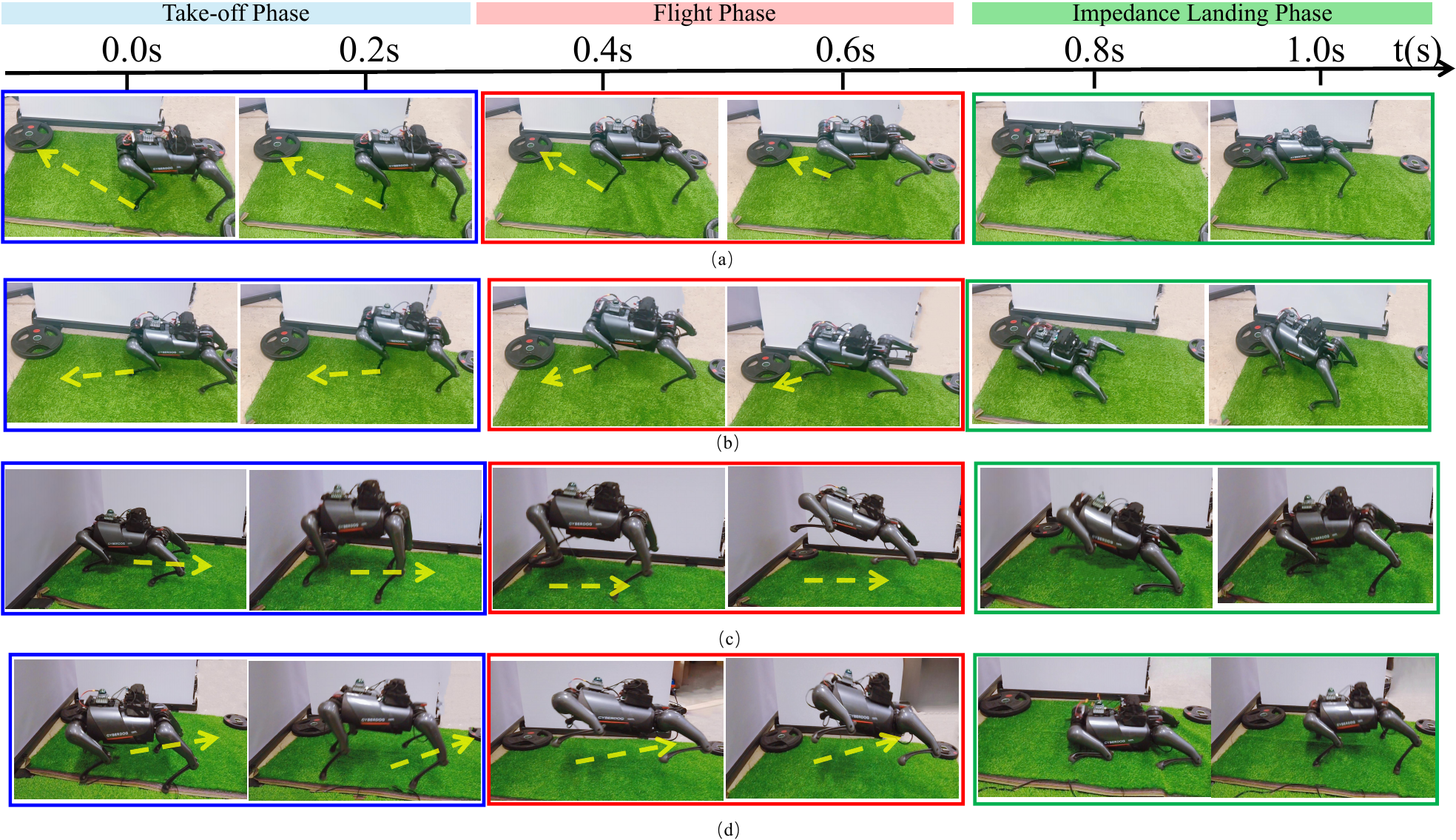}
\caption{Snapshots of the online omnidirectional jump real-world experiments: The blue box represents the takeoff phase, the red box indicates the flight phase, and the green box marks the landing phase. The yellow arrow line shows the direction of the jump. (a) depicts a forward-right (NE) jump, (b) illustrates a forward-left (NW) jump, (c) shows a backward (S) jump, and (d) presents a rear-right (SE) jump. (The details are shown in video:\href{https://youtu.be/1YGi2pMNIdI}{Omni-Jumping})}
\label{fig:show_omini_results}
\end{figure*}
\begin{figure*}[t]
\centering
\includegraphics[width=\linewidth]{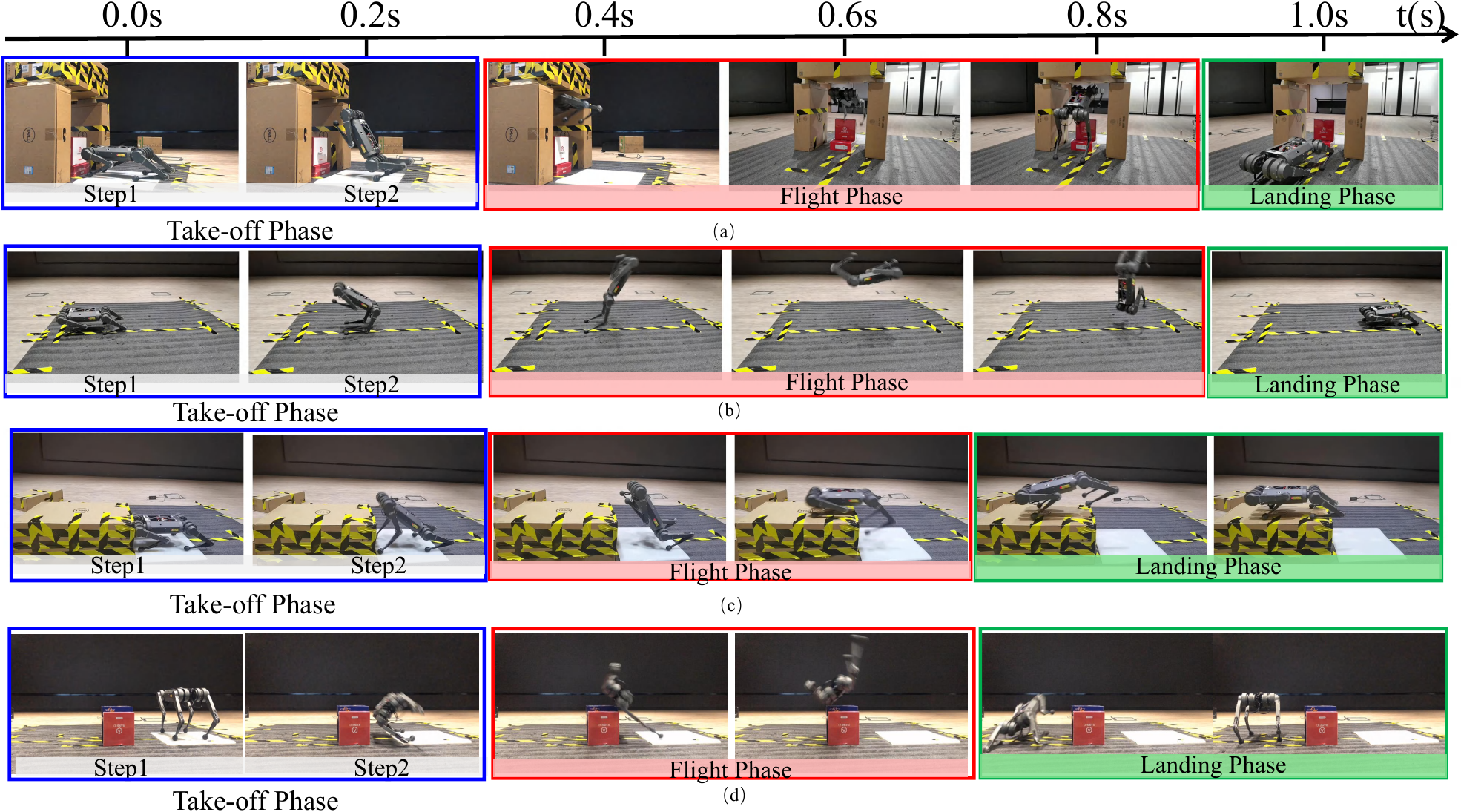}
\caption{Snapshots of offline jump real-world experiments
: the blue box represents the two-step take-off phase, while the red box indicates the flight phase, and the green box shows the landing phase (a) shows a forward jump through a window-shaped obstacle, (b) depicts a backflip, (c) illustrates a jump onto a 30 (cm) high platform, and (d) shows the left-flip over a 32 (cm) high platform. (The details are shown in video: \href{https://youtu.be/PQldIYHTprs}{Agile Motions})}
\label{fig:offline_flip}
\end{figure*}
\begin{table*}[t]
    \scriptsize
    \centering
        \caption{Omnidirectional Solving Success Rate Across Different Target Ranges and Directions.}
     \begin{tabular}{l|c|c|c|c|c|c}
     \hline
     \multirow{6}{*}{} &
      \multicolumn{6}{c}{Jumping Directions} \\
      Target Ranges&  Front (N) & Front Left (NW) & Front Right (NE)  & Left (W)& Rear Left (SW) & Rear Right (SE) \\
      \midrule
    Range 1 & $119/119 (100\%)$ & $1061/1079 (98.33\%)$ & $1063/1079(98.51\%)$ & $55/55 (100.\%)$ & $1067/1079 (98.89\%)$ & $1067/1079 (98.88\%)$ \\
    Range 2 & $146/152 (96.05\%)$ & $1323/1367 (96.78\%)$ & $1338/1367(97.87\%)$ &  $ 71/71 (100.\%)$ & $1298/1367 (94.95\%)$ & $1290/1367 (94.36\%)$\\
    Range 3 & $ 194/199 (97.48\%)$ & $1677/1799 (93.22\%)$ & $1680/1799(93.38\%)$ & $79/79 (100\%)$ & $1314/1511 (86.96\%)$ &$1473/1511 (97.48\%)$ \\
    \bottomrule
    \end{tabular}

    \label{tab:stat_compare}
\end{table*}
\begin{table*}[t]
    \scriptsize
    \centering
        \caption{Target Ranges for Omnidirectional Solving }
     \begin{tabular}{l|c|c}
     \hline
     \multirow{2}{*}{} &
      \multicolumn{2}{c}{Jumping Directions} \\
      Target Ranges& Front Left (NW)\:/Front Right (NE) & Rear Left (SW)\:/Rear Right (SE) \\
      \midrule
    Range 1 & $x_{tg}\in[0.3,1.0]$,\:$y_{tg}\in[-0.6,0.6]$,\: $z_{tg}\in[0.2,0.6]$&  \:$x_{tg}\in[-1.0,-0.3]$,\:$y_{tg}\in[-0.6,0.6]$,\: $z_{tg}\in[0.2,0.6]$ \\
    Range 2 & $x_{tg}\in[0.3,1.2]$,\:$y_{tg}\in[-0.6,0.6]$,\: $z_{tg}\in[0.2,0.6]$&  $x_{tg}\in[-1.2,-0.3]$,\:$y_{tg}\in[-0.6,0.6]$,\: $z_{tg}\in[0.2,0.6]$ \\
    Range 3 & $x_{tg}\in[0.3,1.3]$,\:$y_{tg}\in[-0.6,0.6]$,\: $z_{tg}\in[0.2,0.6]$&  $x_{tg}\in[-1.3,-0.3]$,\:$y_{tg}\in[-0.6,0.6]$,\: $z_{tg}\in[0.2,0.6]$ \\
    \bottomrule
    \hline
      \multicolumn{2}{c}{} \\
      Target Ranges& Front (N)\:/Rear (S) & Left  (W)\:/Right (E) \\
      \midrule
    Range 1 & $x_{tg}\in[0.3,1.0]$,\:$y_{tg}=0.0$,\: $z_{tg}\in[0.2,0.6]$&  $x_{tg}=0.0$,\:$y_{tg}\in[-0.6,0.6]$,\: $z_{tg}\in[0.2,0.6]$ \\
    Range 2 & $x_{tg}\in[0.3,1.2]$,\:$y_{tg}=0.0$, \:$z_{tg}\in[0.2,0.6]$&  $x_{tg}=0.0$,\:$y_{tg}\in[-0.6,0.6]$,\: $z_{tg}\in[0.2,0.6]$ \\
    Range 3 & $x_{tg}\in[0.3,1.3]$,\:$y_{tg}=0.0$,\: $z_{tg}\in[0.2,0.6]$&  $x_{tg}=0.0$,\:$y_{tg}\in[-0.6,0.6]$,\: $z_{tg}\in[0.2,0.6]$ \\
    \bottomrule
    \end{tabular}

    \label{tab:stat_range_data}
\end{table*}

\begin{figure}[htbp]
  \centering
\includegraphics[width=\linewidth]{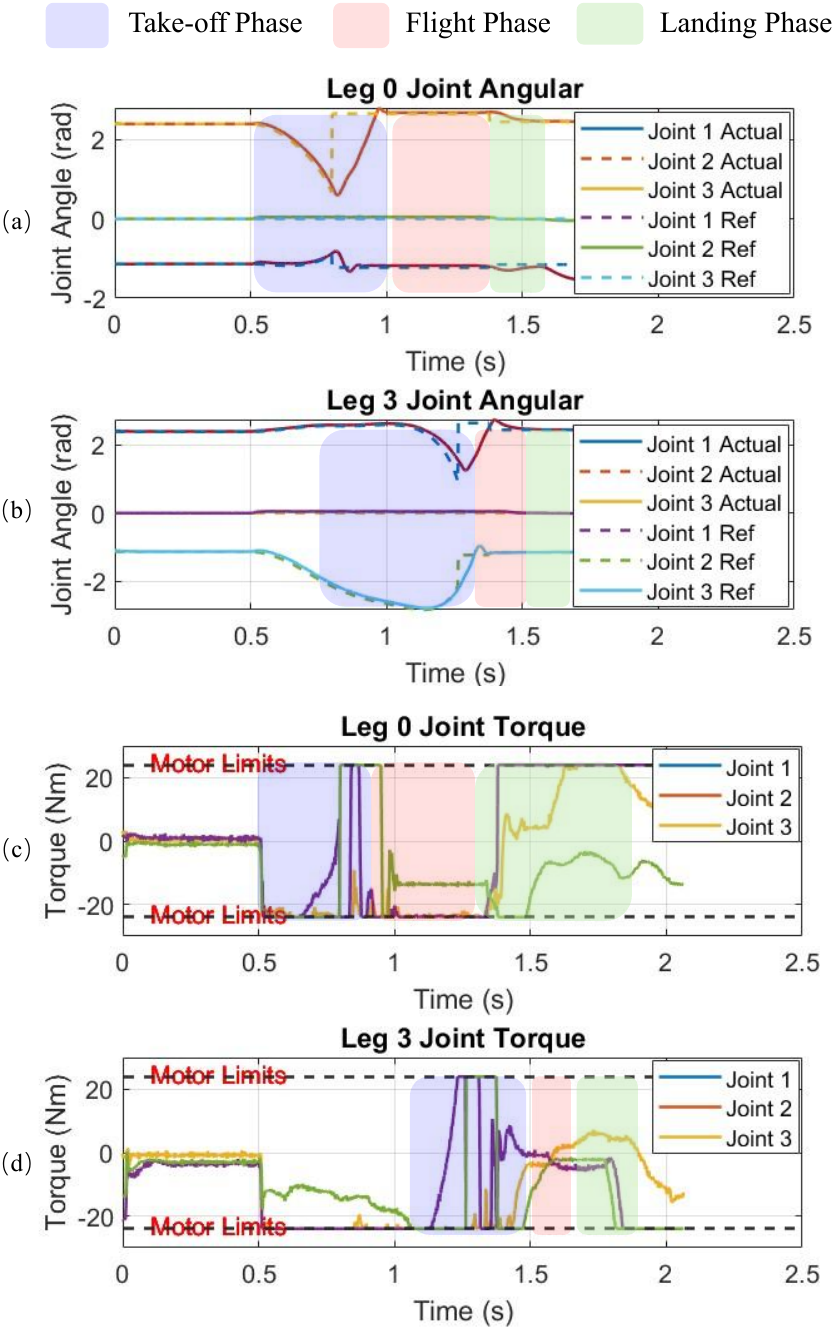}
\vspace{-3mm}
  \caption{Jumping onto a desk: Figures (a) and (c) depict the joint angles of Leg 0 and Leg 3, respectively, while Figures (b) and (d) show the joint torques of Leg 0 and Leg 3.}
  \label{fig:left_flipping_data} 
\end{figure}
\begin{figure}[htbp]
\centering
\includegraphics[width=\linewidth]{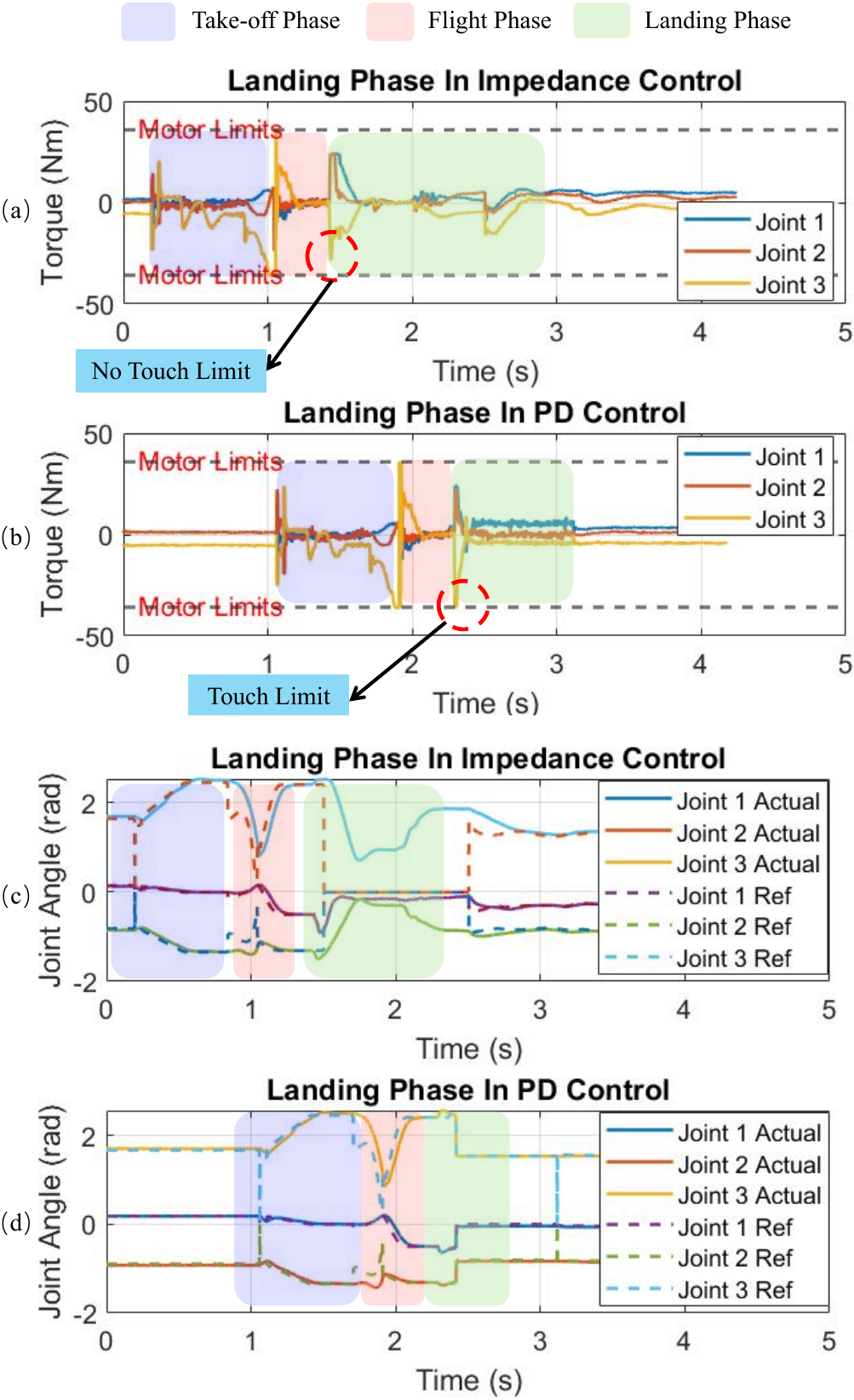}
\caption{
The figure compares Leg 0 torque and joint tracking during forward jumps using impedance control and PD control as landing controllers. The PD control tends to reach the motor limit (36Nm for the knee joint) upon landing, while impedance control, with added active compliance, limits the maximum torque to around 27Nm. (a) shows the torque profile with impedance control. (b) shows the torque profile with PD control. (c) and (d) depict the reference and actual joint angle tracking.}
\label{fig:front_jump_exp_comp}
\end{figure}
\begin{figure}[htbp]
\centering
\includegraphics[width=\linewidth]{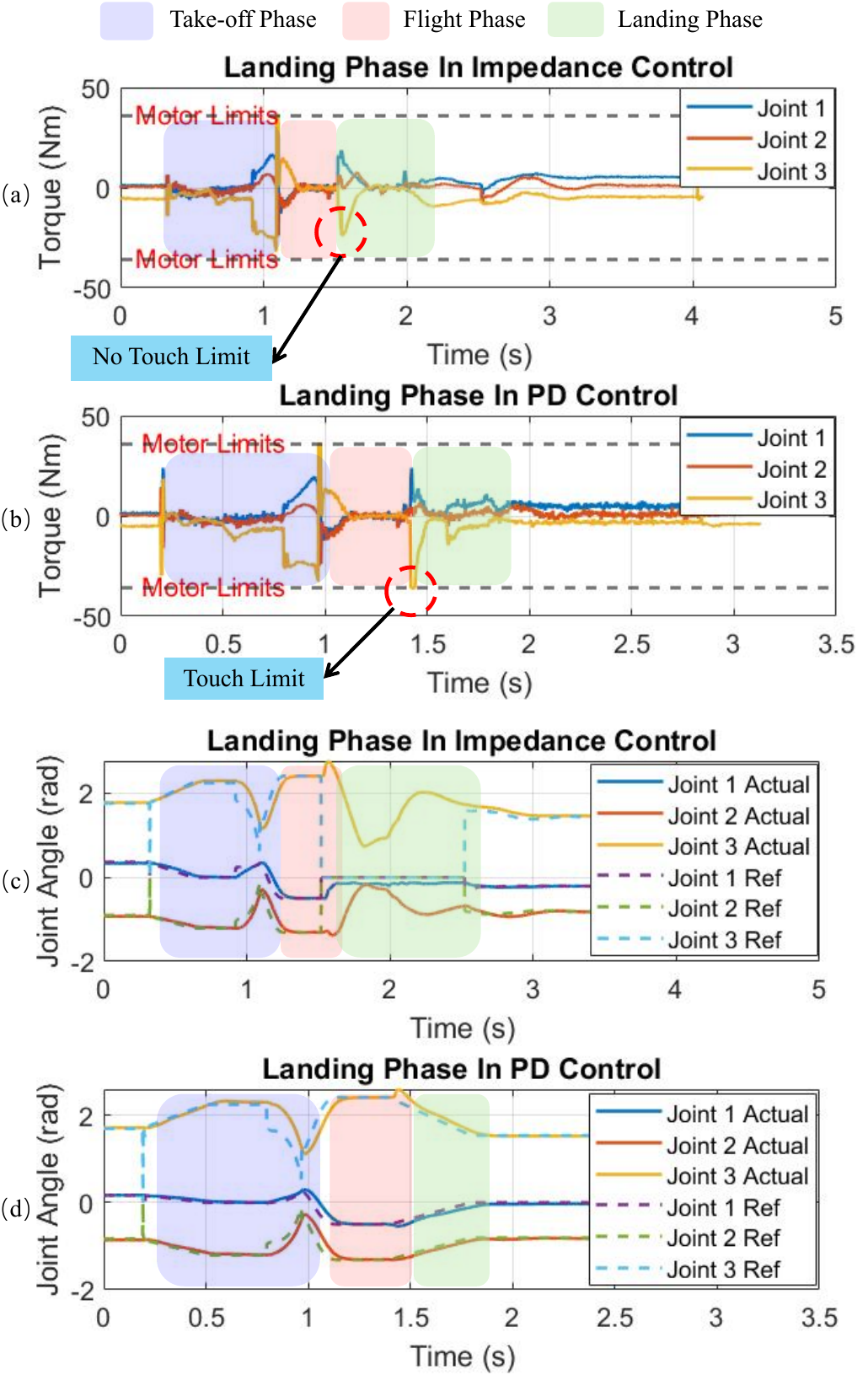}
\caption{The figure compares Leg 0 torque and joint tracking for rear right jumps using impedance and PD control. PD control nears the 36Nm motor limit, while impedance control stays around 25Nm. (a) and (b) show torque profiles, and (c) and (d) show joint angle tracking.}
\label{fig:rear_right_jump_exp_comp}
\end{figure}
\subsection{Experimental Validations}\label{sec:real_exp_vali}
To demonstrate the cascade jump optimization framework proposed in this paper, we conducted real-world experiments involving omnidirectional jumps, including forward, backward, right-front, right-back, left-front, and left-back jumps. Detailed results can be found in Sec.\ref{subsec:different_jump}. To validate the two-step takeoff process, both online and offline real-world experiments were performed. These included a successful execution on a high-jump platform, vertical jumps, and back-flip motions, utilizing pre-optimized trajectories. Further details are provided in Fig. \ref{fig:offline_flip}.

To verify the localization reliability for the jumping task, we integrated mapping and localization with 3D Lidar, slope recognition, traversing trajectory planning, and height-difference detection. Two downward jumps were performed, fully validating the enhancements in the localization by the proposed method. Additionally, to examine the impact of large forces during landing on localization reliability, forward, right-front, and left-front jumps were used for further localization validation.

Furthermore, to highlight the efficacy of the proposed impedance control landing controller, we compare the landing success rates and the frequency of motor limit triggers between PD control and impedance control. See Sec. \ref{subsec:different_landing} for detailed comparisons.

\subsubsection{Jumping with Navigation}\label{subsec:jump_with_navigation}
By jumping, our quadruped robot navigation system can achieve better obstacle traversal capabilities. As shown in Fig. \ref{With_grid_map_long}, when a relatively high platform appears in the planned navigation trajectory, and we cannot pass through it through traditional movement, we can trigger a jump to pass the current obstacle. The jump controller is triggered to determine the height difference between the highest point or the lowest point and the robot's center of mass when the height difference is greater than 0.4m. While jumping improves passability, it also increases the risk of positioning failure. As shown in Fig. \ref{bad}, \cite{fastlio} fails to track the position under the impact of jumping and landing. Therefore, when positioning failure is detected, we will trigger the re-localization to ensure the robustness of positioning. From Fig. \ref{good}, we can find that with the improvement of our re-localization algorithm, the quadruped robot can achieve stable and reliable positioning even in the case of continuous jumping.

\subsubsection{Validation of Various Jumping Motions}\label{subsec:different_jump}
To validate the feasibility of the proposed algorithm, we conducted extensive real-world experiments on omnidirectional jumps, including forward, left, front-right, front-left, rear, and rear-right jumps. Fig. \ref{fig:show_omini_results} provides a snapshot of these experiments. After specifying the target point, the jump trajectory was optimized and executed within 0.1 seconds. Fig. \ref{fig:front_jump_exp_comp} and \ref{fig:rear_right_jump_exp_comp} display the motor torque and position tracking for a full cycle of the forward and rear-right jumps. The data shows that during the take-off phase, the transition from the preparation to the point where the legs leave the ground is completed within 0.3 seconds, with the motor torque nearing the joint’s maximum limit. Additionally, the results demonstrate that the proposed landing controller with active compliance reduces the likelihood of reaching the joint limit during landing.

We further validated the jump construction method described in the paper with two-step take-off jumps, including crossing a window-shaped obstacle, performing a backflip, jumping onto a 30 cm platform, and executing a left flip. Fig. \ref{fig:offline_flip} presents snapshots of these jumps. Notably, the obstacle for the left flip was 32 cm high. Fig. \ref{fig:left_flipping_data} shows the joint information for Leg 0 and 3 during the 30 cm platform jump. Due to the two-step take-off, the front leg leaves the ground first, followed by the rear leg, with a delay of approximately 0.4 seconds between the two (as seen in Figs. \ref{fig:left_flipping_data}(a) and (b)). The torque data in Figs. \ref{fig:left_flipping_data}(c) and (d) further confirms this. Given that the 30 cm platform is the maximum height the robot can jump, both joint 2 and joint 3 reached their torque limits during take-off, and the landing phase also approached the torque limit. These real-world experiments demonstrate the real-time performance, high success rate, and versatility of the proposed algorithm.

\subsubsection{Different Landing Controller Comparison}\label{subsec:different_landing}
This section compares the performance of the PD controller and the impedance controller during forward and rear-right jumps. The target point for the forward jump is $p_{tg}=[1.0, 0.0, 0.25]$, and for the rear-right jump, it is $p_{tg}=[-0.7, -0.4, 0.5]$. In Fig. \ref{fig:front_jump_exp_comp}, for the forward jump, it is evident that the PD controller reaches the motor limit during landing (Fig. \ref{fig:front_jump_exp_comp}(b)), while the impedance controller (Fig. \ref{fig:front_jump_exp_comp}(a)) avoids this issue. Also, the impedance controller takes longer to stabilize after landing than the PD controller. Similarly, in the rear-right jump (Figs. \ref{fig:rear_right_jump_exp_comp}(a) and (b)), the PD controller also reaches the motor limit upon landing, while the impedance controller does not. These results demonstrate that the impedance controller provides better hardware protection during the jump landing.


\subsection{Extention to Humanoid Robot in Simulation}\label{sec:human_exp}
\begin{figure}[htbp]
  \centering
 \subfigure[]{
    \label{bipedal_front} 
    \includegraphics[width=0.445\linewidth]{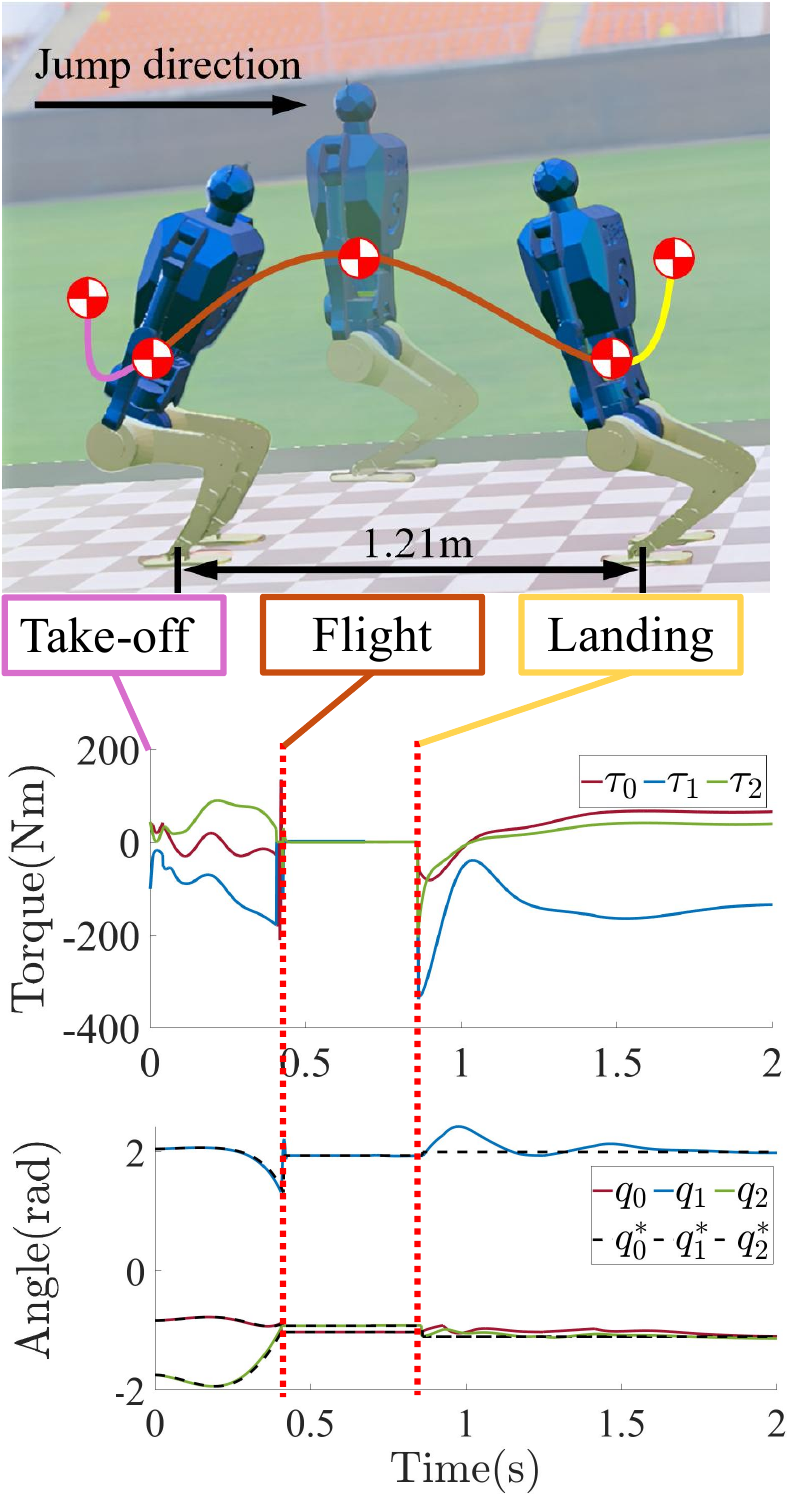}}
  \subfigure[]{
    \label{bipedal_rear_jump} 
    \includegraphics[width=0.445\linewidth]{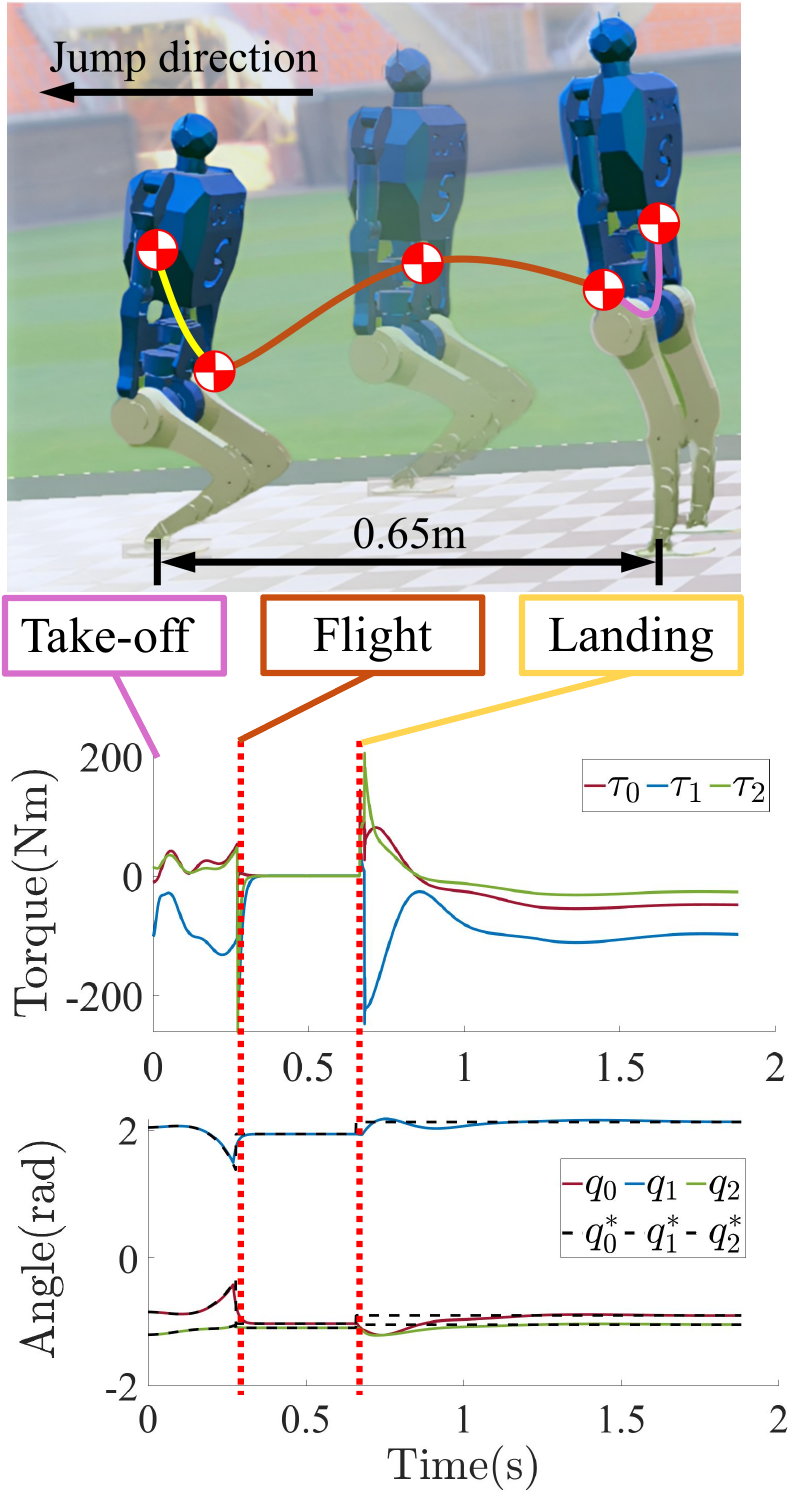}}

  \caption{Humanoid jumping in Webots simulation environment. (a) Joint torque and angle data feedback of backward jumping. (b) Joint torque and angle data feedback of forward jumping. The magenta lines in (a) and (b) represent the CoM trajectories during the take-off phase, the brown lines represent the flight phase, and the yellow lines represent the landing phase. $\tau_i, q_i, q_i^*$ in (3) and (4) represent the Webots feedback torque, joint angle, and desired joint angle, respectively. The subscripts $i$ = 0,1,2 represent the leg ankle, knee, and hip pitch joint, respectively. The red dashed line is the split line for the different jumping phases.}
  \label{Humanoid_jumping_data} 
\end{figure}

In this section, we validated the optimized jumping trajectories from Sec. \ref{sec:humanoid_extension} using the Webots (Ref to~\cite{Webots}) simulation environment, achieving a backward jump of 0.65 m and a forward jump of 1.21 m, as shown in Fig. \ref{Humanoid_jumping_data}. The parameters of the simulation model are shown in Table. \ref{tab:robot_mass_param}, and the torque limits for the ankle, knee, and hip joints are [216,\:320,\:417] (Nm), respectively. In the take-off phase, we map the optimized Center of Mass (CoM) trajectories to joint angles using the kinematic model from Sec. \ref{sec:dyanmics_sec}. Position control is used to drive the robot's motion in Webots. During the flight and landing phases, we set the joints to fixed reference angles and used joint PD control with $K_p$ = 800 and $K_d$ = 30. From Fig.\ref{Humanoid_jumping_data}, we observe that during the take-off phase, the joint angles closely followed the reference values, and the joint torques remained within the torque limits. During the landing phase, the joint PD control requires more time to stabilize the robot due to the impact upon landing and generates higher joint torques compared to the take-off phase, but these torques still remain within limits. Comparing the forward and backward jumps, we find that the joint torques during the take-off and landing phases are higher for the forward jump due to its greater distance. 

\section{Conclusions and Future work}\label{sec:conclution_results}
In this paper, we developed a general approach to modeling the jumping problem and proposed an online jump trajectory planning method that differs from traditional gradient-based approaches. We enabled the quadruped robot to perform omnidirectional jumps in the 2D plane rather than being limited to forward or backward jumps. To minimize motor limit violations during the jumping and landing phases, we introduced impedance control, which outperformed PD control as a landing controller, especially in omnidirectional jumps.

To ensure practical application in real environments, we employed an MPC and WBC framework as the locomotion controller to track the navigation path and used 3D LiDAR for environment scanning and reconstruction to trigger jumps based on height differences. However, we identified that baseline localization methods are prone to failure due to the impact of jumping and landing. To address this, we proposed a coarse-to-fine relocalization method, which ensures accurate positioning despite the disturbances caused by landing impacts.

Through extensive testing in simulations and on real hardware, we achieved an optimization success rate exceeding 90\%. Utilizing the Pre-motion library, we reduced the optimization time to approximately 0.1s, meeting real-time requirements for jump optimization. These results demonstrate the effectiveness, real-time performance, and generality of the proposed online cascade optimization jump framework. Additionally, we extended this approach to humanoid robots and verified it in simulation environments.

Although the use of evolutionary algorithms enabled online jumping without predefined contact schedules or take-off timings, the inherent randomness in the initial guess of the evolutionary algorithm introduces trajectory deviations, resulting in less than 100\% success rates. Furthermore, this paper did not explore the generalization of friction cones across different terrains. For example, achieving stable jumps without slippage across varying surfaces remains challenging. Future work will address these issues, including optimizing jumping trajectories to prevent slippage and improving the overall success rate of trajectory planning.
\section{Acknowledgments}
This work is supported by the InnoHK Clusters of the Hong Kong SAR Government via the Hong Kong Centre for Logistics Robotics and the CUHK T Stone Robotics Institute.
 


\bibliographystyle{SageH.bst}
\bibliography{references}

\begin{thebibliography}{56}
\providecommand{\natexlab}[1]{#1}
\providecommand{\url}[1]{\texttt{#1}}
\providecommand{\urlprefix}{URL }
\expandafter\ifx\csname urlstyle\endcsname\relax
  \providecommand{\doi}[1]{DOI:\discretionary{}{}{}#1}\else
  \providecommand{\doi}{DOI:\discretionary{}{}{}\begingroup \urlstyle{rm}\Url}\fi

\bibitem[{Abdel-Basset et~al.(2023)Abdel-Basset, El-Shahat, Jameel and Abouhawwash}]{EDO_A}
Abdel-Basset M, El-Shahat D, Jameel M and Abouhawwash M (2023) Exponential distribution optimizer (edo): a novel math-inspired algorithm for global optimization and engineering problems.
\newblock \emph{Artificial Intelligence Review} 56(9): 9329--9400.

\bibitem[{Ahmad et~al.(2022)Ahmad, Isa, Lim and Ang}]{ahmad2022differential}
Ahmad MF, Isa NAM, Lim WH and Ang KM (2022) Differential evolution: A recent review based on state-of-the-art works.
\newblock \emph{Alexandria Engineering Journal} 61(5): 3831--3872.

\bibitem[{Ayyarao et~al.(2022)Ayyarao, Ramakrishna, Elavarasan, Polumahanthi, Rambabu, Saini, Khan and Alatas}]{WSO_A}
Ayyarao TSLV, Ramakrishna NSS, Elavarasan RM, Polumahanthi N, Rambabu M, Saini G, Khan B and Alatas B (2022) War strategy optimization algorithm: A new effective metaheuristic algorithm for global optimization.
\newblock \emph{IEEE Access} 10: 25073--25105.

\bibitem[{Ayyub and Lai(1989)}]{ayyub1989structural}
Ayyub BM and Lai KL (1989) Structural reliability assessment using latin hypercube sampling.
\newblock In: \emph{Structural safety and reliability}. ASCE, pp. 1177--1184.

\bibitem[{B{\"a}ck et~al.(1997)B{\"a}ck, Fogel and Michalewicz}]{back1997handbook}
B{\"a}ck T, Fogel DB and Michalewicz Z (1997) Handbook of evolutionary computation.
\newblock \emph{Release} 97(1): B1.

\bibitem[{Bellegarda et~al.(2024)Bellegarda, Nguyen and Nguyen}]{bellegarda2024robust}
Bellegarda G, Nguyen C and Nguyen Q (2024) Robust quadruped jumping via deep reinforcement learning.
\newblock \emph{Robotics and Autonomous Systems} : 104799.

\bibitem[{Bledt et~al.(2018)Bledt, Powell, Katz, Di~Carlo, Wensing and Kim}]{bledt2018cheetah}
Bledt G, Powell MJ, Katz B, Di~Carlo J, Wensing PM and Kim S (2018) Mit cheetah 3: Design and control of a robust, dynamic quadruped robot.
\newblock In: \emph{2018 IEEE/RSJ International Conference on Intelligent Robots and Systems (IROS)}. IEEE, pp. 2245--2252.

\bibitem[{Caccavale et~al.(1999)Caccavale, Natale, Siciliano and Villani}]{caccavale1999six}
Caccavale F, Natale C, Siciliano B and Villani L (1999) Six-dof impedance control based on angle/axis representations.
\newblock \emph{IEEE Transactions on Robotics and Automation} 15(2): 289--300.

\bibitem[{Campbell and Petersson(2016)}]{campbell2016gogma}
Campbell D and Petersson L (2016) Gogma: Globally-optimal gaussian mixture alignment.
\newblock In: \emph{Proceedings of the IEEE conference on computer vision and pattern recognition}. pp. 5685--5694.

\bibitem[{Cheng et~al.(2024)Cheng, Shi, Agarwal and Pathak}]{cheng2024extreme}
Cheng X, Shi K, Agarwal A and Pathak D (2024) Extreme parkour with legged robots.
\newblock In: \emph{2024 IEEE International Conference on Robotics and Automation (ICRA)}. IEEE, pp. 11443--11450.

\bibitem[{Chignoli et~al.(2021{\natexlab{a}})Chignoli, Kim, Stanger-Jones and Kim}]{chignoli2021humanoid}
Chignoli M, Kim D, Stanger-Jones E and Kim S (2021{\natexlab{a}}) The mit humanoid robot: Design, motion planning, and control for acrobatic behaviors.
\newblock In: \emph{2020 IEEE-RAS 20th International Conference on Humanoid Robots (Humanoids)}. IEEE, pp. 1--8.

\bibitem[{Chignoli et~al.(2021{\natexlab{b}})Chignoli, Morozov and Kim}]{chignoli2021rapid}
Chignoli M, Morozov S and Kim S (2021{\natexlab{b}}) Rapid and reliable trajectory planning involving omnidirectional jumping of quadruped robots.
\newblock \emph{arXiv preprint} .

\bibitem[{Choi et~al.(2023)Choi, Ji, Park, Kim, Mun, Lee and Hwangbo}]{choi2023learning}
Choi S, Ji G, Park J, Kim H, Mun J, Lee JH and Hwangbo J (2023) Learning quadrupedal locomotion on deformable terrain.
\newblock \emph{Science Robotics} 8(74): eade2256.

\bibitem[{Di~Carlo et~al.(2018)Di~Carlo, Wensing, Katz, Bledt and Kim}]{di2018dynamic}
Di~Carlo J, Wensing PM, Katz B, Bledt G and Kim S (2018) Dynamic locomotion in the mit cheetah 3 through convex model-predictive control.
\newblock In: \emph{2018 IEEE/RSJ international conference on intelligent robots and systems (IROS)}. IEEE, pp. 1--9.

\bibitem[{Ding et~al.(2020)Ding, Li and Park}]{ding2020kinodynamic}
Ding Y, Li C and Park HW (2020) Kinodynamic motion planning for multi-legged robot jumping via mixed-integer convex program.
\newblock In: \emph{2020 IEEE/RSJ International Conference on Intelligent Robots and Systems (IROS)}. IEEE, pp. 3998--4005.

\bibitem[{{Ding} et~al.(2021){Ding}, {Pandala}, {Li}, {Shin} and {Park}}]{ding_2020}
{Ding} Y, {Pandala} A, {Li} C, {Shin} YH and {Park} HW (2021) Representation-free model predictive control for dynamic motions in quadrupeds.
\newblock \emph{IEEE Transactions on Robotics} : 1--18\doi{10.1109/TRO.2020.3046415}.

\bibitem[{DJI(2020)}]{mid360_lidar}
DJI (2020) Livoxmid360: 3d lidar.
\newblock \urlprefix\url{https://www.livoxtech.com/mid-360}.
\newblock Accessed: 2024-09-19.

\bibitem[{Gilroy et~al.(2021)Gilroy, Lau, Yang, Izaguirre, Biermayer, Xiao, Sun, Agrawal, Zeng, Li et~al.}]{gilroy2021autonomous}
Gilroy S, Lau D, Yang L, Izaguirre E, Biermayer K, Xiao A, Sun M, Agrawal A, Zeng J, Li Z et~al. (2021) Autonomous navigation for quadrupedal robots with optimized jumping through constrained obstacles.
\newblock In: \emph{2021 IEEE 17th International Conference on Automation Science and Engineering (CASE)}. IEEE, pp. 2132--2139.

\bibitem[{Hartley et~al.(2020)Hartley, Ghaffari, Eustice and Grizzle}]{hartley2020contact}
Hartley R, Ghaffari M, Eustice RM and Grizzle JW (2020) Contact-aided invariant extended kalman filtering for robot state estimation.
\newblock \emph{The International Journal of Robotics Research} 39(4): 402--430.

\bibitem[{Hashim and Hussien(2022)}]{SO_A}
Hashim FA and Hussien AG (2022) Snake optimizer: A novel meta-heuristic optimization algorithm.
\newblock \emph{Knowledge-Based Systems} 242: 108320.

\bibitem[{Huang et~al.(2023)Huang, Li, Xiang, Ni, Chi, Li, Yang, Peng and Sreenath}]{huang2023creating}
Huang X, Li Z, Xiang Y, Ni Y, Chi Y, Li Y, Yang L, Peng XB and Sreenath K (2023) Creating a dynamic quadrupedal robotic goalkeeper with reinforcement learning.
\newblock In: \emph{2023 IEEE/RSJ International Conference on Intelligent Robots and Systems (IROS)}. IEEE, pp. 2715--2722.

\bibitem[{Hyun et~al.(2014)Hyun, Seok, Lee and Kim}]{hyun2014high}
Hyun DJ, Seok S, Lee J and Kim S (2014) High speed trot-running: Implementation of a hierarchical controller using proprioceptive impedance control on the mit cheetah.
\newblock \emph{The International Journal of Robotics Research} 33(11): 1417--1445.

\bibitem[{Jenelten et~al.(2020)Jenelten, Miki, Vijayan, Bjelonic and Hutter}]{jenelten2020perceptive}
Jenelten F, Miki T, Vijayan AE, Bjelonic M and Hutter M (2020) Perceptive locomotion in rough terrain--online foothold optimization.
\newblock \emph{IEEE Robotics and Automation Letters} 5(4): 5370--5376.

\bibitem[{Kajita et~al.(2007)Kajita, Nagasaki, Kaneko and Hirukawa}]{kajita2007zmp}
Kajita S, Nagasaki T, Kaneko K and Hirukawa H (2007) Zmp-based biped running control.
\newblock \emph{IEEE robotics \& automation magazine} 14(2): 63--72.

\bibitem[{Karaman(2011)}]{karaman2011incremental}
Karaman S (2011) Incremental sampling-based algorithms for optimal motion planning.
\newblock \emph{Robotics: Science and Systems VI} : 267.

\bibitem[{Katz(2018)}]{katz2018low}
Katz BG (2018) \emph{A low cost modular actuator for dynamic robots}.
\newblock PhD Thesis, Massachusetts Institute of Technology.

\bibitem[{Kim et~al.(2019)Kim, Di~Carlo, Katz, Bledt and Kim}]{kim2019highly}
Kim D, Di~Carlo J, Katz B, Bledt G and Kim S (2019) Highly dynamic quadruped locomotion via whole-body impulse control and model predictive control.
\newblock \emph{arXiv preprint arXiv:1909.06586} .

\bibitem[{Kurtz et~al.(2022)Kurtz, Li, Wensing and Lin}]{kurtz2022mini}
Kurtz V, Li H, Wensing PM and Lin H (2022) Mini cheetah, the falling cat: A case study in machine learning and trajectory optimization for robot acrobatics.
\newblock In: \emph{2022 International Conference on Robotics and Automation (ICRA)}. IEEE, pp. 4635--4641.

\bibitem[{LaValle(1998)}]{lavalle1998rapidly}
LaValle S (1998) Rapidly-exploring random trees: A new tool for path planning.
\newblock \emph{Research Report 9811} .

\bibitem[{Lee et~al.(2020)Lee, Hwangbo, Wellhausen, Koltun and Hutter}]{lee2020learning}
Lee J, Hwangbo J, Wellhausen L, Koltun V and Hutter M (2020) Learning quadrupedal locomotion over challenging terrain.
\newblock \emph{Science robotics} 5(47): eabc5986.

\bibitem[{Lee et~al.(2014)Lee, Hyun, Ahn, Kim and Hogan}]{lee2014dynamics}
Lee J, Hyun DJ, Ahn J, Kim S and Hogan N (2014) On the dynamics of a quadruped robot model with impedance control: Self-stabilizing high speed trot-running and period-doubling bifurcations.
\newblock In: \emph{2014 IEEE/RSJ International Conference on Intelligent Robots and Systems}. IEEE, pp. 4907--4913.

\bibitem[{Li et~al.(2023)Li, Peng, Abbeel, Levine, Berseth and Sreenath}]{li2023robust}
Li Z, Peng XB, Abbeel P, Levine S, Berseth G and Sreenath K (2023) Robust and versatile bipedal jumping control through multi-task reinforcement learning.
\newblock \emph{arXiv preprint arXiv:2302.09450} 1.

\bibitem[{Lin et~al.(2021)Lin, Chen and Yao}]{lin2021unified}
Lin Y, Chen Z and Yao B (2021) Unified method for task-space motion/force/impedance control of manipulator with unknown contact reaction strategy.
\newblock \emph{IEEE Robotics and Automation Letters} 7(2): 1478--1485.

\bibitem[{Liu et~al.(2023)Liu, Xu, Yuan, Mou and Wang}]{liu2023distance}
Liu Q, Xu D, Yuan B, Mou Z and Wang M (2023) Distance-controllable long jump of quadruped robot based on parameter optimization using deep reinforcement learning.
\newblock \emph{IEEE Access} .

\bibitem[{Liu et~al.(2018)Liu, Wang, Song and Wang}]{liu2018efficient}
Liu Y, Wang C, Song Z and Wang M (2018) Efficient global point cloud registration by matching rotation invariant features through translation search.
\newblock In: \emph{Proceedings of the European Conference on Computer Vision (ECCV)}. pp. 448--463.

\bibitem[{Mastalli et~al.(2020)Mastalli, Havoutis, Focchi, Caldwell and Semini}]{mastalli2020motion}
Mastalli C, Havoutis I, Focchi M, Caldwell DG and Semini C (2020) Motion planning for quadrupedal locomotion: Coupled planning, terrain mapping, and whole-body control.
\newblock \emph{IEEE Transactions on Robotics} 36(6): 1635--1648.

\bibitem[{Mirjalili et~al.(2017)Mirjalili, Gandomi, Mirjalili, Saremi, Faris and Mirjalili}]{SSA_A}
Mirjalili S, Gandomi AH, Mirjalili SZ, Saremi S, Faris H and Mirjalili SM (2017) Salp swarm algorithm: A bio-inspired optimizer for engineering design problems.
\newblock \emph{Advances in engineering software} 114: 163--191.

\bibitem[{mit biomimetics(2018)}]{MIT_GitHub}
mit biomimetics (2018) Cheetah-software: Github repository.
\newblock \urlprefix\url{https://github.com/mit-biomimetics/Cheetah-Software}.
\newblock Accessed: 2018-09-19.

\bibitem[{Nguyen et~al.(2019)Nguyen, Powell, Katz, Di~Carlo and Kim}]{nguyen2019optimized}
Nguyen Q, Powell MJ, Katz B, Di~Carlo J and Kim S (2019) Optimized jumping on the mit cheetah 3 robot.
\newblock In: \emph{2019 International Conference on Robotics and Automation (ICRA)}. IEEE, pp. 7448--7454.

\bibitem[{Pandala et~al.(2019)Pandala, Ding and Park}]{pandala2019qpswift}
Pandala AG, Ding Y and Park HW (2019) qpswift: A real-time sparse quadratic program solver for robotic applications.
\newblock \emph{IEEE Robotics and Automation Letters} 4(4): 3355--3362.

\bibitem[{Rudin et~al.(2021)Rudin, Kolvenbach, Tsounis and Hutter}]{rudin2021cat}
Rudin N, Kolvenbach H, Tsounis V and Hutter M (2021) Cat-like jumping and landing of legged robots in low gravity using deep reinforcement learning.
\newblock \emph{IEEE Transactions on Robotics} 38(1): 317--328.

\bibitem[{Shan and Englot(2018)}]{shan2018lego}
Shan T and Englot B (2018) Lego-loam: Lightweight and ground-optimized lidar odometry and mapping on variable terrain.
\newblock In: \emph{2018 IEEE/RSJ International Conference on Intelligent Robots and Systems (IROS)}. IEEE, pp. 4758--4765.

\bibitem[{Shan et~al.(2020)Shan, Englot, Meyers, Wang, Ratti and Rus}]{liosam}
Shan T, Englot B, Meyers D, Wang W, Ratti C and Rus D (2020) Lio-sam: Tightly-coupled lidar inertial odometry via smoothing and mapping.
\newblock In: \emph{2020 IEEE/RSJ International Conference on Intelligent Robots and Systems (IROS)}. pp. 5135--5142.
\newblock \doi{10.1109/IROS45743.2020.9341176}.

\bibitem[{Siciliano et~al.(2008)Siciliano, Sciavicco, Villani and Oriolo}]{Siciliano_}
Siciliano B, Sciavicco L, Villani L and Oriolo G (2008) \emph{Robotics: Modelling, Planning and Control}.
\newblock 1st edition. Springer Publishing Company, Incorporated.
\newblock ISBN 1846286417.

\bibitem[{Song et~al.(2022)Song, Yue, Sun, Ling, Wei, Gui and Liu}]{song2022optimal}
Song Z, Yue L, Sun G, Ling Y, Wei H, Gui L and Liu YH (2022) An optimal motion planning framework for quadruped jumping.
\newblock In: \emph{2022 IEEE/RSJ International Conference on Intelligent Robots and Systems (IROS)}. IEEE, pp. 11366--11373.

\bibitem[{Trojovský and Dehghani(2023)}]{SABO_A}
Trojovský P and Dehghani M (2023) Subtraction-average-based optimizer: A new swarm-inspired metaheuristic algorithm for solving optimization problems.
\newblock \emph{Biomimetics} 8(2).

\bibitem[{Wang et~al.(2022)Wang, Liu and Tan}]{wang2022surrogate}
Wang W, Liu HL and Tan KC (2022) A surrogate-assisted differential evolution algorithm for high-dimensional expensive optimization problems.
\newblock \emph{IEEE Transactions on Cybernetics} 53(4): 2685--2697.

\bibitem[{Webots(2023)}]{Webots}
Webots (2023) http://www.cyberbotics.com.
\newblock \urlprefix\url{http://www.cyberbotics.com}.
\newblock Open-source Mobile Robot Simulation Software.

\bibitem[{Wu and Sreenath(2015)}]{wu2015variation}
Wu G and Sreenath K (2015) Variation-based linearization of nonlinear systems evolving on $\mathit{SO}(3)$ and $\mathbb{S}^{2}$.
\newblock \emph{IEEE Access} 3: 1592--1604.

\bibitem[{xiaomi(2023)}]{Midog_GitHub}
xiaomi (2023) Miroboticslab: Github repository.
\newblock \urlprefix\url{https://github.com/MiRoboticsLab/cyberdog_motor_sdk}.
\newblock Accessed: 2023-09-19.

\bibitem[{Xu et~al.(2022)Xu, Cai, He, Lin and Zhang}]{fastlio}
Xu W, Cai Y, He D, Lin J and Zhang F (2022) Fast-lio2: Fast direct lidar-inertial odometry.
\newblock \emph{IEEE Transactions on Robotics} 38(4): 2053--2073.

\bibitem[{Yang et~al.(2015)Yang, Li, Campbell and Jia}]{yang2015go}
Yang J, Li H, Campbell D and Jia Y (2015) Go-icp: A globally optimal solution to 3d icp point-set registration.
\newblock \emph{IEEE transactions on pattern analysis and machine intelligence} 38(11): 2241--2254.

\bibitem[{Yue et~al.(2023)Yue, Song, Zhang, Zeng, Zhang and Liu}]{yue2023evolutionary}
Yue L, Song Z, Zhang H, Zeng X, Zhang L and Liu YH (2023) Evolutionary-based online motion planning framework for quadruped robot jumping.
\newblock In: \emph{2023 IEEE/RSJ International Conference on Intelligent Robots and Systems (IROS)}. IEEE, pp. 767--773.

\bibitem[{Zhang et~al.(2014)Zhang, Singh et~al.}]{zhang2014loam}
Zhang J, Singh S et~al. (2014) Loam: Lidar odometry and mapping in real-time.
\newblock In: \emph{Robotics: Science and systems}, volume~2. Berkeley, CA, pp. 1--9.

\bibitem[{Zhou et~al.(2021)Zhou, Pan, Gao and Shen}]{plan}
Zhou B, Pan J, Gao F and Shen S (2021) Raptor: Robust and perception-aware trajectory replanning for quadrotor fast flight.
\newblock \emph{IEEE Transactions on Robotics} 37(6): 1992--2009.
\newblock \doi{10.1109/TRO.2021.3071527}.

\bibitem[{Zhuang et~al.(2023)Zhuang, Fu, Wang, Atkeson, Schwertfeger, Finn and Zhao}]{zhuang2023robot}
Zhuang Z, Fu Z, Wang J, Atkeson C, Schwertfeger S, Finn C and Zhao H (2023) Robot parkour learning.
\newblock \emph{arXiv preprint arXiv:2309.05665} .

\end{thebibliography}

\clearpage
\end{document}